\documentclass[lettersize,journal]{IEEEtran}
\usepackage{amsmath,amsfonts}
\usepackage[ruled,vlined]{algorithm2e}
\usepackage{array}
\usepackage[caption=true,font=normalsize,labelfont=sf,textfont=sf]{subfig}
\usepackage{textcomp}
\usepackage{stfloats}
\usepackage{url}
\usepackage{verbatim}
\usepackage{graphicx}
\usepackage{cite}
\usepackage{hyperref}
\usepackage{multirow}
\usepackage{cleveref}
\usepackage{threeparttable}
\usepackage{xcolor}
\usepackage{amsmath}
\usepackage{amssymb}
\usepackage{tabularx}
\usepackage{caption}
\usepackage{booktabs}
\usepackage{algorithm2e}

\hypersetup{
    colorlinks=true, 
    linkcolor=blue,  
    filecolor=blue,  
    urlcolor=black,   
    citecolor=blue,  
    pdfborder={0 0 0} 
}

\Crefname{figure}{Fig.}{Figs.}
\Crefname{equation}{Eq.}{Eqs.}

\def\eg{e.g.}

\def\etal{\emph{et al.}}
\def\ie{{i.e.}}

\newcommand{\verticaltext}[3][0pt]{%
  \raisebox{#1}{\parbox[t]{1em}{\rotatebox[origin=c]{90}{#2 #3}}}%
}

\begin{document}

\title{Fast-Convergent and Communication-Alleviated Heterogeneous Hierarchical Federated Learning \\in Autonomous Driving}

\author{Wei-Bin Kou, Qingfeng Lin, Ming Tang, Rongguang Ye, Shuai Wang, \\Guangxu Zhu*, Yik-Chung Wu*
\thanks{Wei-Bin Kou, Qingfeng Lin and Yik-Chung Wu are with the Department of Electrical and Electronic Engineering, The University of Hong Kong, Hong Kong, China.}
\thanks{Guangxu Zhu is with Shenzhen Research Institute of Big Data, Shenzhen, China.}
\thanks{Ming Tang and Rongguang Ye are with the Department of Computer Science and Engineering, Southern University of Science and Technology, Shenzhen, China.}
\thanks{Shuai Wang is with Shenzhen Institute of Advanced Technology, Chinese Academy of Sciences, Shenzhen, China.}
\thanks{\textit{(Corresponding author: Guangxu Zhu and Yik-Chung Wu.)}}
}

\markboth{IEEE TRANSACTIONS ON INTELLIGENT TRANSPORTATION SYSTEMS,~Vol.~xx, No.~x, August~2024}%
{Shell \MakeLowercase{\textit{et al.}}: A Sample Article Using IEEEtran.cls for IEEE Journals}

\maketitle

\begin{abstract}
\underline{S}treet \underline{S}cene \underline{S}emantic \underline{U}nderstanding (denoted as TriSU) is a complex task for autonomous driving (AD). However, inference model trained from data in a particular geographical region faces poor generalization when applied in other regions due to inter-city data domain-shift. Hierarchical Federated Learning (HFL) offers a potential solution for improving TriSU model generalization by collaborative privacy-preserving training over distributed datasets from different cities. Unfortunately, it suffers from slow convergence because data from different cities are with disparate statistical properties. Going beyond existing HFL methods, we propose a Gaussian heterogeneous HFL algorithm (FedGau) to address inter-city data heterogeneity so that convergence can be accelerated. In the proposed FedGau algorithm, both single RGB image and RGB dataset are modelled as Gaussian distributions for aggregation weight design. This approach not only differentiates each RGB image by respective statistical distribution, but also exploits the statistics of dataset from each city in addition to the conventionally considered data volume. With the proposed approach, the convergence is accelerated by 35.5\%-40.6\% compared to existing state-of-the-art (SOTA) HFL methods. On the other hand, to reduce the involved communication resource, we further introduce a novel performance-aware adaptive resource scheduling (AdapRS) policy. Unlike the traditional static resource scheduling policy that exchanges a fixed number of models between two adjacent aggregations, AdapRS adjusts the number of model aggregation at different levels of HFL so that unnecessary communications are minimized. Extensive experiments demonstrate that AdapRS saves 29.65\% communication overhead compared to conventional static resource scheduling policy while maintaining almost the same performance.
\end{abstract}


\section{Introduction}
\label{sec1}
\underline{\textbf{S}}treet \underline{\textbf{S}}cene \underline{\textbf{S}}emantic \underline{\textbf{U}}nderstanding (denoted as TriSU) is a crucial but complex task for autonomous driving (AD) \cite{rizzoli2022multimodal,10337777,9913352,kou2024adverse}. Recently, a number of new approaches \cite{can2022understanding,natan2022towards} for TriSU have been proposed, achieving impressive results. For example, \cite{wan2022edge} utilizes a multi-modal linear feature combination to segment vehicles from video frames; \cite{nesti2022evaluating} utilizes 3D world geometric information to craft adversarial patches against real-world attacks in the realm of AD; \cite{can2022understanding} introduces road view semantic comprehension using onboard Bird's Eye View (BEV) cameras; \cite{natan2022towards} designs an adaptive loss weighting scheme to surmount the imbalanced learning issue and utilizes multi-sensor fusion technique to enable better understanding of a dynamically changing environment. However, these methods typically face a challenge in generalization, even in relatively minor domain-shift \cite{muhammad2022vision}. This challenge becomes more pronounced when dealing with significant inter-city environmental changes.

\begin{figure*}[t]
\includegraphics[width=\linewidth]{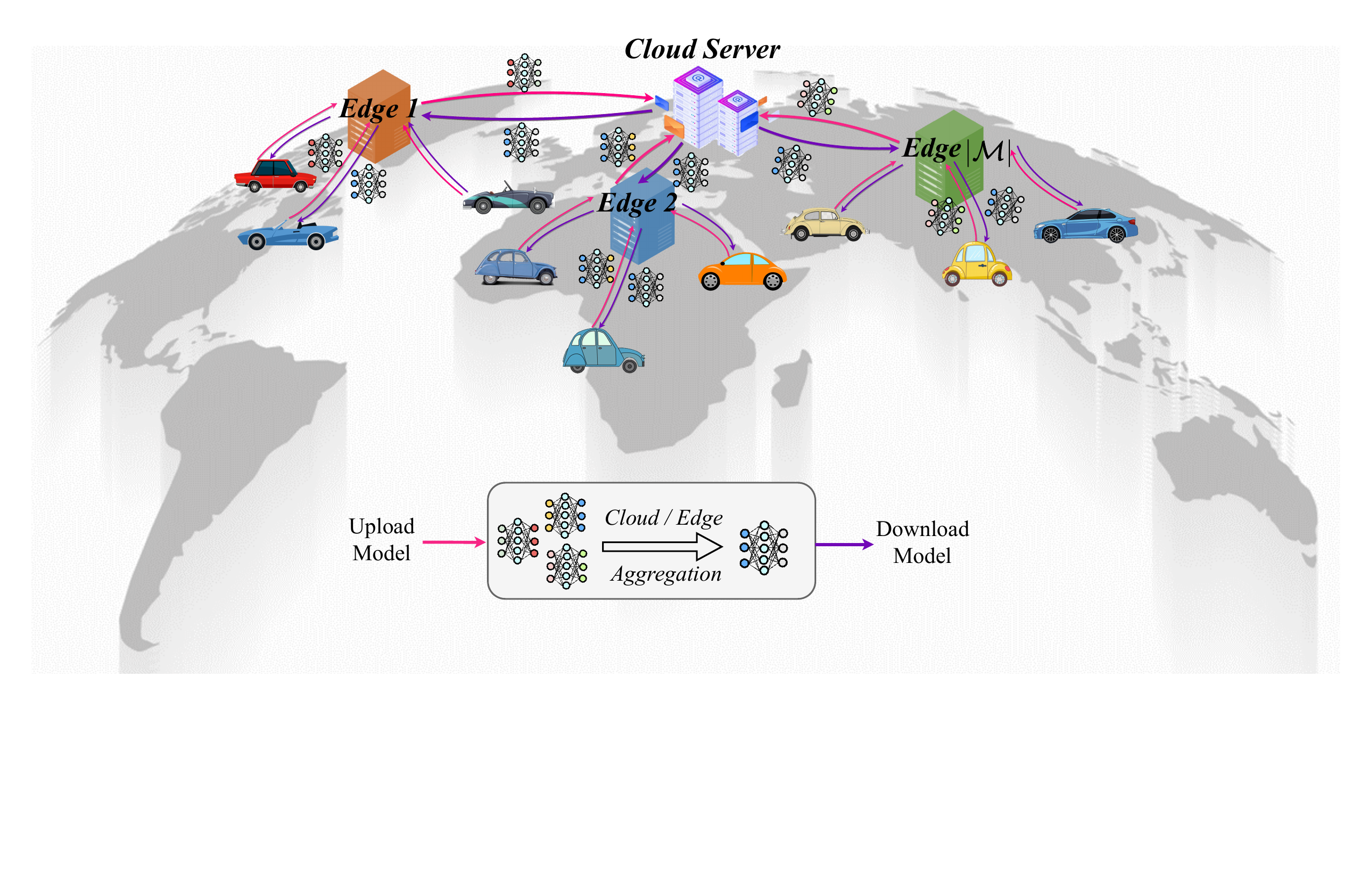}
\vspace{-3.2cm}
\caption{Illustration of HFL in inter-city setting. $\mathcal{M}$ is the set of participating cities.}
\label{Fig.HFL_TriSU}
\vspace{-0.5cm}
\end{figure*}

Federated Learning (FL) \cite{https://doi.org/10.48550/arxiv.1602.05629,feddrive2022,kou2024pfedlvm,10346209,9831009,10324362,10101681,9505307} offers a potential solution to enhance model generalization in inter-city setting but lack of flexibility and scalability when involving a large number of participating vehicles, where a massive number of direct connections to a central cloud server is impractical. To enhance the flexibility and scalability in FL, Hierarchical Federated Learning (HFL) \cite{liu2019clientedgecloud,kou2023communication,wu2024hierarchical} provides a promising alternative by introducing mid-point edge servers. 
Specifically, in additional to the cloud server used in traditional FL, HFL establishes an edge server in each city. All participating vehicles in each city communicate their TriSU models with the edge server, and the global cloud server aggregates models from all the edge servers. In this way, the cloud server only needs to communicate with the edge servers, which is a much more manageable task. The HFL in the context of TriSU is summarized in \Cref{Fig.HFL_TriSU}.

The training process of HFL involves multiple $rounds$ of model aggregations \cite{kou2024fedrc}. Each HFL $round$ involves: (\textbf{I}) multiple edge aggregations; and followed by (\textbf{II}) a single cloud aggregation. In stage (\textbf{I}), TriSU model aggregation at each edge server occurs through weighted averaging of all connected vehicles' models. Traditionally, the weight is defined as the ratio of local data size from each vehicle to the size of the total dataset covered by the edge server. In this stage, owing to low data heterogeneity within one city, the aggregated model converges fast, where $proportion$-based weight approximately represents the vehicle's contribution in edge aggregation. On the other hand, in the cloud aggregation at stage (\textbf{II}), due to the significant inter-city heterogeneity, the model converges slowly or even diverges. This is due to the fact that the conventional $proportion$-based weight (i.e., the $proportion$ of the size of dataset covered by the edge server  compared to the size of the total dataset covered by the cloud server) treats all RGB images to be of equal importance. Without measuring the data heterogeneity among inter-city datasets, HFL model converges slowly, which is a well-known phenomenon in HFL.

Recently, the research community begins to explore aggregation weight beyond data size proportionality. For example, \cite{wu2021fast} uses the gradient gap between edge server and cloud server as a criterion to design the aggregation weight. Unfortunately, this design does not measure the data heterogeneity among inter-city datasets, thus suffers from slow convergence of HFL in the context of TriSU. On the other hand, \cite{10.1145/3594779} defines the aggregation weight based on the histograms of pixels from each RGB image, but inappropriately large or small bin sizes can obscure important details or exaggerate minor variations in the data. Worse still, the involved histograms transmission consumes already stringent communication resource and potentially leaks privacy. 

\begin{figure*}[tp]
\centering
\subfloat[Gaussian distribution of a single RGB image]{\includegraphics[width=0.46\linewidth, height=0.26\linewidth]{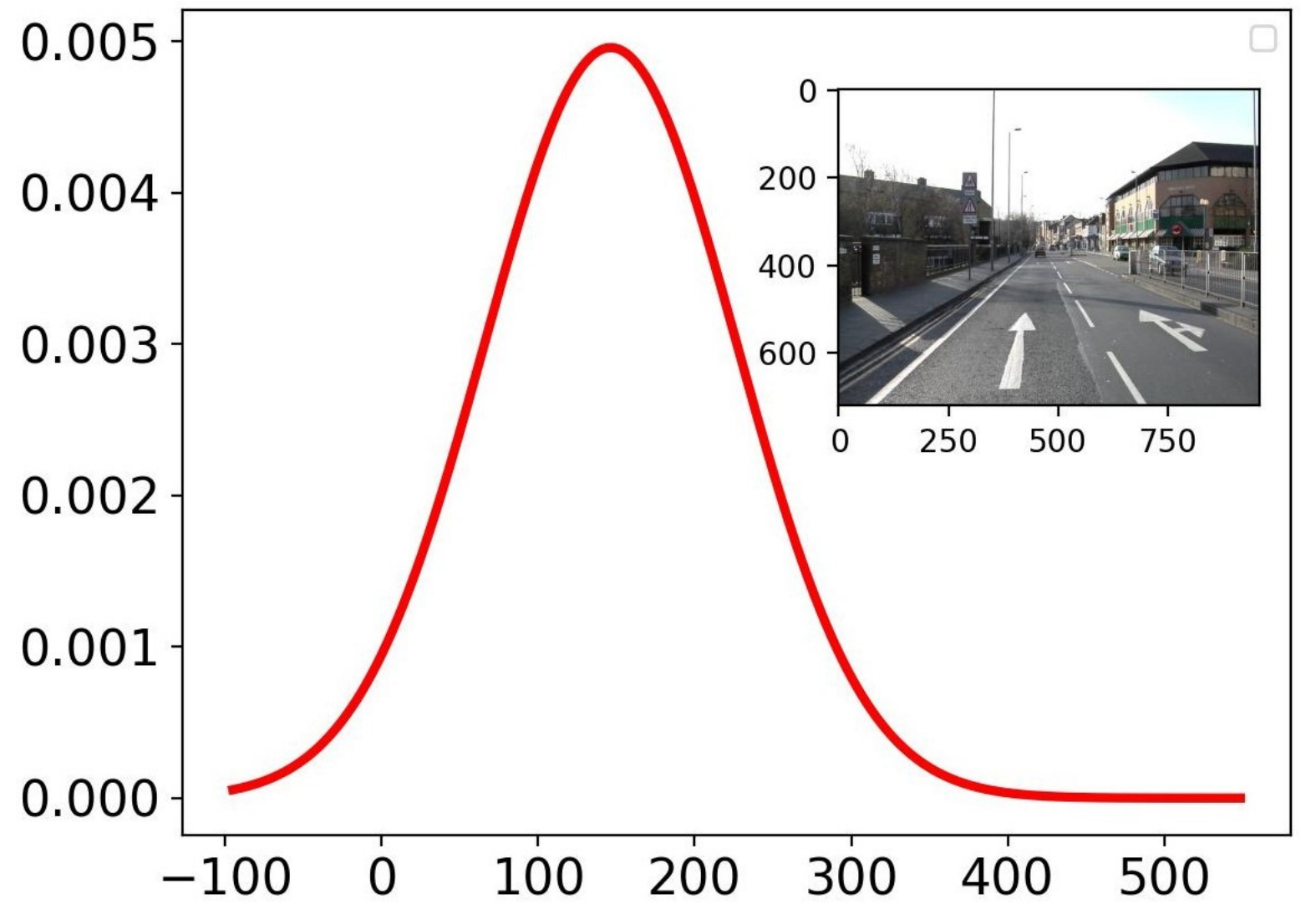}
\label{Fig.rgb_gaussian}
}
\hspace{0.02\linewidth}
\subfloat[Gaussian distribution of RGB dataset]{\includegraphics[width=0.46\linewidth, height=0.26\linewidth]{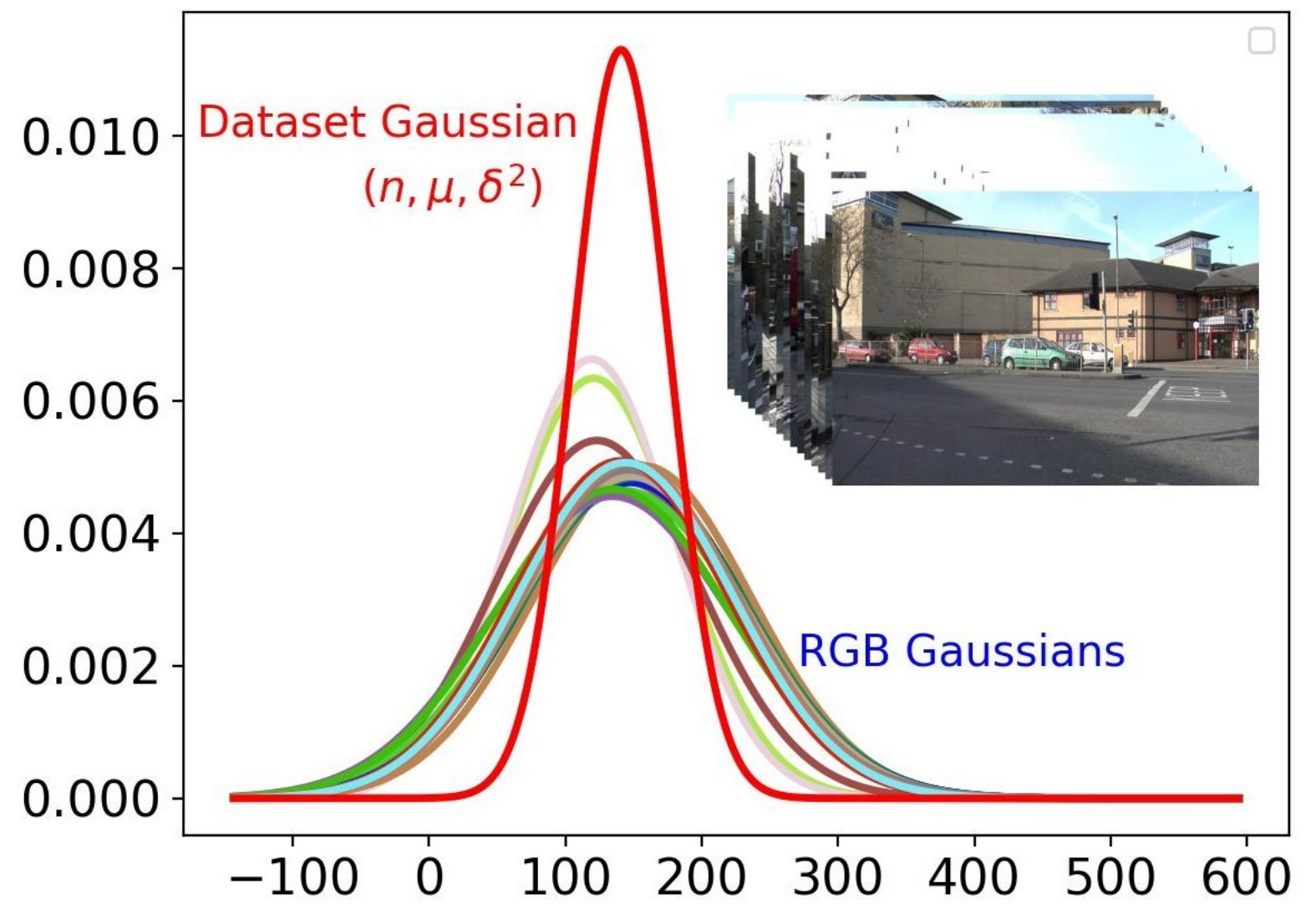}
\label{Fig.dataset_gaussian}
}
\caption{(a) Gaussian distribution of a single RGB image's pixel values. (b) Gaussian distribution of RGB dataset estimated by averaging Gaussian distributions of all included RGB images. $n, \mu, \delta^2$ represent the dataset size, mean and variance of dataset Gaussian distribution, respectively.
}
\label{Fig:rgb_related_gaussian}
\vspace{-0.5cm}
\end{figure*}

\begin{figure*}[t]
\centering 
\includegraphics[width=\linewidth]{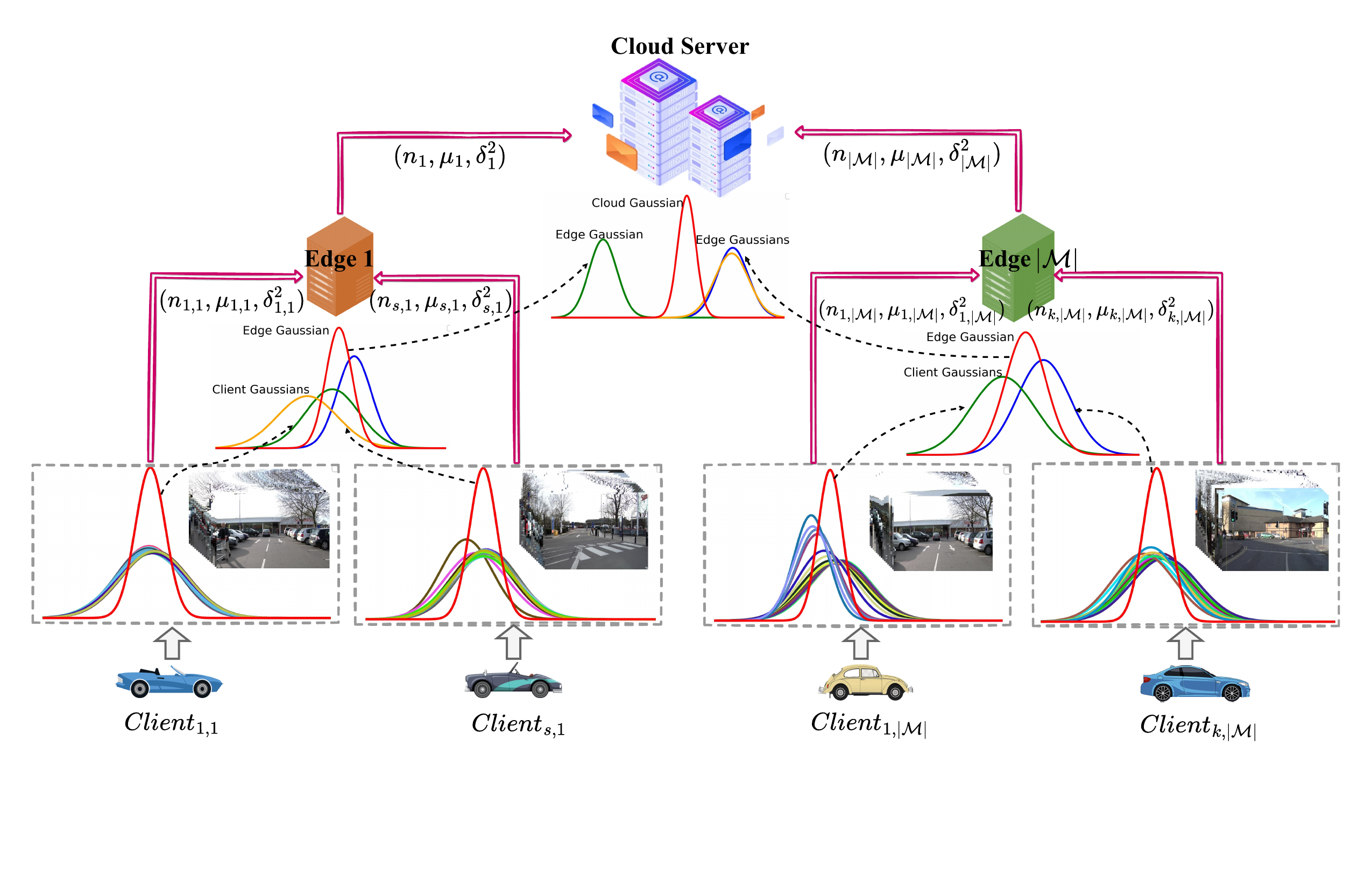}
\vspace{-2.5cm}
\caption{Overview of FedGau. In FedGau, datasets on vehicles or covered by edge servers and cloud server, are all modelled as Gaussian distributions, which are subsequently utilized to measure data heterogeneity and to accelerate HFL model convergence.}
\label{Fig.fedgau}
\vspace{-0.5cm}
\end{figure*}

Therefore, how to design an aggregation weight to handle HFL inter-city data heterogeneity is still an open problem. To this end, this paper presents the FedGau algorithm. Concretely, the proposed FedGau exploits three strategies: \textbf{(I)} we model the distribution of each RGB image's pixel values as a Gaussian distribution (\Cref{Fig.rgb_gaussian}); \textbf{(II)} we represent the RGB dataset on each vehicle using a Gaussian distribution by averaging all RGB images' Gaussian distributions (\Cref{Fig.dataset_gaussian}); and (\textbf{III}) we further calculate the Gaussian distribution of RGB dataset covered by each edge server (or cloud server) by averaging all connected vehicles' Gaussian distributions (or edge servers' Gaussian distributions).
With each dataset modelled as Gaussian distributed, the discrepancy between vehicle's dataset and the corresponding edge server's dataset can be readily computed using the Bhattacharyya distance \cite{bhattacharyya1943measure}, from which the weighting of each vehicle's model at edge aggregation is designed. Furthermore, the Bhattacharyya distance between edge server's dataset and cloud server's dataset can be measured, which subsequently is utilized to calculate aggregation weight of edge server's model at the cloud server. In general, smaller Bhattacharyya distance yield higher weight, as it prioritizes the integration of more closely related RGB images. This weight design facilitates faster convergence of HFL TriSU model. FedGau framework is summarized in \Cref{Fig.fedgau}.

Although FedGau's success in speeding up the convergence enables saving of the communication resource, there still exists room to further reduce communication resource due to the conventional performance-agnostic static communication resource scheduling (StatRS) policy \cite{8664630}. In StatRS, each $round$ contains a single cloud aggregation and the fixed number of edge aggregation, therefore consuming fixed amount of communication resource in each $round$.
Thus, as the training progresses, the amount of communication resource consumption grows linearly, but HFL TriSU model performance (e.g., mIoU) grows more and more slowly. The mismatch between the usage of communication resource and model performance improvement inevitably leads to unnecessary communication resource consumption. 
Therefore, this paper also proposes a performance-aware adaptive communication resource scheduling (AdapRS) policy to dynamically adjust the number of edge aggregation and the number of vehicles' local updates between two adjacent edge aggregations. In AdapRS, the required communication resource varies with performance of the HFL system, thus further saving unnecessary model exchanges. 

To summarize, our main contributions are highlighted as follows:
\begin{itemize}
    \item This work models both RGB image and RGB dataset as Gaussian distributions. This approach can handle inter-city data heterogeneity, because it can, on the one hand, quantify the importance of each RGB image rather than typically treating all RGB images equally; on the other hand, characterize statistical properties to measure data heterogeneity on top of solely considering data volume. 
    \item We measure the data heterogeneity (\ie, computing vehicle-edge distance and edge-cloud distance) based on the established Gaussian distributions, which subsequently are utilized to calculate aggregation weight. This design accelerates convergence by preferentially integrating RGB images that exhibit greater similarity, and subsequently facilitates faster learning and better model's performance.
    \item To conserve communication resource, we propose AdapRS, a performance-aware adaptive optimization method that reduces communication resource consumption yet without sacrificing model performance. Notably, AdapRS is designed for FedGau but remains applicable to other existing HFL aggregation algorithms.
    \item Extensive experiments and empirical analyses are conducted on FedGau and AdapRS on TriSU task. The experimental results suggest that FedGau converges faster than the existing state-of-the-art (SOTA) benchmarks by 38.7\%, 37.5\%, 35.5\%, and 40.6\% in terms of mIoU, mPrecision, mRecall, and mF1, respectively. On the other hand, extensive experiments also demonstrate that AdapRS can save communication resource by 29.65\% compared to StatRS.
\end{itemize}

The remainder of this paper is organized as follows. \Cref{related_work} provides an overview of the related work. \Cref{methodology} elaborates on the proposed FedGau and AdapRS. \Cref{experiments} presents a comprehensive set of experiments along with an empirical analysis. \Cref{conclusion} concludes this paper.

\section{Related Work}
\label{related_work}
\subsection{Hierarchical Federated Learning (HFL)}
FL is a decentralized and distributed machine learning paradigm that prioritizes data privacy preservation \cite{dong2022federated,han2022fedx,ye2024prafflpreferenceawareschemefair}. 
For the initial FedAvg \cite{https://doi.org/10.48550/arxiv.1602.05629}, it aggregates vehicles' model parameters through weighted averaging at the server. However, recent studies \cite{wang2021addressing,huang2021personalized} have found that data heterogeneity can negatively slow down convergence.
To address this issue, several strategies have been proposed \cite{li2020federated,acar2021federated}. For example, 
FedProx \cite{li2020federated} introduces a proximal regularization term on local models, ensuring that the updated local parameters remain close to the global model and preventing gradient divergence. 
FedDyn \cite{acar2021federated} uses a dynamic regularizer for each device to align global and local objectives.
However, these existing methods often underperforms in complex tasks, such as object detection and semantic segmentation \cite{miao2023fedseg}. In addition, although some existing works consider data heterogeneity from statistical prospective to accelerate the FL model convergence, they generally sacrifice data privacy because of involving transfer of raw data, histogram, etc\cite{10.1145/3594779, kou2023communication}.

In the early stage of FL research, the majority of studies are cloud-based FL (CFL). In CFL, the participating clients in total can reach millions\cite{DBLP:journals/corr/abs-1902-01046}, providing massive datasets and then leading to good generalization, but suffering from high communication cost and large latency\cite{DBLP:journals/corr/abs-2010-11612, DBLP:journals/corr/abs-1902-01046}. Recently, researchers have begun exploring edge-based FL (EFL)\cite{8664630,wang2022collaborative}, where a nearby edge server instead of cloud server acts as the parameter server. In EFL, the communication is faster and cheaper. Nevertheless, one disadvantage of edge-based FL is the limited number of clients each server can access, leading to inevitable training performance decline.
HFL could combine strengths of both CFL and EFL. Compared with CFL, HFL will significantly reduce the costly communication with the cloud, resulting a significant reduction in the runtime and cost. Meanwhile, HFL will outperform EFL in generalization as more accessible data by the cloud server. 
In this paper, we present FedGau framework to overcome inter-city domain-shift and accelerates HFL model convergence on complex TriSU task, considering data statistics yet involving almost zero privacy leakage.

\subsection{Communication Resource Scheduling}
In reality, communication resource is limited. Facing stringent communication resource, an obvious question is how to allocate the resource to edge aggregation level and cloud aggregation level in HFL, aiming at trading off communication resource between aforementioned levels and avoiding communication resource waste. Allocating most of the traffic to edge aggregation will result in insufficient rounds of global model aggregation at the cloud and severe model discrepancy among edge servers. Allocating most of the traffic to cloud aggregation will not necessarily lead to performance improvement as the cloud is repeatably aggregating models that do not sufficiently updated by the edge servers. 
In a nutshell, allocating the communication resource to different HFL stages is a tricky problem as it affects the performance and convergence of HFL model. Therefore, optimizing the communication resource at different HFL stages need careful consideration. In this aspect, \cite{DBLP:journals/corr/abs-2010-11612,9834296,yang2022hierarchical} proposed to optimize the aggregation interval, including cloud aggregation interval (CAI) and edge aggregation interval (EAI). However, all these works consider StatRS, where CAI and EAI remains unchanged across $rounds$. As mentioned earlier, StatRS can lead to a significant waste in communication resources\cite{8664630}. 
In order to mitigate the above communication resource waste, this paper presents a performance-aware AdapRS policy to reduce communication resource waste while maintaining as good performance as StatRS.

\subsection{\underline{S}treet \underline{S}cene \underline{S}emantic \underline{U}nderstanding (TriSU)}
TriSU is a field within AD and robotics focused on enabling machines to interpret and understand the content of street scenes, typically through various forms of sensory data such as images and point clouds. This capability is crucial for AD \cite{meletis2022holistic}. 
TriSU assigns a class label to every pixel in an image. This process is crucial for understanding the layout of the street scene, including the road, pedestrian, sidewalks, buildings, and other static and dynamic elements.
Modern TriSU heavily relies on machine learning, particularly deep learning techniques. Initially, Fully Convolutional Networks (FCNs)-based models significantly improve the performance of this task \cite{yang2022deaot,zhou2022rethinking,yu2020bisenet}. In recent years, Transformer-based approaches \cite{xie2021segformer,cheng2021maskformer} have also been proposed for semantic segmentation. 
Recently, Bird's Eye View (BEV) \cite{can2022understanding} technique is widely adopted for road scene understanding \cite{9913352}. 

\section{Methodology}
\label{methodology}
\subsection{HFL Formulation}
The key notations in HFL formulation are summarized in \Cref{tab:HFRS}. We consider a HFL consisting of a cloud server, $\mathcal{|M|}$ edge servers and $ \sum_{e=1}^{\mathcal{|M|}} |\mathcal{C}_e|$ vehicles. Vehicle $\{c,e\}$ denotes the $c$-th vehicle connected to edge 
server $e$, where $c = 1, 2, \cdots, |\mathcal{C}_e|$. Vehicle $\{c,e\}$ has a local dataset $\mathcal{D}_{c,e}$ with the size of $|\mathcal{D}_{c,e}|$. The edge server $e$ virtually covers dataset $\mathcal{D}_e \triangleq \cup_{c=1}^{|\mathcal{C}_e|} \mathcal{D}_{c,e}$ with the size of $|\mathcal{D}_e|$. In the same way, the cloud server virtually covers dataset $\mathcal{D} \triangleq \cup_{e=1}^{|\mathcal{M}|} \mathcal{D}_{e}$ with the size of $|\mathcal{D}|$.

\begin{table}[tp]
    \centering
    \renewcommand{\arraystretch}{1.0}
    \setlength{\tabcolsep}{15.0pt}
    \caption{Key Notations of HFL Formulation}
    \begin{tabularx}{\linewidth}{ll}
    \hline
        \textbf{Symbols} & \textbf{Definitions} \\ \hline
        $\mathop{e}$ & Edge server ID \\
        $\mathop{\{c,e\}}$ & Vehicle ID \\
        $\mathcal{C}_e$ & Vehicle set connected to edge server $\mathop{e}$ \\ 
        $\mathcal{M}$ & Edge server set \\
        $\mathcal{D}_{c,e}$ & Training dataset on vehicle $\mathop{\{c,e\}}$ \\  
        $\mathcal{D}_{e}$ & Training dataset virtually covered by edge server $\mathop{e}$ \\  
        $\mathcal{D}$ & Entire training dataset covered by the cloud server \\ 
        $\mathcal{\omega}_{c,e}$ & Model parameters on vehicle $\mathop{\{c,e\}}$ \\  
        $\mathcal{\omega}_{e}$ & Aggregated model parameters on edge server $\mathop{e}$ \\  
        $\mathcal{\omega}$ & Global aggregated model parameters on cloud server \\  
        $p_{c,e}$ & Aggregation weight for $\mathcal{\omega}_{c,e}$ \\  
        $p_{e}$ & Aggregation weight for $\mathcal{\omega}_{e}$\\  
        $\tau_1$ & Edge aggregation interval (EAI) \\ 
        $\tau_2$ & Cloud aggregation interval (CAI) \\
        $\mathcal{K}$ & Total number of edge aggregation \\ 
        $\mathcal{R}$ & Total number of cloud aggregation \\
        \hline
    \end{tabularx}
\label{tab:HFRS}
\vspace{-0.5cm}
\end{table}

\subsubsection{Vehicle Update}
In each local iteration, vehicle $\{c,e\}$ trains its local model $\mathbf{\omega}_{c,e}$ based on the onboard dataset $\mathcal{D}_{c,e}$. We define the loss function of the $j$-th mini-batch images $\mathcal{D}_{c,e}^{(j)}$ as $\mathcal{E}(\mathbf{\omega}_{c,e},\mathcal {D}_{c,e}^{(j)})$.
The goal of training $\mathcal{\omega}_{c,e}$ on vehicle $\{c,e\}$ is to minimize its expected loss as follow:
\begin{equation}
\label{equa:robot}
\mathop{\mathrm{min}}_{\mathbf{\omega}_{c,e}}~\mathcal{L}_{c,e}(\mathcal{\omega}_{c,e}) =
\frac{1}{|\mathcal{D}_{c,e}|}
\sum_{\mathcal{D}_{c,e}^{(j)}\in\mathcal{D}_{c,e}
}\mathcal{E}(\mathbf{\omega}_{c,e},\mathcal{D}_{c,e}^{(j)}).
\end{equation}

\subsubsection{Edge Aggregation}
For edge server $e \in \mathcal{M}$, it aggregates all connected vehicles' model in a weighted summation manner to get the aggregated model $\omega_e$. This operation runs once per $\tau_1$ vehicles' training iterations. Specifically, when the training iteration of vehicle $\{c,e\}$ is equal to $k\tau_1$, $k=1,2,\cdots, \mathcal{K}$, edge server $e$ collects all connected vehicles' models and then performs edge aggregation in a weighted summation fashion, i.e.,
\begin{align}
    \omega_e &= \sum_{c=1}^{|\mathcal{C}_e|} p_{c,e}\omega_{c,e},
\end{align}
where $p_{c,e}$ is the aggregation weight of vehicle $\{c,e\}$ in edge aggregation. $\mathcal{C}_e$ is the set of the vehicles connected to edge server $e$. Once the model $\omega_e$ is obtained, it will be sent back to all vehicles that are connected to edge server $e$.

\subsubsection{Cloud Aggregation}
For cloud aggregation, the cloud server aggregates all edge servers' models  in a weighted summation manner to get the global model $\omega$. This operation runs once every $\tau_2$ edge aggregations. Specifically, when the training iteration of vehicle $\{c,e\}$ is equal to $r\tau_1\tau_2$, $r=1,2,\cdots, \mathcal{R}$, the cloud server receives models from all edge servers and then performs cloud aggregation, i.e.,
\begin{align}
    \omega &= \sum_{e=1}^{|\mathcal{M}|} p_e\omega_{e} = \sum_{e=1}^{|\mathcal{M}|} p_e\sum_{c=1}^{|\mathcal{C}_e|} p_{c,e}\omega_{c,e}, 
    \label{eq:global_sum}
\end{align}
where $p_{e}$ is the aggregation weight of the model of edge server $e$ in cloud aggregation. After cloud aggregation, the cloud server will redistribute the aggregated model $\omega$ to all edge servers and further to all vehicles via the corresponding edge server.

\subsection{FedGau}
\label{fedgau}
In this section, we will present the proposed FedGau. The underlying mathematical foundation of FedGau stems from FL convergence analysis \cite{wang2023batch}. This analysis demonstrates that under inter-city setting (characterized by non-i.i.d. data), discrepancy of BatchNorm statistical parameters between vehicle (or edge server) and the corresponding edge server (or cloud server) leads to gradient divergence, and consequently impedes and biases the FL convergence. Wang \etal~ \cite{wang2023batch} also notes that such impeded and biased convergence can be attributed to the statistical discrepancy between vehicles' dataset (or edge servers' dataset) and the corresponding edge server's dataset (or cloud server's dataset), particularly in a non-i.i.d. setting. However, the conventional aggregation weights $p_{c,e}$ and $p_e$ in \Cref{eq:global_sum}, typically defined as follows,
\begin{align}
    p_{c,e} = \frac{|\mathcal{D}_{c,e}|}{|\mathcal{D}_{e}|}, ~~~
    p_e = \frac{|\mathcal{D}_{e}|}{|\mathcal{D}|}, 
\end{align}
regard each RGB image contributes equally and ignore the statistical discrepancy among involved datasets. Such weight definition inevitably fails to represent the statistical discrepancy between vehicles' dataset (or edge servers' dataset) and the corresponding edge server's dataset (or cloud server's dataset) in model aggregation.

\begin{figure*}[tp]
\centering 
\vspace{-0.3cm}
\includegraphics[width=\linewidth, height=0.6\linewidth]{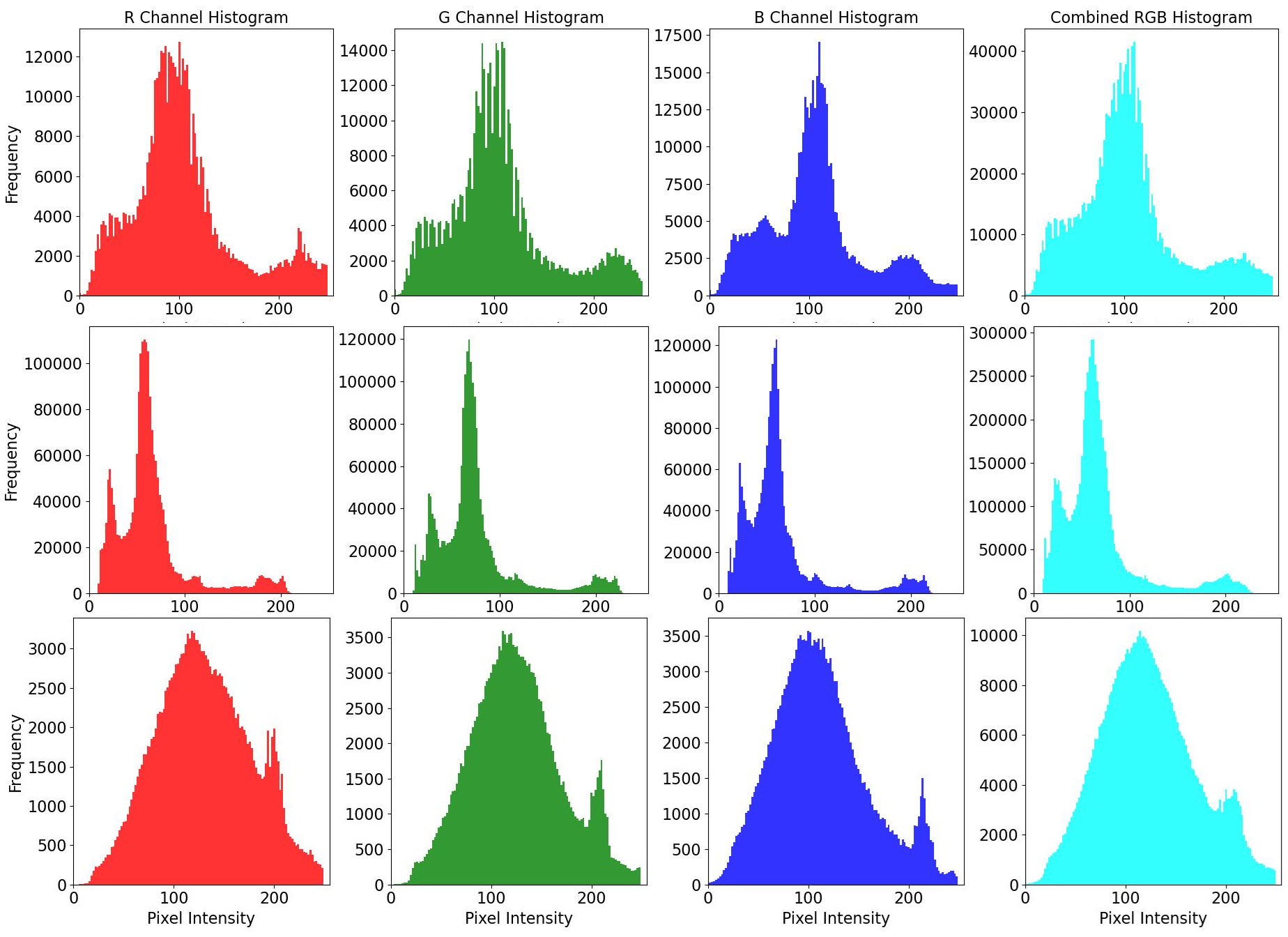}
\vspace{-0.5cm}
\caption{Histograms of pixel densities. The first raw represents histograms of one RGB image from CamVid dataset \cite{brostow2008segmentation}. The second raw represents histograms of one RGB image from Cityscapes dataset \cite{Cordts2016Cityscapes}. The last raw represents histograms of one RGB image from Internet.}
\label{Fig:gaus_examples}
\vspace{-0.3cm}
\end{figure*}

Motivated by this, we propose to measure the statistical discrepancy to accelerate HFL model convergence in inter-city setting. Our observation indicates that the distribution of pixel intensities of each RGB image (or individual channel of RGB image) displays a bell-curve shape when visualized as a histogram (exemplified in \Cref{Fig:gaus_examples}), which is a characteristic of a Gaussian distribution. Based on this observation, we propose to use Gaussian distribution as an approximation to model the statistical distribution of RGB image.

On top of above Gaussian assumption, we introduce FedGau that built on a tractable approximate quantification of the discrepancy of the statistical parameters between vehicles' dataset (or edge servers' dataset) and the corresponding edge server's dataset (or cloud server's dataset). With this tractable measure at hand, FedGau is designed to measure the discrepancy, and thus allows faster convergence. More specifically, in the proposed FedGau framework, both individual RGB image and RGB dataset are modelled as Gaussian distributions. This approach stands out by recognizing the uniqueness of each RGB image, as opposed to conventional $proportion$-based method that treats all RGB images uniformly and ignores their statistical properties. Furthermore, FedGau's modelling technique takes into account both the volume of data and its underlying statistical characteristics, offering a more comprehensive information than approaches focused solely on data quantity. Based on the estimitated distributions, we can measure the distance between vehicles' dataset (or edge servers' dataset) and the corresponding edge server's dataset (or cloud server's dataset), and then determine the contribution of each involved model in the aggregation.

\begin{figure}[tp]
\centering 
\vspace{-0.3cm}
\includegraphics[width=\linewidth]{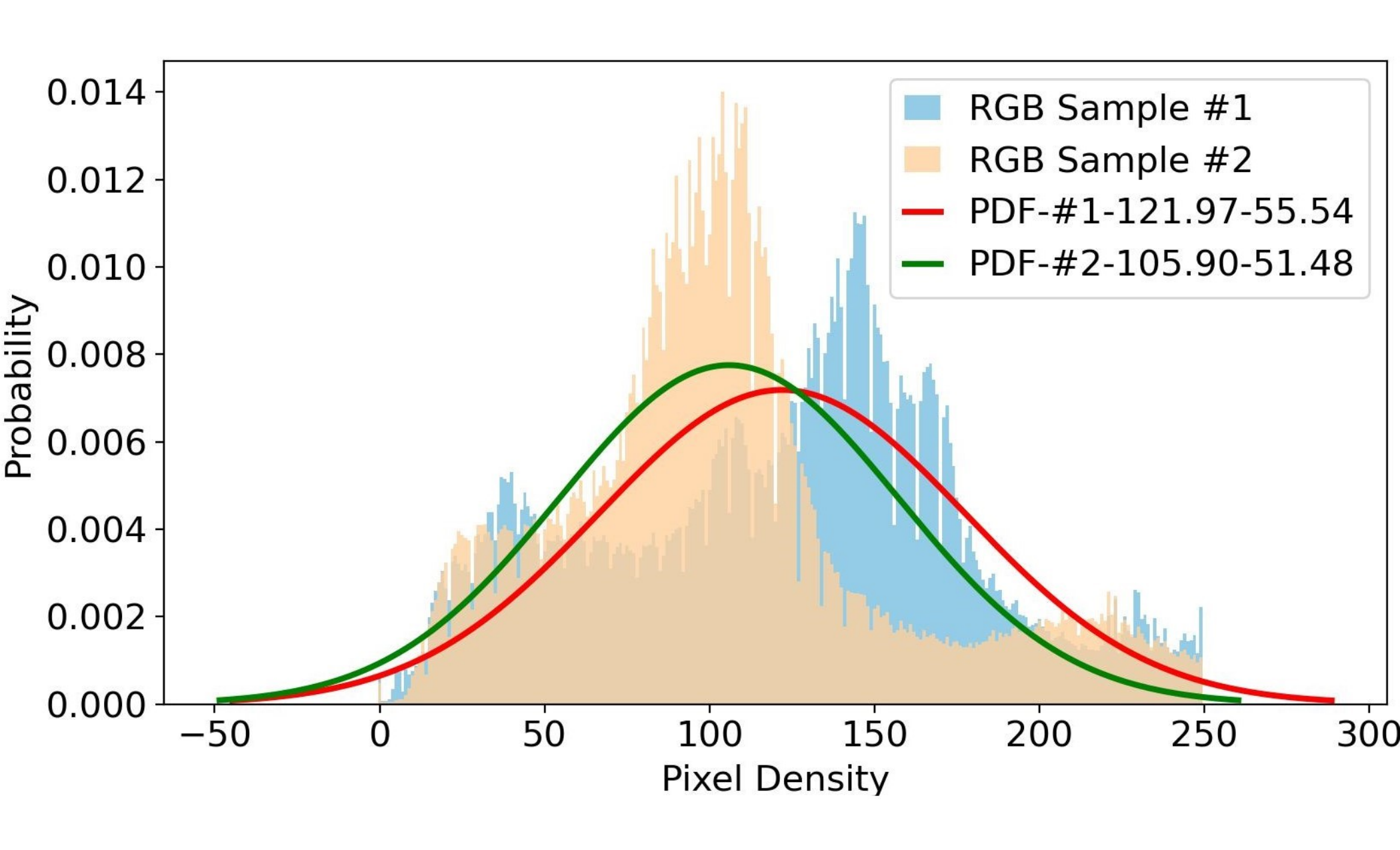}
\vspace{-0.8cm}
\caption{Illustration of two normalized histograms and the corresponding estimated probability density functions (pdfs). These two pdfs are estimated by our proposed scheme. For example, the mean and variance of ``RGB Sample \#1'' are 121.97 and 55.54.}
\label{Fig.2_rgb_comp}
\vspace{-0.5cm}
\end{figure}

We detail the proposed FedGau as following progressive steps.

\subsubsection{Step I: Distribution Estimation of Single RGB Image}
For a single RGB image with the dimension $\mathcal{W} \times \mathcal{H}$, we assume that the pixel value $\mathcal{X}$ follows a Gaussian distribution, i.e., $\mathcal{X} \sim \mathcal{N}(\mu_s, \delta_s^2)$, where $\mu_s$ and $\delta_s^2$ can be estimated by using $L = 3 \times \mathcal{W} \times \mathcal{H}$ samples (regarding R, G and B channels to follow one shared distribution) according to following equations:
\begin{align}
   \mu_s = \frac{1}{L}\sum_{l=1}^L x_l, ~~~ 
   \delta_s^2 = \frac{1}{L-1} \sum_{l=1}^L (x_l - \mu_s)^2,
   \label{eq:rgb_delta}
\end{align}
where $x_l$ means a pixel value from any one channel of the RGB image. 
\Cref{Fig.2_rgb_comp} exemplifies the estimated Gaussian distributions of two RGB images based on this method.

\subsubsection{Step II: Distribution Estimation of RGB Dataset}
For vehicle $\{c,e\}$, its onboard RGB dataset $\mathcal{D}_{c,e}$ contains $n_{c,e} = |\mathcal{D}_{c,e}|$ RGB images. Based on \textit{Step I}, the $i$-th ($1 \leq i \leq n_{c,e}$) image in $\mathcal{D}_{c,e}$ is modelled as a Gaussian distribution, i.e., $\mathcal{X}_i \sim \mathcal{N}(\mu_i, \delta_i^2)$. Based on this, we can estimate the statistical distribution of the vehicle dataset $\mathcal{D}_{c,e}$ as the average of the Gaussian distributions of all included RGB images, i.e., $\mathcal{X}_{c,e} = 1 / n_{c,e} \sum_{i=1}^{n_{c,e}} \mathcal{X}_i$, which also follows a Gaussian distribution, i.e., $\mathcal{X}_{c,e} \sim \mathcal{N}(\mu_{c,e}, \delta_{c,e}^2)$, where the parameters $\mu_{c,e}$ and $\delta_{c,e}^2$ can be calculated as 
\begin{align}
    \mu_{c,e} = \frac{1}{n_{c,e}} \sum_{i=1}^{n_{c,e}} \mu_i, ~~~ 
    \delta_{c,e}^2 = \frac{1}{n_{c,e}^2} \sum_{i=1}^{n_{c,e}} \delta_i^2.
\label{eq:dataset_delta}
\end{align}
Taking the dataset size $n_{c,e}$ into consideration, we can use a three-element tuple $(n_{c,e}, \mu_{c,e}, \delta_{c,e}^2)$ to represent $\mathcal{D}_{c,e}$.

For the edge server $e$, it receives the information $(n_{c,e}, \mu_{c,e}, \delta_{c,e}^2)$ from all connected vehicles. Then, edge server $e$ can calculate the parameters of the Gaussian distribution of $\mathcal{D}_e$ by averaging the Gaussian distributions of all covered RGB images, which still is a Gaussian distribution. The Gaussian distribution parameters of $\mathcal{D}_e$ can be calculated by
\begin{align}
    n_{e} = \sum_{c=1}^{|\mathcal{C}_e|} n_{c,e},~ 
    \mu_{e} = \frac{1}{n_{e}} \sum_{c=1}^{|\mathcal{C}_e|} n_{c,e}\mu_{c,e}, ~
    \delta_{e}^2 = \frac{1}{n_{e}^2} \sum_{c=1}^{|\mathcal{C}_e|} n_{c,e}^2\delta_{c,e}^2.
    \label{eq:edge_delta}
\end{align}
Therefore, we can use $(n_{e}, \mu_{e}, \delta_{e}^2)$ to represent the dataset $\mathcal{D}_{e}$.

For cloud server, it receives the information 
$(n_e, \mu_{e}, \delta_{e}^2)$ from all edge servers, and then can calculate the Gaussian distribution parameters of dataset $\mathcal{D}$ by averaging the Gaussian distributions of all covered RGB images, which is a Gaussian distribution as well. The Gaussian distribution parameters of dataset $\mathcal{D}$ based on following equations:
\begin{align}
    n = \sum_{e=1}^{|\mathcal{M}|} n_{e}, ~~ 
    \mu = \frac{1}{n} \sum_{e=1}^{|\mathcal{M}|} n_{e}\mu_{e}, ~~
    \delta^2 = \frac{1}{n^2} \sum_{e=1}^{|\mathcal{M}|} n_{e}^2\delta_{e}^2.
    \label{eq:cloud_delta}
\end{align}
Therefore, we can use $(n, \mu, \delta^2)$ to represent dataset $\mathcal{D}$.

\subsubsection{Step III: Distance Calculation between RGB Datasets}
As discussed in \textit{Step II}, Gaussian distributions of data at vehicles, edge servers, and cloud server are obtained. Therefore, we can estimate the distance between vehicle (or edge server) and the corresponding edge server (or cloud server) by calculating the distance between the corresponding Gaussian distributions. Specifically, we propose to utilize Bhattacharyya distance \cite{bhattacharyya1943measure} to quantify the distance between such Gaussian distributions. The main advantage of Bhattacharyya distance is that it can be more robust to measure the distance between distributions, as it takes into account the entire distribution, rather than only focusing on the statistic parameters of distribution.
Another advantage is that the Bhattacharyya distance is a symmetric measure, meaning that the distance between P and Q is the same as the distance between Q and P. This property is not shared by KL divergence \cite{kullback1951information}. In addition, Bhattacharyya distance satisfies non-negativity and the triangle inequality. This makes it a useful tool for a wide range of data analysis tasks.

Given two Gaussian distributions $\mathcal{D}_1 \sim \mathcal{N}(\mu_{D_1}, \delta_{D_1}^2)$ and $\mathcal{D}_2 \sim \mathcal{N}(\mu_{D_2}, \delta_{D_2}^2)$ of two datasets, we start with the definition of the Bhattacharyya coefficient $\sigma(\mathcal{D}_1, \mathcal{D}_2)$ for two distributions as follow:
\begin{equation}
\sigma(\mathcal{D}_1, \mathcal{D}_2) = \int \sqrt{f_1(x) f_2(x)} \, dx,
\label{eq:bc_def}
\end{equation}
where $f_1(x)$ and $f_2(x)$ are the probability density function (pdf) of $\mathcal{D}_1$ and $\mathcal{D}_2$, respectively. Bhattacharyya coefficient measures the degree of overlap between two pdfs. For Gaussian distributions, their pdfs are given by
\begin{equation}
f_i(x) = \frac{1}{\sqrt{2\pi}\delta_{D_i}} \exp\left(-\frac{(x - \mu_{D_i})^2}{2\delta_{D_i}^2}\right), \quad i=1,2.
\label{eq:pdfs}
\end{equation}
Substituting \Cref{eq:pdfs} into \Cref{eq:bc_def}, we can calculate the Bhattacharyya coefficient of two Gaussian distributions as
\begin{equation}
\hspace{-0.3cm}\sigma(\mathcal{D}_1, \mathcal{D}_2)\hspace{-0.1cm} = \hspace{-0.1cm}\exp\hspace{-0.1cm}\left(\hspace{-0.1cm}-\frac{1}{4} \frac{(\mu_{D_1}\hspace{-0.1cm} -\hspace{-0.1cm} \mu_{D_2})^2}{\delta_{D_1}^2\hspace{-0.1cm} +\hspace{-0.1cm} \delta_{D_2}^2}\hspace{-0.1cm} -\hspace{-0.1cm}  \frac{1}{2} \ln\hspace{-0.1cm}\left(\hspace{-0.1cm}\frac{\delta_{D_1}^2 \hspace{-0.1cm}+\hspace{-0.1cm} \delta_{D_2}^2}{2 \delta_{D_1} \delta_{D_2}}\hspace{-0.1cm}\right)\hspace{-0.1cm}\right).
\label{eq:BC}
\end{equation}
Based on this, the Bhattacharyya distance is then given by
\begin{equation}
D_B(\mathcal{D}_1, \mathcal{D}_2) = -\ln(\sigma(\mathcal{D}_1, \mathcal{D}_2)).
\label{eq:BD}
\end{equation}
Then substituting \Cref{eq:BC} into \Cref{eq:BD}, the Bhattacharyya distance can be formulated as follow:
\begin{equation}
\hspace{-0.3cm}D_B(\mathcal{D}_1, \mathcal{D}_2)\hspace{-0.1cm} = \hspace{-0.1cm}\frac{1}{4} \frac{(\mu_{D_1}\hspace{-0.1cm} -\hspace{-0.1cm} \mu_{D_2})^2}{\delta_{D_1}^2\hspace{-0.1cm} +\hspace{-0.1cm} \delta_{D_2}^2}\hspace{-0.1cm} +\hspace{-0.1cm}  \frac{1}{2} \ln\hspace{-0.1cm}\left(\hspace{-0.1cm}\frac{\delta_{D_1}^2 \hspace{-0.1cm}+\hspace{-0.1cm} \delta_{D_2}^2}{2 \delta_{D_1} \delta_{D_2}}\hspace{-0.1cm}\right)\hspace{-0.1cm},
\label{eq:bd}
\end{equation}
where the first term measures the squared difference in means and reflects the separation between the two distributions, the second term accounts for the difference in the spread of the two distributions.

On top of the calculation of the Bhattacharyya distance in \Cref{eq:bd}, we can calculate the distance between vehicle $\{c,e\}$ and edge server $e$ by $D_B(\mathcal{D}_{c,e}, \mathcal{D}_e)$, and the distance between edge server $e$ and cloud server by $D_B(\mathcal{D}_{e}, \mathcal{D})$.

\subsubsection{Step IV: FedGau Weights Calculation}
Based on distances $D_B(\mathcal{D}_{c,e}, \mathcal{D}_e)$ and $D_B(\mathcal{D}_{e}, \mathcal{D})$ in \textit{Step III}, the aggregation weights $p_{c,e}$ and $p_e$ can be calculated by assigning a normalized weight based on how similar the distribution of vehicle $\{c,e\}$ (or edge server $e$) is to the distribution of edge server $e$ (or cloud server), and they can be formulated as follows:
\begin{align}
    p_{c,e} = \frac{\frac{1}{D_B(\mathcal{D}_{c,e}, \mathcal{D}_e)}}{\sum_{c} \frac{1}{D_B(\mathcal{D}_{c,e}, \mathcal{D}_e)}}, ~~~
    p_{e} = \frac{\frac{1}{D_B(\mathcal{D}_{e}, \mathcal{D})}}{\sum_{e} \frac{1}{D_B(\mathcal{D}_{e}, \mathcal{D})}}, 
\label{eq:coef}
\end{align}
which implies that closer distance yields higher aggregation weight. When compared with $proportion$-based weights that equalize all RGB images, the proposed FedGau's weights can leverage tailored Gaussian distribution of each RGB image to accelerate HFL convergence on TriSU task.

\begin{figure*}[h]
\centering
\subfloat[Edge1]{\includegraphics[width=0.24\linewidth]{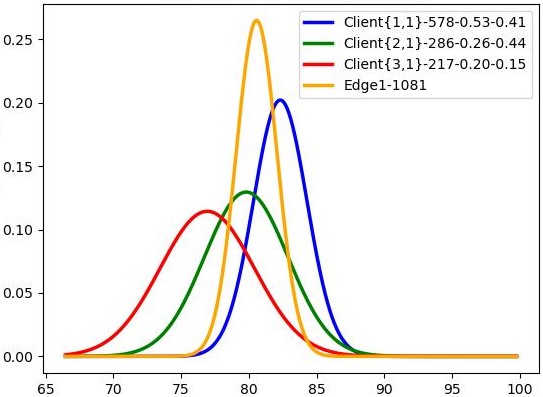}
\label{Fig.city_edge1}
}
\subfloat[Edge2]{\includegraphics[width=0.24\linewidth]{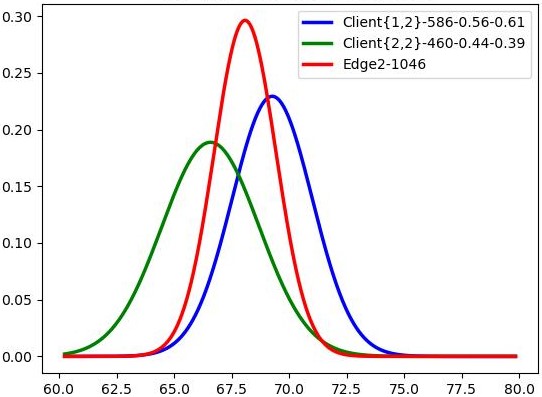}
\label{Fig.city_edge2}
}
\subfloat[Edge3]{\includegraphics[width=0.24\linewidth]{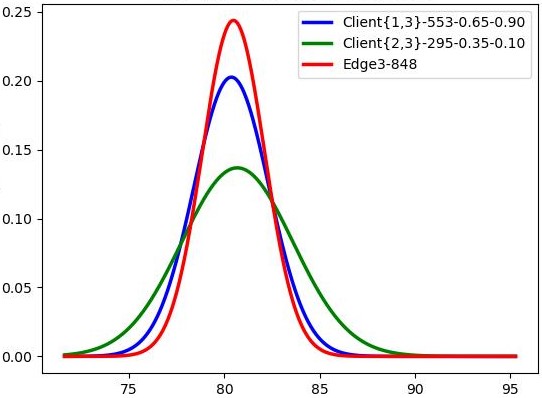}
\label{Fig.city_edge3}
}
\subfloat[Cloud]{\includegraphics[width=0.24\linewidth]{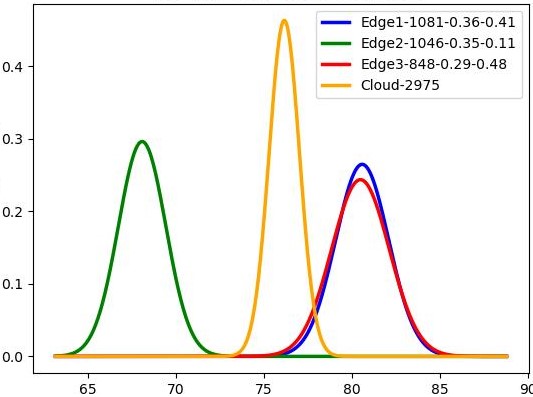}
\label{Fig.city_cloud}
}
\caption{Illustration of FedGau on Cityscapes Dataset. The figure legend $'Client\{1,1\}-578-0.53-0.41'$ in \Cref{Fig.city_edge1}, can be separated into four parts by $'-'$. They represent vehicle ID, vehicle dataset size, $proportation$-based weight and FedGau weight, respectively. The legend $'Edge1-1081'$ means Edge $1$ has virtual dataset with 1081 size. The legends in \Cref{Fig.city_edge2} and \Cref{Fig.city_edge3} share the similar meaning with \Cref{Fig.city_edge1}. The legend $'Edge1-1081-0.36-0.42'$ in \Cref{Fig.city_cloud} also can be separated into four parts by $'-'$. They represent edge ID, edge dataset size, $proportion$-based weight and FedGau weight, respectively. It is observed that FedGau weights are better than $proportion$-based weights for model aggregation. For example, in the \Cref{Fig.city_cloud}, the distribution of edge server 2 is far away from the distribution of cloud server, therefore, it should have a smaller aggregation weight, where FedGau weight (\ie, 0.11) fits well whereas $proportion$-based weight (\ie, 0.35) does not.}
\label{Fig.dataset_dist1}
\vspace{-0.5cm}
\end{figure*}

\begin{figure*}[h]
\centering
\subfloat[Edge1]{\includegraphics[width=0.24\linewidth]{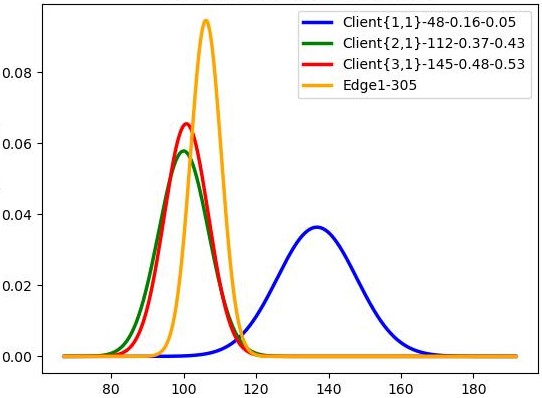}
\label{Fig.cam_edge1}
}
\subfloat[Edge2]{\includegraphics[width=0.24\linewidth]{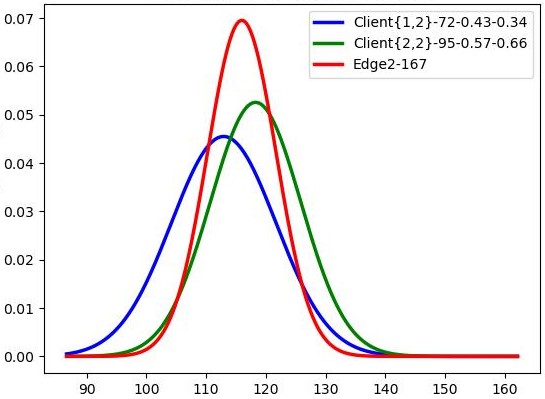}
\label{Fig.cam_edge2}
}
\subfloat[Edge3]{\includegraphics[width=0.24\linewidth]{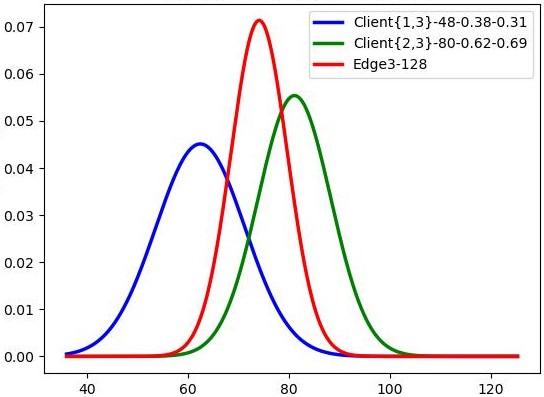}
\label{Fig.cam_edge3}
}
\subfloat[Cloud]{\includegraphics[width=0.24\linewidth]{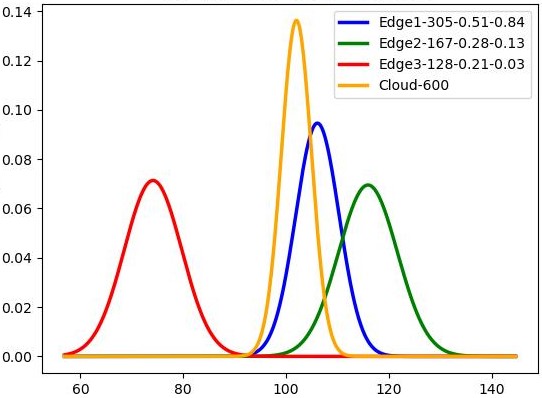}
\label{Fig.cam_cloud}
}
\caption{Illustration of FedGau on CamVid Dataset. This figure shares the similar meaning with \Cref{Fig.dataset_dist1}.}
\label{Fig.dataset_dist2}
\vspace{-0.5cm}
\end{figure*}

We summarize FedGau in \Cref{alg:HierFedGau} (overall framework) and \Cref{alg:FedGau} (basic operation unit). In \Cref{alg:FedGau}, \textbf{Server} can be either cloud server or edge server. When \textbf{Server} represents the cloud server, \textbf{NS} means the set of all edge servers, while when \textbf{Server} represents the edge server, \textbf{NS} means the set of all connected vehicles. Furthermore, we visualize FedGau's result of Cityscapes dataset in \Cref{Fig.dataset_dist1} and FedGau's result of CamVid dataset in \Cref{Fig.dataset_dist2}. 

\begin{algorithm}[tp]
\SetAlgoLined
\caption{FedGau}
\KwIn{Cloud server: \textbf{Cloud}, Edge set: $\mathbf{\mathcal{M}}$, Vehicle set: $\cup_{e=1}^{|\mathcal{M}|}\mathcal{C}_e$}
\KwOut{Aggregation Weights: $\mathcal{P}$}

\SetKwFunction{FedGauBase}{FedGau\_Base}

\textbf{Edge Server Side:} \\
\ForEach{$\mathbf{edge\ server}\ e \in \mathbf{\mathcal{M}}$}{
    \FedGauBase{$\mathbf{edge\ server}\ e, \mathcal{C}_e$}
}

\textbf{Cloud Server Side:} \\
\FedGauBase{$\mathbf{Cloud}, \mathcal{M}$} \tcp{Algo. 2}
\label{alg:HierFedGau}
\end{algorithm}

\setlength{\textfloatsep}{8pt}
\begin{algorithm}[tp]
\SetAlgoLined
\caption{FedGau\_Base}
\KwIn{One server: $\mathbf{Server}$, Connected node set: $\mathbf{NS}$}
\KwOut{Aggregation Weights: $\mathcal{P}$}

\textbf{Node Side:} \\
\ForEach{$\mathbf{node}\ \mathcal{S} \in \mathbf{NS}$}{
    $n_\mathcal{S}, \mu_\mathcal{S}, \delta^2_\mathcal{S} \leftarrow \Cref{eq:dataset_delta}\ \text{or}\ \Cref{eq:edge_delta}$ \\
    Send $n_\mathcal{S}, \mu_\mathcal{S}, \delta^2_\mathcal{S}$ to Server
}

\textbf{Server Side:} \\
\If{$\mathbf{Server} ==$ cloud server}{
    $n, \mu, \delta^2 \leftarrow \Cref{eq:cloud_delta}$
}
\ElseIf{$\mathbf{Server} ==$ edge server}{
    $n, \mu, \delta^2 \leftarrow \Cref{eq:edge_delta}$
}

\ForEach{$\mathbf{node}\ \mathcal{S} \in \mathbf{NS}$}{
    $\mathcal{P}_{\mathcal{S}} = D_B((n_\mathcal{S}, \mu_{\mathcal{S}}, \delta^2_{\mathcal{S}}), (n, \mu, \delta^2))$
}
\label{alg:FedGau}
\end{algorithm}

\subsection{AdapRS}
\label{adaprs}
For the considered HFL system, the amount of communication resource consumption in each $round$ (denoted as $\mathcal{T}_{crc}$) is equal to the product of the model size and the number of model exchanges (including vehicle-edge exchanges and edge-cloud exchange) in each $round$, \ie,
\begin{align}
    \label{eq:round_traffic}
\mathcal{T}_{crc} \triangleq N_{exc} \times D_{\omega} =2(\tau_{2}\sum_{e=1}^{\mathcal{|M|}}|\mathcal{C}_e|+|\mathcal{M}|)D_{\omega}, 
\end{align}
where $D_{\omega}$ means the model size (Unit: MB) and $N_{exc}$ represents the number of model exchanges in each $round$. According to \Cref{eq:round_traffic}, we can observe that $\mathcal{T}_{crc}$ is only determined by $\tau_2$ if the HFL topology is fixed (\ie, $\mathcal{M}$ and $\mathcal{C}_e$ are fixed). Therefore, $\mathcal{T}_{crc}$ of traditional StatRS is fixed across $rounds$, which fails to adapt to the changing model performance (\eg, mIoU) and inevitably results in unnecessary communication resource usage. 

To mitigate StatRS's waste of communication resource, we propose a performance-aware adaptive communication resource scheduling policy by assigning dynamic $\tau_2$ across $rounds$. To this end, we firstly build a $round$-wise HFL convergence modelling, and subsequently propose a performance-aware adaptive optimization method based on the convergence model. 

\begin{table}[tp]
    \centering
    \renewcommand{\arraystretch}{1.0}
    \setlength{\tabcolsep}{15.0pt}
    \caption{Key Notations of HFL Convergence Modelling}
    \begin{tabularx}{\linewidth}{ll}
    \hline
        \textbf{Symbols} & \textbf{Definitions} \\ \hline
        \multirow{2}{*}{$\theta_{c,e}$} & The upper bound of gradient variance \\
        ~ & between vehicle ${\{c,e\}}$ and edge server $e$ \\
        \multirow{2}{*}{$\rho_{c,e}$} & The upper bound of Lipschitz constant variance\\
        ~ & between vehicle ${\{c,e\}}$ and edge server $e$ \\
        \multirow{2}{*}{$\beta_{c,e}$} & The upper bound of smoothness constant variance \\
        ~ & between vehicle ${\{c,e\}}$ and edge server $e$ \\
        $\theta{e}$ & The weighted sum of $\theta_{c,e}$ \\  
        $\theta$ & The weighted sum of $\theta_{e}$ \\ 
        $\rho$ & Lipschitz constant \\ 
        $\beta$ & Smoothness constant \\ 
        \hline
    \end{tabularx}
\label{tab:HFCM}
\end{table}

\subsubsection{$Round$-wise HFL Convergence Modelling}
The key notations used in HFL convergence modelling are summarized in \Cref{tab:HFCM}. 

The global loss $\mathcal{L}(\omega)$ of HFL can be formulated as the weighted summation of the training loss of all involved vehicles, \ie, 
\begin{align}
    \mathcal{L}(\omega) &= \sum_{e=1}^{|\mathcal{M}|} p_e\sum_{c=1}^{|\mathcal{C}_e|} p_{c,e}\mathcal{L}_{c,e}(\omega_{c, e}).
\end{align}
According to \cite{liu2019clientedgecloud, yang2022hierarchical, 8664630, DBLP:journals/corr/abs-2010-11612,9834296}, some assumptions about $\mathcal{L}(\omega)$ are made, such as convex, $\rho$-Lipschitz, $\beta$-smoothness and bounded gradient between vehicle $\{c, e\}$ and edge server $e$. On top of these assumptions, we can model HFL convergence as the gap between each $round$'s $\mathcal{L}(\omega)$ and $C^*$ =  $\mathcal{L}(\mathbf{\omega^*})$ ($\mathbf{\omega^*}$ is the optimal model) at the end of each $round$. Specifically, for the $r$-th $round$, the HFL model convergence can be modelled as
\begin{align}
\label{equa:convergence}
\mathcal{L}(\mathbf{\omega}_r)-C^*\leq &
    \frac{C_r}{\tau_{1, r}\tau_{2, r}}+\rho_r p(\tau_{1, r}, \tau_{2, r}, \theta_{e, r}, \theta_r)\ +\ \nonumber \\
    &\sqrt{\frac{C_r^2}{\tau_{1, r}^2\tau_{2, r}^2}\!+\!\frac{2C_r\rho_r p(\tau_{1, r}, \tau_{2, r}, \theta_{e, r}, \theta_r)}{\tau_{1, r}\tau_{2, r}}}. 
\end{align}
where $r = 2,\ 3, \cdots$. Here one $round$ contains one cloud aggregation, $\tau_{2, r}$ edge aggregations, and $\tau_{1, r}\tau_{2, r}$ vehicles' training iterations. The relevant parameters within \Cref{equa:convergence} are formulated as follows:
\begin{align}
p(\tau_{1, r}, \tau_{2, r}, \theta_{e, r}, \theta_r) \triangleq &q_c(\tau_{1, r}\tau_{2, r}, \theta_r) \nonumber \\ 
&+ (\tau_{2, r} + 1)\sum_{e=1}^{|\mathcal{M}|} p_e q_e(\tau_{1, r}, \theta_{e, r}),
\end{align}
\begin{align}
&q_c(\tau_{1, r}\tau_{2, r}, \theta_r)\triangleq \theta_r(\frac{1}{\beta_r}(1+\eta\beta_r)^{\tau_{1, r}\tau_{2, r}}-\frac{1}{\beta_r}-\eta\tau_{1, r}\tau_{2, r}),
\\
&q_e(\tau_{1, r}, \theta_{e, r})\triangleq \theta_{e, r}(\frac{1}{\beta_{e, r}}(1+\eta\beta_{e, r})^{\tau_{1, r}}-\frac{1}{\beta_{e, r}}-\eta \tau_{1, r}), 
\\
&C_r \triangleq \frac{{||\omega_r - \omega^*||^2}}{\eta(2-\eta\beta_r)} \stackrel{Estimate}{\approx} \frac{{||\nabla \mathcal{L}(\omega_r)||^2}}{\eta\beta_r^2(2-\eta\beta_r)},
\end{align}
\begin{align}
&\rho_r \triangleq \sum_{e \in \mathcal{M}} p_e\sum_{c \in C_e} p_{c,e} \rho_{c,e,r},
\\
&\mathop{\theta_r} \triangleq \sum_{e \in \mathcal{M}} p_e \mathop{\theta_{e, r}}, 
\\
&\mathop{\theta_{e, r}} \triangleq \sum_{c \in C_e} p_{c,e} \mathop{\theta_{c, e, r}}, 
\\
&\beta_r \triangleq \sum_{e \in \mathcal{M}} p_e \beta_{e,r},
\\
&\mathop{\beta_{e, r}} \triangleq \sum_{c \in C_e} p_{c,e} \mathop{\beta_{c, e, r}}, 
\end{align}
where $\rho_{c,e,r}$ is the estimated constant of $\rho$-Lipschitz for vehicle $\{c, e\}$, $\beta_{c,e,r}$ is the estimated constant of $\beta$-smoothness for vehicle $\{c, e\}$, $\theta_{c, e, r}$ is the estimated boundary of gradient between vehicle $\{c, e\}$ and edge server $e$, at the end of $r$-th $round$. These parameters are all based on previous works but customized for our specific setting. 

\subsubsection{Performance-Aware Adaptive Optimization}
Generally, the training of the model in HFL involves multiple $rounds$ before the model converges. Based on above $round$-wise HFL convergence modelling, we propose a performance-aware adaptive optimization scheme to minimize the loss gap by dynamically adjusting $\tau_{1,r}$ and $\tau_{2,r}$ across $rounds$. Precisely, for the $r$-th $round$, the performance-aware adaptive resource scheduling optimization can be formulated as follows:
\begin{alignat}{2}
&\hspace{-0.5cm}\min_{\tau_{1,r}, \tau_{2,r}} &\quad& \frac{C_r}{\tau_{1,r}\tau_{2,r}} + \rho_r p(\tau_{1,r}, \tau_{2,r}, \theta_{e,r}, \theta_r) \nonumber \\ 
&&& + \sqrt{\frac{C_r^2}{\tau_{1,r}^2\tau_{2,r}^2} + \frac{2C_r\rho_r p(\tau_{1,r}, \tau_{2,r}, \theta_{e,r}, \theta_r)}{\tau_{1,r}\tau_{2,r}}}, \label{eq:obj} \\
&\mathrm{s.t.} 
\quad& & \tau_{1,r}\tau_{2,r}=\mathcal{I}, \label{eq:cons_fixed_iterations} \\
&&& 1 \leq \tau_{2,r} \leq \vartheta_r\tau_{1,r}, \label{eq:cons}
\end{alignat}
where $\mathcal{I}$ means the the number of vehicles' training iterations in the $r$-th $round$, $\vartheta_r$ is our proposed performance-aware factor to adapt $\tau_{1,r}$ and $\tau_{2,r}$ dynamically across $rounds$.

In the optimization, we keep $\mathcal{I}$ as a fixed value in \Cref{eq:cons_fixed_iterations}. The rationale behind keeping $\mathcal{I}$ to remain constant is to ensure that the computation resource should stay fixed in each $round$, as the computational resource required for server-side aggregation are considerably negligible compared to vehicle-side training. By doing so, we can confidently attribute any performance improvement in a $round$ to the communication resource investment. 

With respect to the proposed performance-aware factor $\vartheta_r$ in \Cref{eq:cons}, it is defined as follow:
\begin{equation}
\label{eq:adapfactor}
    \mathop{\vartheta_r} \triangleq max(0, ~\frac{QoC_r}{QoC_{max}}),
\end{equation}
where $QoC_r$ (the abbreviation of Quality of Communication) is defined as the ratio of the performance increment (e.g., $\Delta$mIoU) to the number of model exchanges (\ie, $N_{exc}$ in \Cref{eq:round_traffic}) in the $r$-th $round$, and $QoC_{max}$ means the maximum value out of the set \{$QoC_1$, $QoC_2$, $\cdots$, $QoC_r$\}. Concretely, $QoC_r$ and $QoC_{max}$ can be formulated as follows:
\begin{align}
    QoC_r &\triangleq \frac{\Delta mIoU_r}{N_{exc, r}} = \frac{mIoU_r - mIoU_{r-1}}{2(\tau_{2,r}\sum_{e=1}^{\mathcal{|M|}}|\mathcal{C}_e|+|\mathcal{M}|)},
    \label{eq:qoc_r}
    \\
    QoC_{max} &\triangleq max(QoC_1, QoC_2, \cdots, QoC_r).
    \label{eq:qoc_max}
\end{align}
As per \Cref{eq:adapfactor}, the factor $\vartheta_r$ is designated as a metric to quantify the distance between $QoC_r$ and $QoC_{max}$, and it is specifically employed to align $QoC_r$ with $QoC_{max}$ as much as possible in the proposed performance-aware adaptive optimization problem.

At the end of the $r$-th $round$, we run Scipy \cite{2020SciPy-NMeth} to solve above optimization problem. The optimization outcomes $\tau_{1,r}$ and $\tau_{2,r}$ are used to set $\tau_{1,r+1}$ and $\tau_{2,r+1}$ to adjust resource allocation for the $(r+1)$-th $round$. Specifically, as $\vartheta_r$ increases, $\tau_{2,r+1}$ also increases while $\tau_{1,r+1}$ decreases, thereby promoting more model exchanges to improve model performance. Conversely, when $\vartheta_r$ decreases, $\tau_{2,r+1}$ reduces and $\tau_{1,r+1}$ grows, it helps to conserve communication resource as per \Cref{eq:round_traffic}.

Our proposed AdapRS is summarized in \Cref{alg:AdapRS}.

\begin{algorithm}[tp]
\SetAlgoLined
\caption{AdapRS}
\KwIn{Cloud Server, Edge set: $\mathcal{M}$, Vehicle set: $\cup_{e=1}^{|\mathcal{M}|}\mathcal{C}_e$}
\KwOut{$\tau_{1,r+1}, \tau_{2,r+1}$}

\For{round $r = 1, \cdots, \mathcal{R}$}{
    \textbf{Vehicle Side:} \\
    \ForEach{$\mathbf{Vehicle} \in \cup_{e=1}^{|\mathcal{M}|}\mathcal{C}_e$}{
        $\rho_{c,e,r} \leftarrow \frac{||\mathcal{L}(\omega_{c,e,r}) - \mathcal{L}(\omega_{e,r})||}{||\omega_{c,e,r}-\omega_{e,r}||}$ \\
        $\beta_{c,e,r} \leftarrow \frac{||\nabla \mathcal{L}(\omega_{c,e,r}) - \nabla \mathcal{L}(\omega_{e,r})||}{||\omega_{c,e,r} - \omega_{e,r}||}$ \\
        $\theta_{c,e,r} \leftarrow ||\nabla \mathcal{L}(\omega_{c,e,r}) - \nabla \mathcal{L}(\omega_{e,r})||$ \\
        Send $\rho_{c,e,r}, \beta_{c,e,r}, \theta_{c,e,r}$ to $\mathbf{edge\ server}\ e$
    }
    
    \textbf{Edge Server Side:} \\
    \ForEach{$\mathbf{edge\ server}\ e \in \mathcal{M}$}{
        $\rho_{e,r}, \beta_{e,r}, \theta_{e,r} \leftarrow \sum_cp_{c,e}\rho_{c,e,r}, \sum_cp_{c,e}\beta_{c,e,r}, \sum_cp_{c,e}\theta_{c,e,r}$ \\
        Send $\rho_{e,r}, \beta_{e,r}, \theta_{e,r}$ to $\mathbf{cloud\ server}$
    }
    
    \textbf{Cloud Server Side:} \\
    $\rho_{r}, \beta_{r}, \theta_{r} \leftarrow \sum_ep_e\rho_{e,r}, \sum_ep_e\beta_{e,r}, \sum_ep_e\theta_{e,r}$ \\
    $\vartheta_{r} \leftarrow \Cref{eq:adapfactor}$ \\
    $\tau_{1,r+1}, \tau_{2,r+1} \leftarrow$ Optimize Obj. \Cref{eq:obj} s.t. \Cref{eq:cons_fixed_iterations,eq:cons}
}
\label{alg:AdapRS}
\end{algorithm}

\subsection{Complexity Analysis}
In this section, we will analyze the time and space complexity of FedGau and AdapRS to understand their scalability in practical use. Recall that the considered HFL system includes one cloud server, $|\mathcal{M}|$ edge servers, and $|\mathcal{V}|$ vehicles. Additionally, the considered HFL system uses a collective dataset of $n$ RGB images, as per \Cref{eq:cloud_delta}.

\subsubsection{FedGau Complexity Analysis}
For space complexity of FedGau, $n$ RGB images require $2n$ units of space; $|\mathcal{V}|$ vehicles need $3|\mathcal{V}|$ units; $|\mathcal{M}|$ edge servers need $3|\mathcal{M}|$ units; and the cloud server requires 3 units. Therefore, the total space requirement $S_{c, \text{FedGau}}$ is given by:
\begin{align}
S_{c, FedGau} = 2n + 3|\mathcal{V}| + 3|\mathcal{M}| + 3.
\label{eq:fedgau_space_comp}
\end{align}
Under typical conditions where $n$ significantly exceeds $|\mathcal{V}|$ and $|\mathcal{M}|$ (\ie, $n \gg |\mathcal{V}|$ and $n \gg |\mathcal{M}|$), we can approximate $S_{c, \text{FedGau}}$ to be roughly $2n$. Thus, the space complexity of FedGau is O($n$).

For time complexity, assuming that the basic summation operation takes time $t_p$. Based on this assumption, the computation time for estimating Gaussian distributions are listed as follows: for $n$ RGB images, it is $6n\mathcal{W}\mathcal{H}t_p$; for all vehicles, it is $2nt_p$; for all edge servers, it is $3|\mathcal{V}|t_p$; and for the cloud server, it is $3|\mathcal{M}|t_p$. Therefore, the cumulative time $T_{c, \text{FedGau}}$ of FedGau is thus given as follow:
\begin{align}
T_{c, FedGau} = 6n\mathcal{W}\mathcal{H}t_p + 2nt_p + 3|\mathcal{V}|t_p + 3|\mathcal{M}|t_p.
\label{eq:fedgau_time_comp}
\end{align}
Considering that $n \gg |\mathcal{V}|$ and $n \gg |\mathcal{M}|$, $T_{c, FedGau}$ simplifies to:
\begin{align}
T_{c, FedGau} \approx 2n(3\mathcal{W}\mathcal{H} + 1)t_p.
\label{eq:approx_fedgau_time_comp}
\end{align}
Moreover, the term $3\mathcal{W}\mathcal{H}$ is typically much greater than 1, the computation time can be further simplified as follow:
\begin{align}
T_{c, FedGau} \approx 6n\mathcal{W}\mathcal{H}t_p.
\label{eq:approx_fedgau_time_comp_simp}
\end{align}
Given this simplification, we can conclude the time complexity of FedGau is \(\text{O}(n\mathcal{W}\mathcal{H})\).

\subsubsection{AdapRS Complexity Analysis}
To analyze clear, we establish several foundational assumptions: (I) For each vehicle, the primary operation is the calculation of $\rho_{c,e,r}$, $\beta_{c,e,r}$, and $\theta_{c,e,r}$, which requires computation time $t_v$ and $3$ memory units. (II) For each edge server, the fundamental operation is summing parameters from the vehicles, which requires summation time $t_p$ and $3$ memory units. (III) For the cloud server, the basic operation is summing parameters from all edge servers, which requires summation time $t_p$ and $3$ memory units; additionally, running Scipy requires time $t_o$ and extra $3$ memory units.

Given these assumptions, the total memory space of AdapRS $S_{c, \text{AdapRS}}$ is given as follow:
\begin{align}
    S_{c, AdapRS} &= 3|\mathcal{V}| + 3|\mathcal{M}| + 6.
    \label{eq:adaprs_space_comp}
\end{align}
Based on the fact $|\mathcal{V}| \gg |\mathcal{M}|$, thus, AdapRS's space complexity is O($|\mathcal{V}|$).

As for the time complexity, for all vehicles, the time is $3|\mathcal{V}|t_v$; for all edge servers, the total time is $3|\mathcal{V}|t_p$; for the cloud server, the total time is $3|\mathcal{M}|t_p$ plus the optimization time $t_o$. Therefore, the total time $T_{c, AdapRS}$ is formalized as follow:
\begin{align}
    T_{c, AdapRS} = (3|\mathcal{V}|(t_v + t_p) + 3|\mathcal{M}|t_p + t_o) \times r,
    \label{eq:adaprs_time_comp}
\end{align}
where $r$ refers to the number the $rounds$. Since  $|\mathcal{V}| \gg |\mathcal{M}|$, $T_{c, AdapRS}$ can be simplified further as follow:
\begin{align}
    T_{c, AdapRS} \approx 3|\mathcal{V}|(t_v + t_p) \times r.
    \label{eq:adaprs_time_comp_simplified}
\end{align}
Thus, the time complexity of AdapRS is O($|\mathcal{V}| \times r$).

\section{Experiments}
\label{experiments}
In this section, we firstly introduce the experimental settings in \Cref{datasets_metrics_imp}. Based on such settings, we will evaluate the FedGau's convergence in \Cref{fedgau_convergence} and the performance in \Cref{fedgau_perf}. In addition, we will also evaluate AdapRS in \Cref{adaprs_eval}. Finally, we conduct ablation studies in \Cref{abl_study}.

\begin{table}[tp]
    \centering
    \renewcommand{\arraystretch}{1.0}
    \setlength{\tabcolsep}{15.0pt}
    \caption{Hardware/Software configurations}
    \begin{tabularx}{\linewidth}{ll}
    \hline
        \textbf{Items} & \textbf{Configurations} \\ \hline
        CPU  & AMD Ryzen 9 3900X 12-Core \\ 
        GPU  & NVIDIA GeForce 3090 $\times$ 3, 4090 $\times$ 4\\ 
        RAM  & DDR4 32G \\ 
        DL Framework  & PyTorch @ 1.13.0+cu116 \\ 
        GPU Driver  & 470.161.03 \\ 
        CUDA  & 11.4 \\ 
        cuDNN  & 8302 \\ \hline
    \end{tabularx}
\label{Tab:configs}
\end{table}

\begin{table}[tp]
    \centering
    \renewcommand{\arraystretch}{1.0}
    \setlength{\tabcolsep}{15.0pt}
    \caption{Training configurations}
    \begin{tabularx}{\linewidth}{ll}
    \hline
        \textbf{Items} & \textbf{Configurations} \\ \hline
        Loss  & nn.CrossEntropyLoss \\ 
        Optimizer  & nn.Adam \\ 
        Adam Betas  & (0.9, 0.999) \\ 
        Weight Decay  & 1e-4 \\ 
        Batch Size  & 8 \\ 
        Learning Rate  & 3e-4 \\ 
        \multirow{1}{*}{DNN Models}  & DeepLabv3+ \cite{chen2018encoderdecoder}, BiSeNetV2 \cite{yu2021bisenet}, SegNet \cite{badrinarayanan2017segnet} \\
        \multirow{3}{*}{FL Algorithms} 
        ~ &FedProx \cite{li2020federated}, FedDyn \cite{acar2021federated}, FedAvgM \cite{hsu2019measuring} \\
        ~ &FedIR \cite{hsu2020federated}, FedCurv \cite{shoham2019overcoming},  FedNova \cite{wang2020tackling} \\
        ~ &MOON \cite{li2021model}, SCAFFOLD \cite{karimireddy2020scaffold}, FedAvg \cite{https://doi.org/10.48550/arxiv.1602.05629} \\
        \hline
    \end{tabularx}
\label{Tab:train}
\end{table}

\subsection{Datasets, Metrics and Implementation}
\label{datasets_metrics_imp}
\subsubsection{Datasets}
The \textbf{Cityscapes} dataset \cite{Cordts2016Cityscapes}, captured in multiple cities, consists of 2,975 training images, 500 validation images, and 1,525 test images, with ground truth. Cityscpaes dataset includes pixel-level label of 19 classes, including vehicles, pedestrians and so forth. The \textbf{CamVid} dataset \cite{brostow2008segmentation} totally includes 701 images with pixel-level label of 11 classes. We select 600 images randomly as training dataset and the remaining 101 images are used as test dataset. 

\subsubsection{Evaluation Metrics}
We evaluate the proposed FedGau and AdapRS on TriSU task using four metrics: \textbf{mIoU} (mean intersection-over-union between predicted masks and ground truth), \textbf{mPrecision (mPre for short)} (mean ratio of true positive pixels to total predicted positive pixels), \textbf{mRecall (mRec for short)} (mean ratio of true positive pixels to total positive pixels in ground truth), and \textbf{mF1} (mean harmonic mean of precision and recall), across all semantic classes.
These metrics are formulated as follows: 
\begin{align}
    &mIoU = \frac{1}{\mathcal{C}}\sum_{c=1}^{\mathcal{C}}IoU_c = \frac{1}{\mathcal{C} \mathcal{N}}\sum_{c=1}^{\mathcal{C}} \sum_{n=1}^{\mathcal{N}} \frac{TP_{n, c}}{FP_{n, c}\hspace{-0.1cm}+\hspace{-0.1cm} TP_{n, c}\hspace{-0.1cm} +\hspace{-0.1cm} FN_{n, c}},
    \nonumber
    \\
    \vspace{-0.2cm}
    &mPre = \frac{1}{\mathcal{C}}\sum_{c=1}^{\mathcal{C}}Pre_c = \frac{1}{\mathcal{C} \mathcal{N}}\sum_{c=1}^{\mathcal{C}} \sum_{n=1}^{\mathcal{N}} \frac{TP_{n, c}}{FP_{n, c} + TP_{n, c}},
    \nonumber
    \\
    \vspace{-0.2cm}
    &mRec = \frac{1}{\mathcal{C}}\sum_{c=1}^{\mathcal{C}}Rec_c = \frac{1}{\mathcal{C} \mathcal{N}}\sum_{c=1}^{\mathcal{C}} \sum_{n=1}^{\mathcal{N}} \frac{TP_{n, c}}{TP_{n, c} + FN_{n, c}},
    \nonumber
    \\
    \vspace{-0.2cm}
    &mF1 = \frac{1}{\mathcal{C}}\sum_{c=1}^{\mathcal{C}}F1_c = \frac{1}{\mathcal{C}}\sum_{c=1}^{\mathcal{C}} \frac{2 * Pre_c * Rec_c}{Pre_c + Rec_c},
    \label{Eq:mF1}
\end{align}
where $TP$, $FP$, $TN$ and $FN$ stand for True Positive, False Positive, True Negative and False Negative, respectively. $\mathcal{C}$ denotes the number of semantic classes within the dataset, with value of 19 for the Cityscapes dataset and 11 for the CamVid dataset. Similarly, $\mathcal{N}$ signifies the size of the test dataset, which is 500 for Cityscapes dataset and 101 for CamVid dataset.

\subsubsection{Implementation Details}
The main hardware and software configurations are listed in \Cref{Tab:configs}. The training details are summarized in \Cref{Tab:train}.

As detailed in \Cref{Tab:train}, we compare the proposed FedGau against other nine FL algorithms, where FedDyn, FedProx, and FedAvgM each includes a hyperparameter. In our experiments, we configured the hyperparameter for FedDyn to two distinct values (\ie, 0.005 and 0.01) and denoted them as FedDyn(0.005) and FedDyn(0.01). In a similar fashion, we denoted configurations of FedProx as FedProx(0.005) and FedProx(0.01), and FedAvgM as FedAvgM(0.7) and FedAvgM(0.9).

\subsection{Evaluation on FedGau's convergence}
\label{fedgau_convergence}
\begin{figure*}[tp]
\centering
\subfloat[mIoU]{\includegraphics[width=0.24\linewidth]{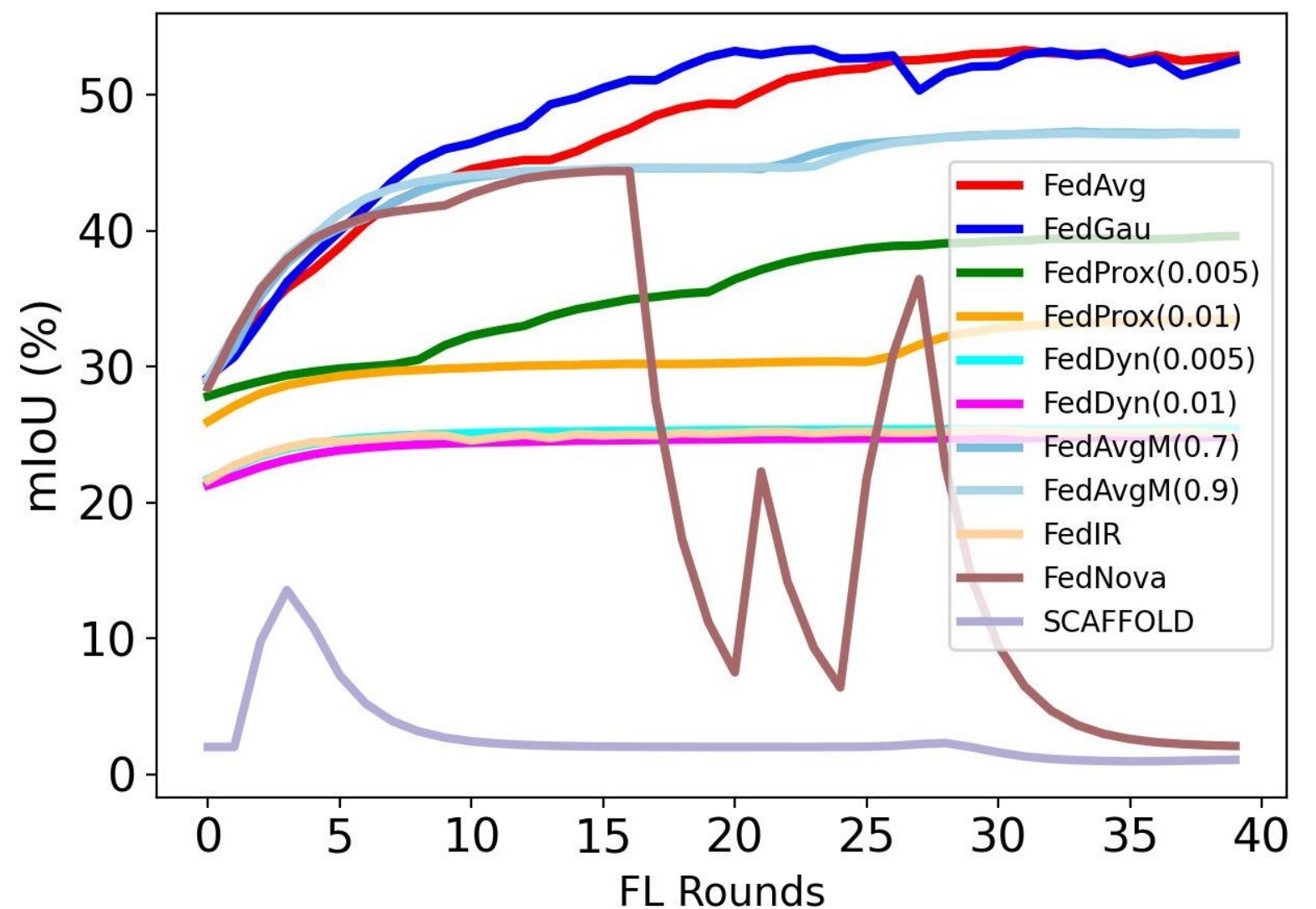}
\label{Fig.Metrics_mIoU}
}
\subfloat[mPrecision]{\includegraphics[width=0.24\linewidth]{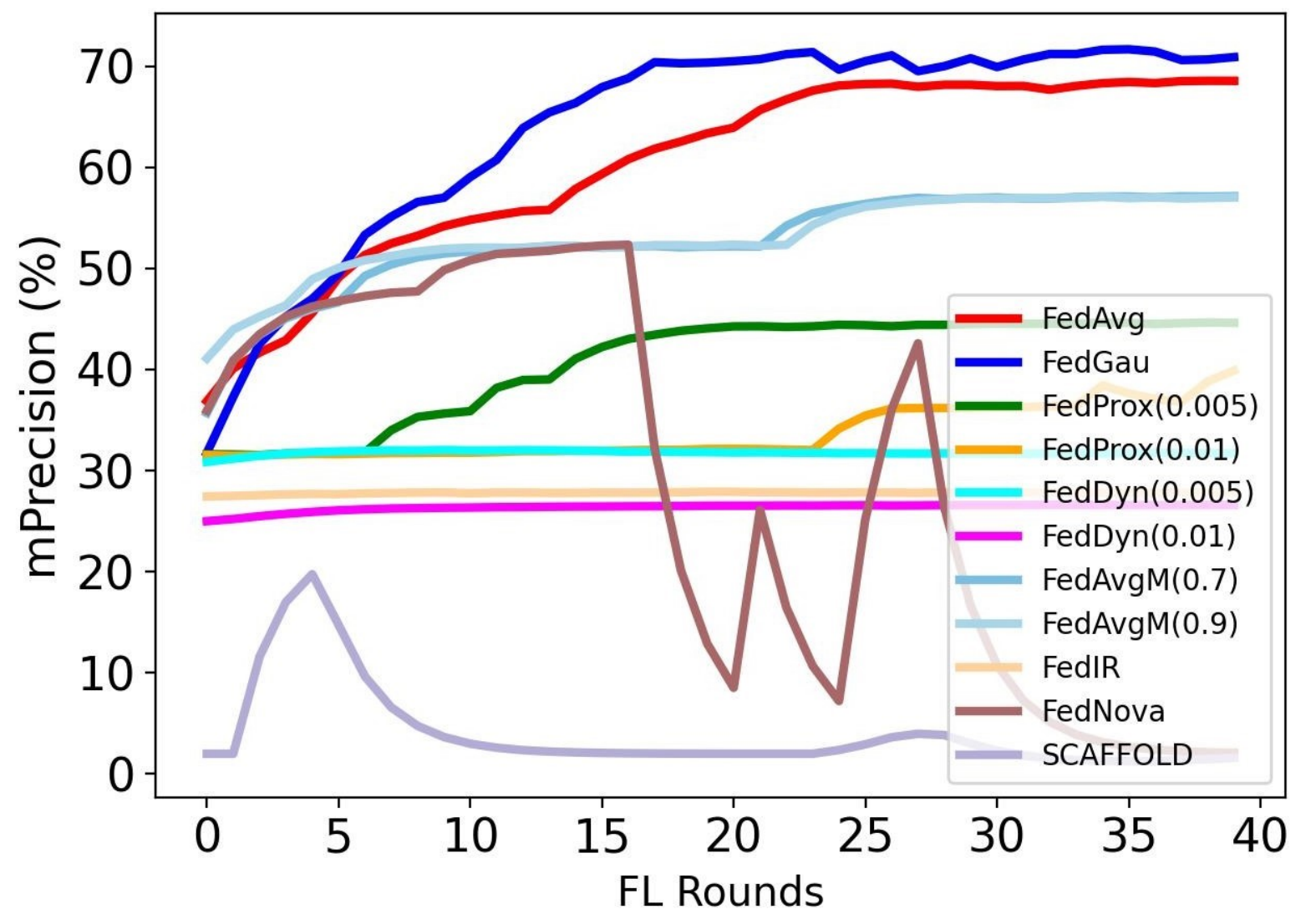}
\label{Fig.Metrics_mPre}
}
\subfloat[mRecall]{\includegraphics[width=0.24\linewidth]{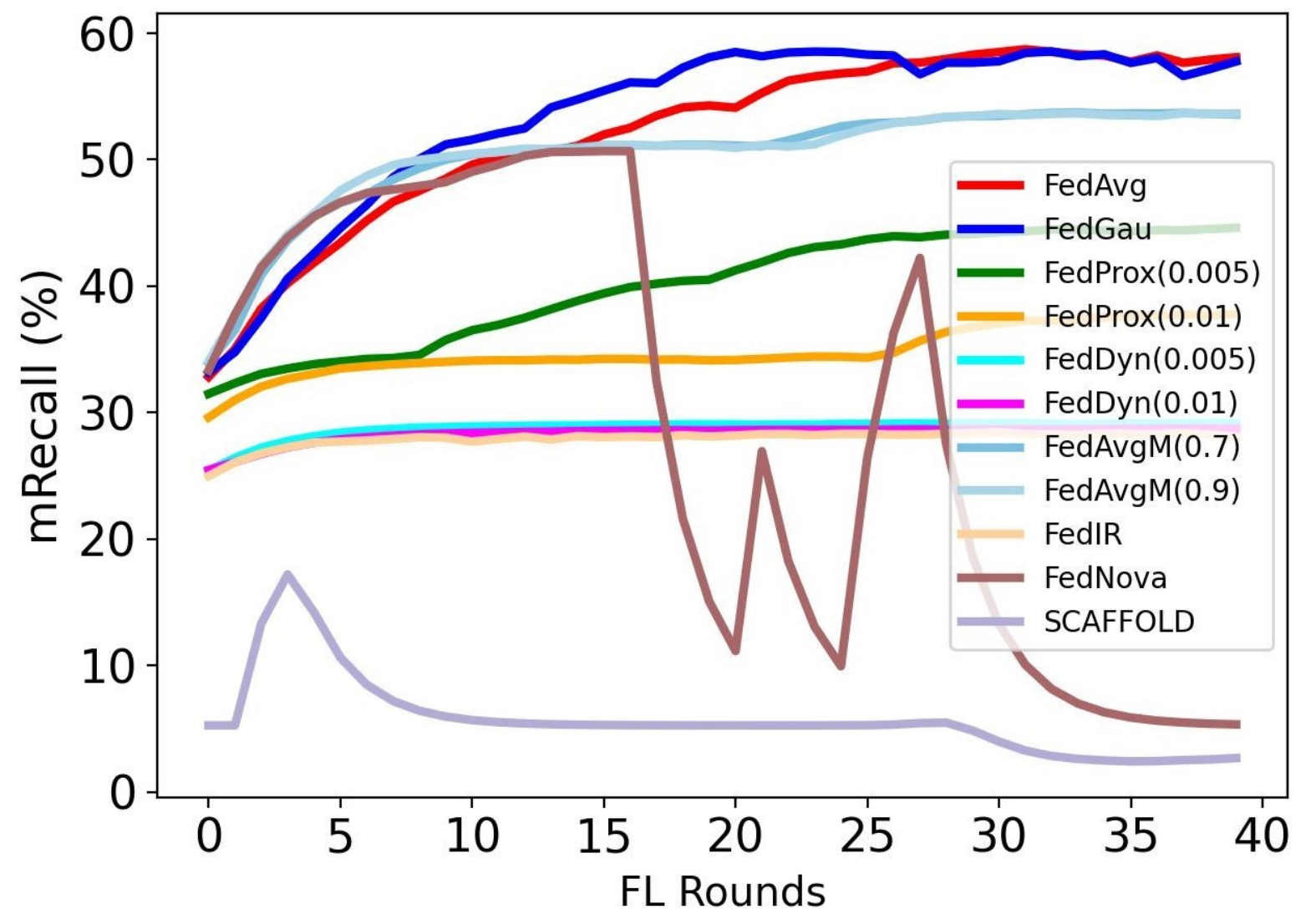}
\label{Fig.Metrics_mRec}
}
\subfloat[mF1]{\includegraphics[width=0.24\linewidth]{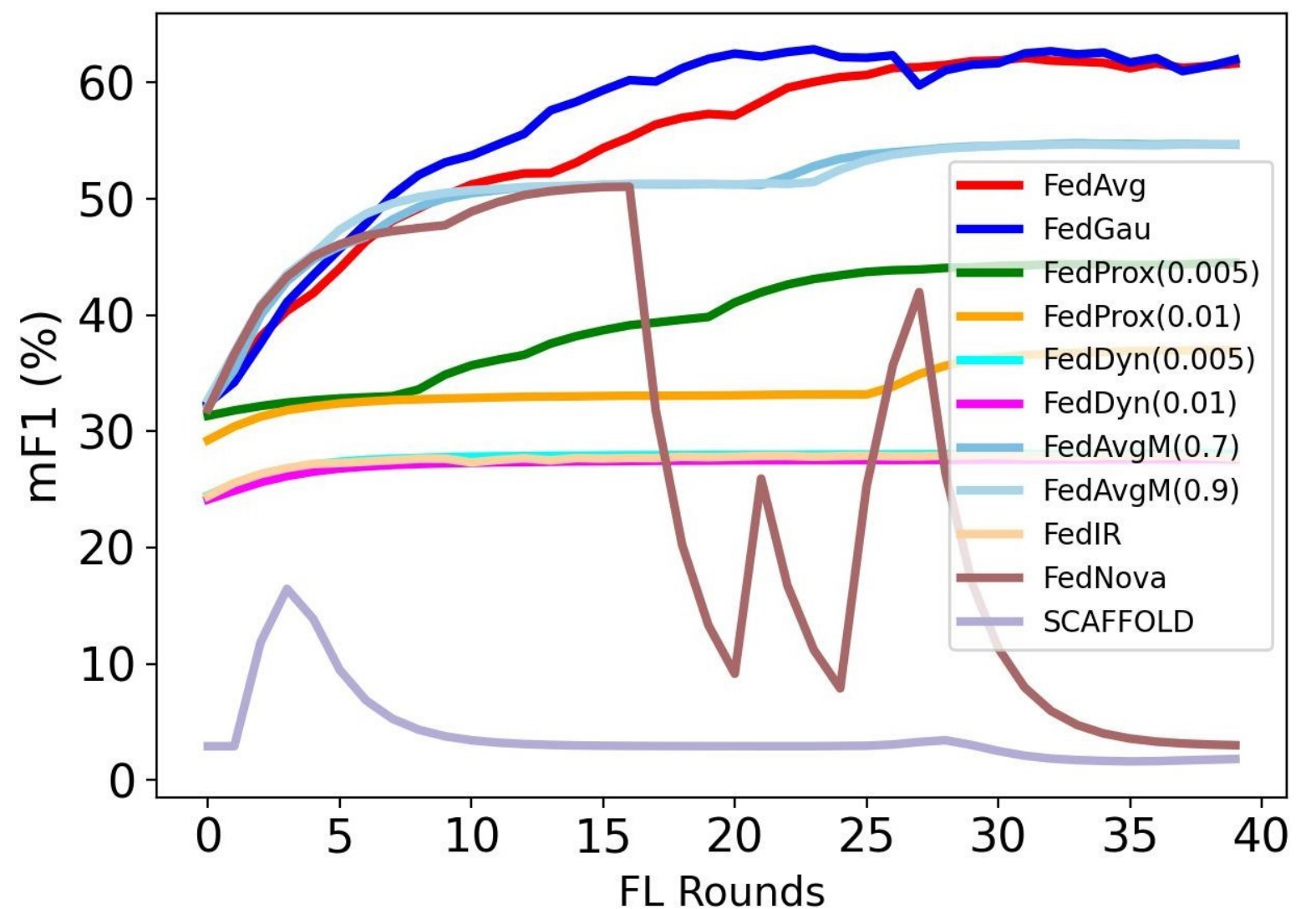}
\label{Fig.Metrics_mF1}
}
\caption{Convergence comparison. Results show that FedGau converges faster than all other FL algorithms across all metrics.}
\label{Fig.Metrics}
\vspace{-0.3cm}
\end{figure*}

We evaluate the model convergence of our proposed FedGau algorithm against other FL algorithms, utilizing the Cityscapes dataset \cite{Cordts2016Cityscapes} and CamVid dataset \cite{brostow2008segmentation}. \Cref{Fig.Metrics} containing various metrics depicts the convergence of the proposed FedGau against other considered FL algorithms. 
It is evident that FedGau, FedAvg, FedAvgM(0.7) and FedAvgM(0.9) converges faster than the rest algorithms with significant margins. Consequently, we mainly compare FedGau against FedAvg, FedAvgM(0.7) and FedAvgM(0.9) hereafter. At the start of training, FedAvgM(0.7) and FedAvgM(0.9) with the server's gradient momentum, exhibit a steeper increase across all metrics compared to FedAvg and FedGau. However, as training progresses, the improvement speed of FedAvg and FedGau surpasses that of FedAvgM(0.7) and FedAvgM(0.9), and this trend continues until the training concludes. Overall, FedGau and FedAvg demonstrate a faster convergence compared to the other FL algorithms.

Focusing on the convergence comparison between FedGau and FedAvg, FedGau consistently exhibits a faster convergence than FedAvg, as detailed in \Cref{Fig.Metrics_mIoU,Fig.Metrics_mPre,Fig.Metrics_mRec,Fig.Metrics_mF1}. To quantify this, we note that  FedGau and FedAvg reach convergence at approximately the 19th and 31st rounds, respectively, as observed in  \textbf{mIoU}. This implies that FedGau's convergence is accelerated by  (31 - 19) / 31 = 38.71\% relative to FedAvg. Similarly, for \textbf{mPrecision}, \textbf{mRecall} and \textbf{mF1}, FedGau's convergence is accelerated by 37.5\%, 35.5\%, and 40.6\%, respectively, compared to FedAvg. 
The reason why FedGau outperforms FedAvg in convergence is that, as emphasized before, FedGau distinguishes each RGB image rather than treating all RGB images uniformly, as is often the case. Furthermore, FedGau takes into account the data's volume and statistical characteristics, rather than just focusing on the amount of data. In other words, FedAvg is a special case of FedGau when datasets on all vehicles are independent identically distributed (i.i.d).

In summary, FedGau holds a substantial advantage in convergence over all other competing FL algorithms across all metrics.

\begin{table}[tp]
\centering
\setlength{\tabcolsep}{1.65pt}
\caption{Quantitative performance of FedGau against other FL algorithms paired with \textbf{DeepLabv3+} on Cityscapes and CamVid datasets}
\begin{tabularx}{\linewidth}{cccccccccc}
\hline
\multirow{2}{*}{FL Algorithms}                                   & \multicolumn{4}{c}{Cityscapes Dataset (\%)}          &                                                                                & \multicolumn{4}{c}{CamVid Dataset (\%)}              \\ \cmidrule{2-5}   \cmidrule{7-10}
                                                                 & mIoU                               & mF1                                & mPre                         & mRec    &                        & mIoU           & mF1            & mPre     & mRec        \\ \hline
FedAvg \cite{https://doi.org/10.48550/arxiv.1602.05629}        & 53.61       & 62.49       & 68.90       & 59.06   &    & 76.72       & 85.59       & 89.89       & 84.45    \\
FedProx (0.005) \cite{li2020federated}                   & 41.51       & 47.22       & 50.22       & 46.78   &    & 75.46       & 82.10       & 82.46       & 81.78    \\
FedProx (0.01) \cite{li2020federated}                    & 33.67       & 37.24       & 41.86       & 38.16   &    & 73.57       & 80.81       & 81.47       & 80.44    \\
FedDyn (0.005) \cite{acar2021federated}               & 25.53       & 28.17       & 32.11       & 29.28    &   & 75.44       & 82.07       & 82.65       & 81.70    \\
FedDyn (0.01) \cite{acar2021federated}                & 24.85       & 27.64       & 26.65       & 28.77    &   & 64.55       & 71.60       & 80.85       & 71.55    \\
FedAvgM (0.7) \cite{hsu2019measuring}                  & 47.28       & 54.79       & 57.14       & 53.74   &    & 76.29       & 82.67       & 83.21       & 82.28    \\
FedAvgM (0.9) \cite{hsu2019measuring}                  & 47.17       & 54.71       & 57.07       & 53.66   &    & 79.23       & 87.07       & 90.03       & 85.26    \\
FedIR \cite{hsu2020federated}                                  & 25.31       & 27.94       & 27.91       & 28.46  &     & 60.38       & 67.27       & 77.12       & 63.89    \\
FedNova \cite{wang2020tackling}                                & 44.38       & 51.03       & 52.34       & 50.68   &    & 75.90       & 82.41       & 83.40       & 81.63    \\
MOON \cite{li2021model}                                        & 48.46       & 56.56       & 59.70       & 55.38   &    & 76.40   & 82.69       & 83.22       & 82.22       \\
SCAFFOLD \cite{karimireddy2020scaffold}                        & 13.55       & 16.44       & 19.76       & 17.19   &    & 23.74       & 30.12       & 42.85       & 31.48    \\
\textbf{\begin{tabular}[c]{@{}c@{}}FedGau (Ours)\end{tabular}} &\textbf{55.44} & \textbf{65.76} & \textbf{75.66} & \textbf{61.12}& & \textbf{80.12} & \textbf{87.70} & \textbf{91.34} & \textbf{86.16} \\ \hline
\end{tabularx}
\label{Tab:metrics_deeplabv3}
\end{table}

\begin{table}[tp]
\centering
\setlength{\tabcolsep}{1.65pt}
\caption{Quantitative performance of FedGau against other FL algorithms paired with \textbf{BiSeNetV2} on Cityscapes and CamVid datasets}
\begin{tabularx}{\linewidth}{cccccccccc}
\hline
\multirow{2}{*}{FL Algorithms}                                   & \multicolumn{4}{c}{Cityscapes Dataset (\%)}     &                                                                                     & \multicolumn{4}{c}{CamVid Dataset (\%)}              \\ \cmidrule{2-5}  \cmidrule{7-10}
                                                                 & mIoU                               & mF1                                & mPre                & mRec     &                         & mIoU           & mF1            & mPre     & mRec        \\ \hline
FedAvg \cite{https://doi.org/10.48550/arxiv.1602.05629}        & 29.32       & 33.29       & 32.47       & 34.52  &     & 50.50       & 58.83       & 60.29       & 59.22          \\
FedProx (0.005) \cite{li2020federated}                   & 24.87       & 27.68       & 26.48       & 29.11    &   & 39.94       & 45.87       & 45.25       & 47.52          \\
FedProx (0.01) \cite{li2020federated}                    & 23.98       & 27.05       & 25.91       & 28.47    &   & 39.54       & 45.68       & 45.50       & 46.56          \\
FedDyn (0.005) \cite{acar2021federated}               & 18.41       & 21.45       & 19.98       & 23.52     &  & 31.63       & 39.22       & 39.73       & 40.65          \\
FedDyn (0.01) \cite{acar2021federated}                & 18.07       & 21.19       & 19.76       & 23.25     &  & 27.49       & 33.87       & 34.91       & 34.37          \\
FedAvgM (0.7) \cite{hsu2019measuring}                  & 30.28       & 34.58       & 34.00 &\textbf{36.34} & & 46.90 & 53.85       & 57.04       & 54.28          \\
FedAvgM (0.9) \cite{hsu2019measuring}                  & 30.33       & 34.61       & 33.80       & 36.28    &   & 50.64       & 59.07       & 60.70       & 58.82          \\
FedIR \cite{hsu2020federated}                                  & 21.58       & 23.70       & 22.68       & 24.95  &     & 37.46       & 43.87       & 45.36       & 43.64          \\
FedCurv \cite{shoham2019overcoming}                            & 29.81       & 33.89       & 33.47       & 35.15  &     & 47.34       & 54.00       & 54.22       & 54.32          \\
FedNova \cite{wang2020tackling}                                & 30.11       & 34.23       & 33.63       & 35.66  &     & \textbf{52.24} & 59.85    & \textbf{62.63} & 58.74          \\
MOON \cite{li2021model}                                        & 30.32       & 34.53       & \textbf{34.01} & 35.79  &     & 49.57       & 58.06    & 60.04       & 57.23          \\
SCAFFOLD \cite{karimireddy2020scaffold}                        & 4.76        & 6.56        & 9.91        & 8.57     &   & 15.50       & 19.61       & 18.32       & 22.72          \\
\textbf{\begin{tabular}[c]{@{}c@{}}FedGau (Ours)\end{tabular}} &\textbf{30.65} & \textbf{34.88} & 33.84 & 36.11 & & 49.21 & \textbf{60.15} & 61.20 & \textbf{59.62} \\ \hline
\end{tabularx}
\label{Tab:metrics_bisenetv2}
\end{table}

\begin{table}[tp]
\centering
\setlength{\tabcolsep}{1.65pt}
\caption{Quantitative performance of FedGau against other FL algorithms paired with \textbf{SegNet} on Cityscapes and CamVid datasets}
\begin{tabularx}{\linewidth}{cccccccccc}
\hline
\multirow{2}{*}{FL Algorithms}                                   & \multicolumn{4}{c}{Cityscapes Dataset (\%)}          &                                                                                & \multicolumn{4}{c}{CamVid Dataset (\%)}              \\ \cmidrule{2-5}  \cmidrule{7-10}
                                                                 & mIoU                               & mF1                                & mPre                         & mRec         &                   & mIoU           & mF1            & mPre     & mRec        \\ \hline
FedAvg \cite{https://doi.org/10.48550/arxiv.1602.05629}        & 28.99       & 33.38       & 33.02       & 34.51  &     & 45.96       & 46.17       & 49.16       & \textbf{52.78}  \\
FedProx (0.005) \cite{li2020federated}                   & 20.48       & 22.99       & 21.43       & 24.85   &    & 42.89       & 47.79       & 46.89       & 49.21          \\
FedProx (0.01) \cite{li2020federated}                    & 11.43       & 13.18       & 11.89       & 15.14   &    & 43.36       & 48.12       & 47.26       & 49.62          \\
FedDyn (0.005) \cite{acar2021federated}               & 20.48       & 22.99       & 21.57       & 24.70   &    & 33.76       & 38.44       & 36.98       & 40.32          \\
FedDyn (0.01) \cite{acar2021federated}                & 16.00       & 18.14       & 16.67       & 20.03   &    & 32.85       & 37.89       & 36.75       & 39.66          \\
FedAvgM (0.7) \cite{hsu2019measuring}                  & 26.91       & 30.50       & 30.33       & 31.33  &     & 42.68       & 47.75       & 47.27       & 48.69          \\
FedAvgM (0.9) \cite{hsu2019measuring}                  & 27.84       & 32.06       & 32.57       & 32.47  &     & 42.50       & 47.61       & 47.41       & 48.30          \\
FedIR \cite{hsu2020federated}                                  & 12.81       & 14.83       & 13.85       & 16.40  &     & 26.83       & 30.79       & 29.12       & 33.38          \\
FedCurv \cite{shoham2019overcoming}                    & 29.30   & 33.49   & 33.19   & 34.61 &  & 43.15   & 48.03   & 47.64    & 48.87          \\
FedNova \cite{wang2020tackling}                                & 27.46        & 31.05    & 36.38  & 33.20 &    & 41.93       & 47.23       & 46.86       & 48.27          \\
MOON \cite{li2021model}                                        & \textbf{31.10}  & \textbf{36.17}   & \textbf{37.39}   & \textbf{36.47} &  & 42.37    & 47.50   & 47.25   & 48.25          \\
SCAFFOLD \cite{karimireddy2020scaffold}                        & 0.68        & 1.27        & 5.21        & 5.49   &     & 10.86       & 15.17       & 13.21       & 18.30          \\
\textbf{\begin{tabular}[c]{@{}c@{}}FedGau (Ours)\end{tabular}} &29.60 & 34.40 & 29.50 & 34.76 & & \textbf{47.80} & \textbf{49.45} & \textbf{50.77} & 49.39 \\ \hline
\end{tabularx}
\label{Tab:metrics_segnet}
\end{table}

\subsection{Evaluation on FedGau's performance}
\label{fedgau_perf}
In this section, we will evaluate the model performance of FedGau against other competitors from both quantitative and qualitative perspectives.

\subsubsection{Quantitative Evaluation}
We conduct a comprehensive set of experiments to compare the model performance of FedGau against other FL algorithms when paired with DeepLabv3+ \cite{chen2018encoderdecoder}, BiSeNetV2 \cite{yu2021bisenet}, and SegNet \cite{badrinarayanan2017segnet}. The results of the DeepLabv3+ are presented in \Cref{Tab:metrics_deeplabv3}, which clearly indicates that FedGau surpasses all other algorithms in almost all metric on both Cityscapes dataset and CamVid dataset. Specifically, for Cityscapes dataset, FedGau significantly outperforms the second-best FL algorithm (i.e., FedAvg) by margins of (55.44 - 53.61) / 53.61  = 3.41\%, (65.76 - 62.49) / 62.49 = 5.23\%, (75.66 - 68.90) / 68.90  = 9.81\% and (61.12 - 59.06) / 59.06 \%  = 3.49\% in mIoU, mPrecision, mRecall, and mF1, respectively. For CamVid dataset, the improvements of FedGau over FedAvg are (80.12 - 76.72) / 76.72  = 4.43\%, (87.70 - 85.59) / 85.59 = 2.47\%, (91.34 - 89.89) / 89.89 = 1.61\% and (86.16 - 84.45) / 84.45 = 2.02\% in mIoU, mPrecision, mRecall, and mF1, respectively. The outcomes of BiSeNetV2 and SegNet are documented in \Cref{Tab:metrics_bisenetv2,Tab:metrics_segnet}, respectively. They also demonstrate that FedGau consistently achieves superior or competitive performance in almost all metrics.

In addition, by analyzing \Cref{Tab:metrics_deeplabv3}, \Cref{Tab:metrics_bisenetv2}, and \Cref{Tab:metrics_segnet}, we can deduce two main observations: First, there seems to exist a negative correlation between the model performance of FL algorithms and the complexity of tasks or the complexity of datasets. For instance, algorithms like FedProx, FedDyn, MOON, and FedNova show better results in simpler classification tasks but fall behind in TriSU tasks. Likewise, almost all FL algorithms exhibit worse performance on more complex Cityscapes dataset compared to simpler CamVid dataset. Second, deep neural network (DNN) architecture significantly impacts the model performance across all FL algorithms. As evidence, FedGau achieves a mIoU of 55.44\% with the DeepLabv3+, which drops to 30.65\% and 29.60\% with the BiSeNetV2 and SegNet, respectively. This trend is generally observable across other FL algorithms as well.

\subsubsection{Qualitative Evaluation}
\Cref{tab:semantic_pred} illustrates the qualitative performance of semantic segmentation of FedGau against other FL algorithms, including FedAvg, SCAFFOLD, FedAvgM(0.7), FedIR, FedDyn(0.005), FedProx(0.005), FedNova, on a set of five RGB images. To gauge the effectiveness of each algorithm's predictions, we examine how closely their segmented outputs align with the ground truth and the original images. The comparison reveals that FedGau's outputs are consistently more accurate in capturing both the overall scene and intricate details across all images. Notably, FedGau is the only algorithm that reliably identifies subtle elements such as poles, depicted in light yellow, which tend to be overlooked by most other FL algorithms.

\begin{table*}[tp]
\centering
\renewcommand{\arraystretch}{0.24}
\addtolength{\tabcolsep}{-0.45pt}
\caption{Qualitative performance of FedGau against other FL algorithms}
\begin{tabularx}{\linewidth}{|l|lllll|}
\hline
\verticaltext[26pt]{\tiny}{Raw RGBs} &\includegraphics[width=0.187\linewidth, height=0.12\linewidth]{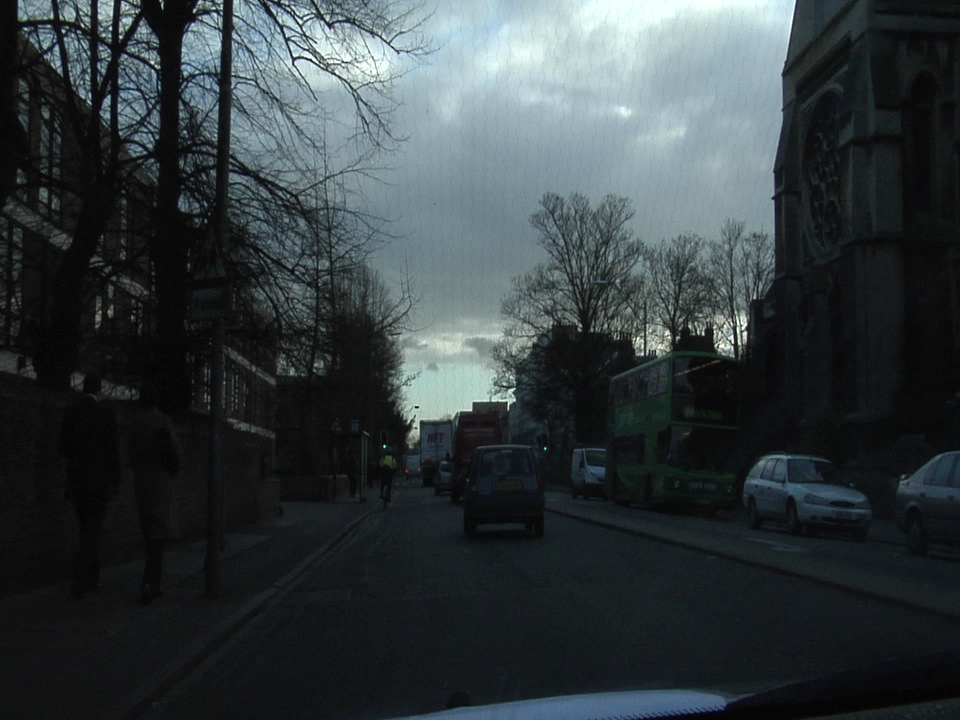} &\hspace{-0.47cm}
\includegraphics[width=0.187\linewidth, height=0.12\linewidth]{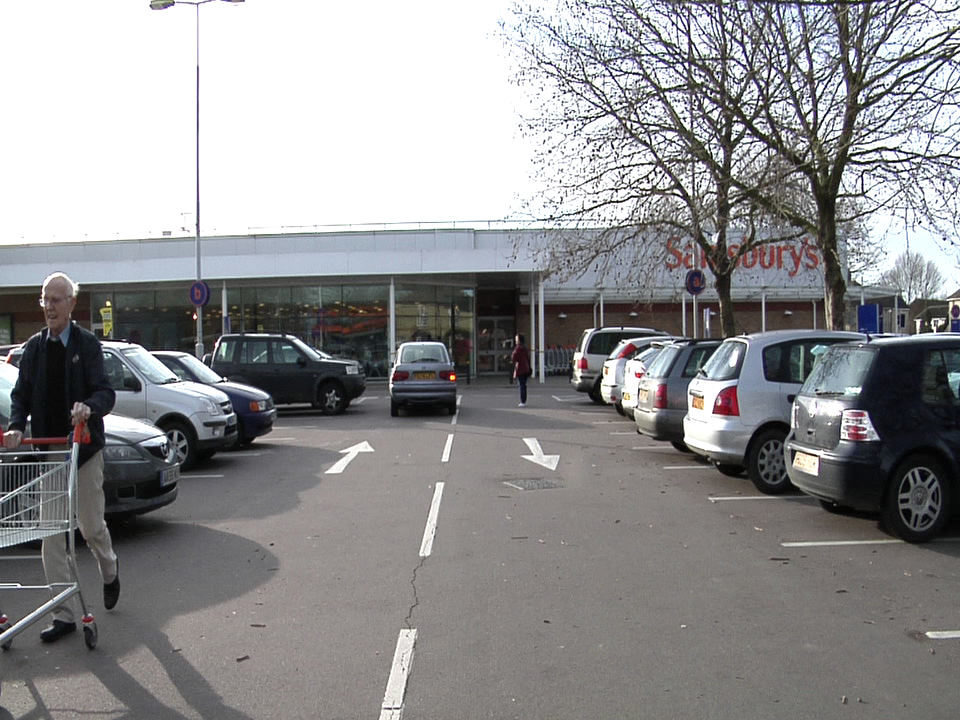} &\hspace{-0.47cm}
\includegraphics[width=0.187\linewidth, height=0.12\linewidth]{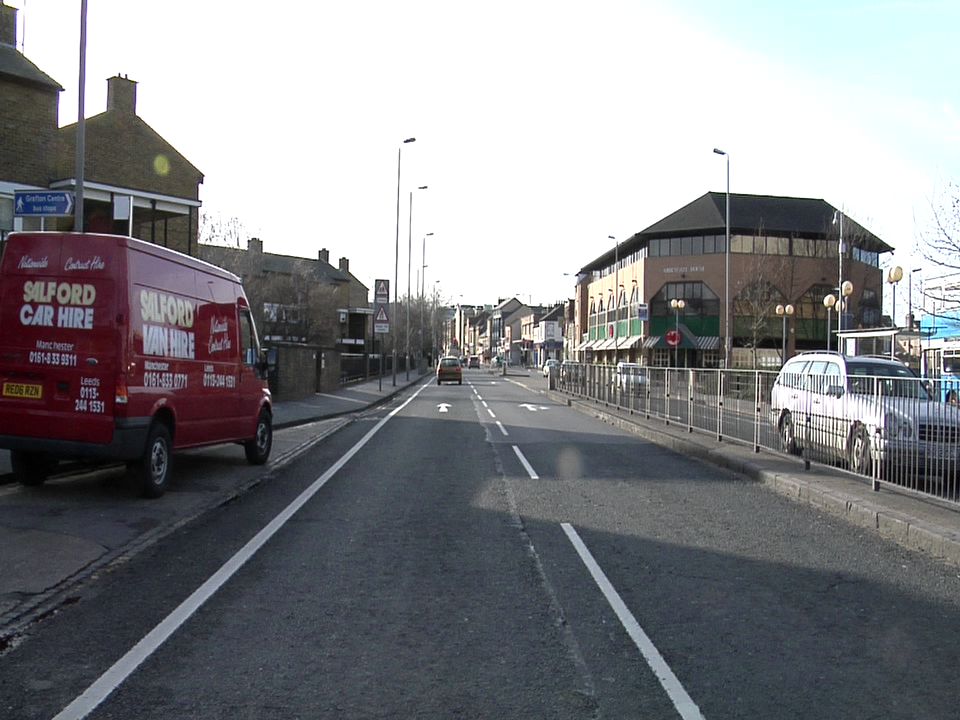} &\hspace{-0.47cm}
\includegraphics[width=0.187\linewidth, height=0.12\linewidth]{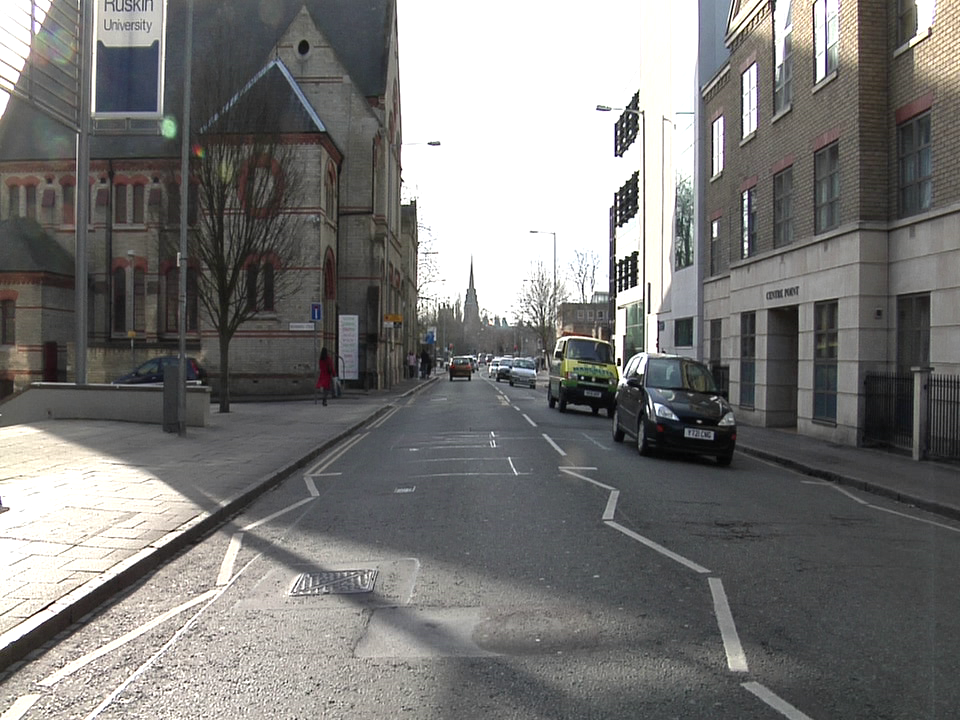} &\hspace{-0.47cm}
\includegraphics[width=0.187\linewidth, height=0.12\linewidth]{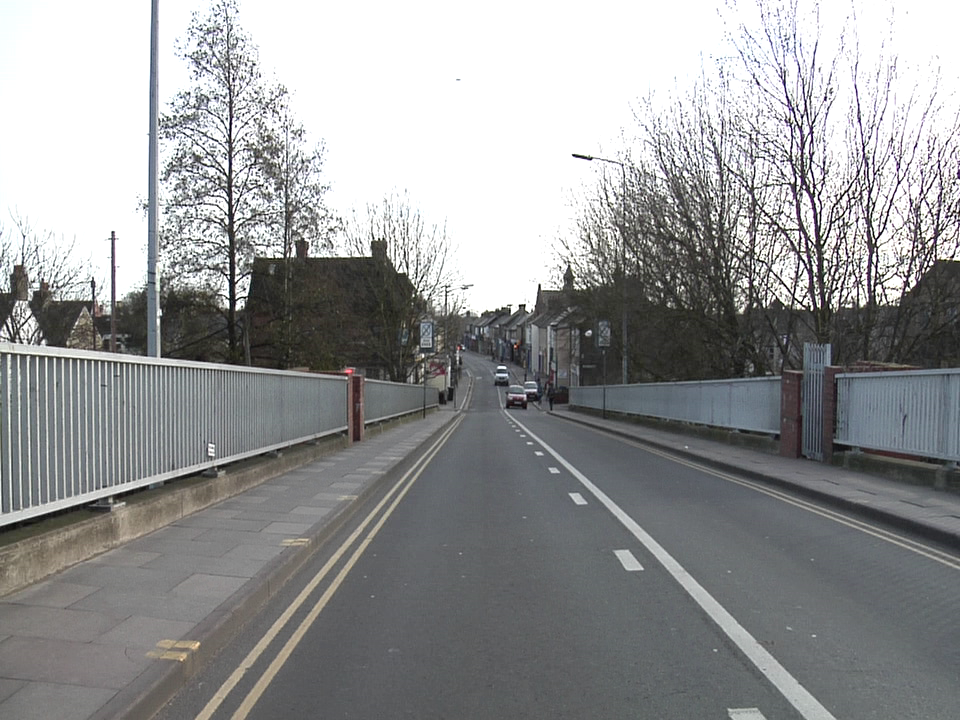}\\
\hline

\verticaltext[26pt]{\tiny}{Ground Truth} &\includegraphics[width=0.187\linewidth, height=0.12\linewidth]{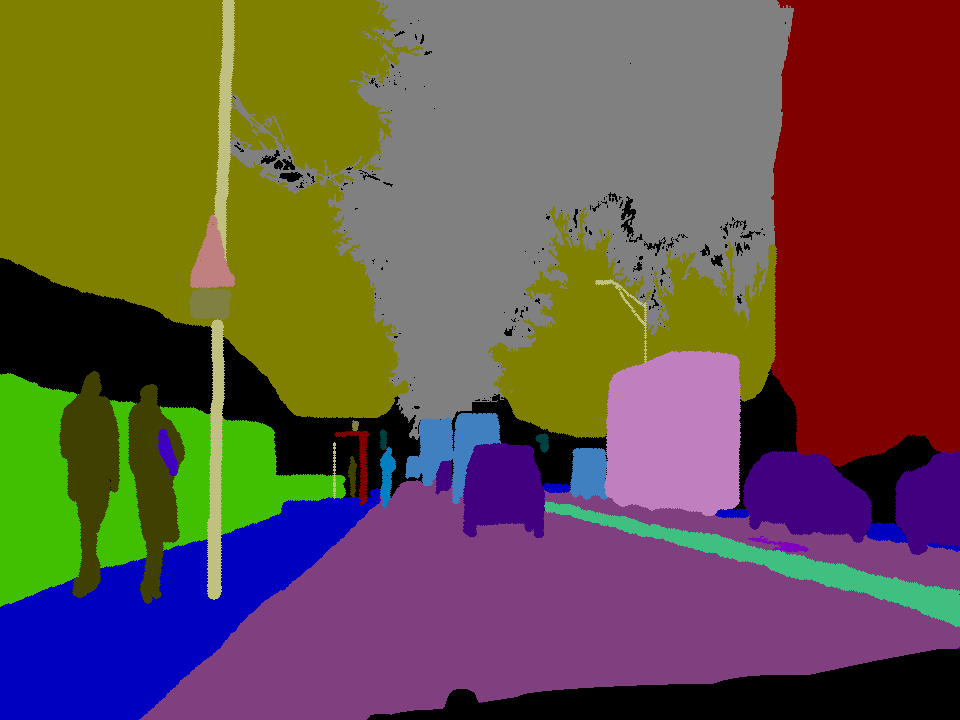} &\hspace{-0.47cm}
\includegraphics[width=0.187\linewidth, height=0.12\linewidth]{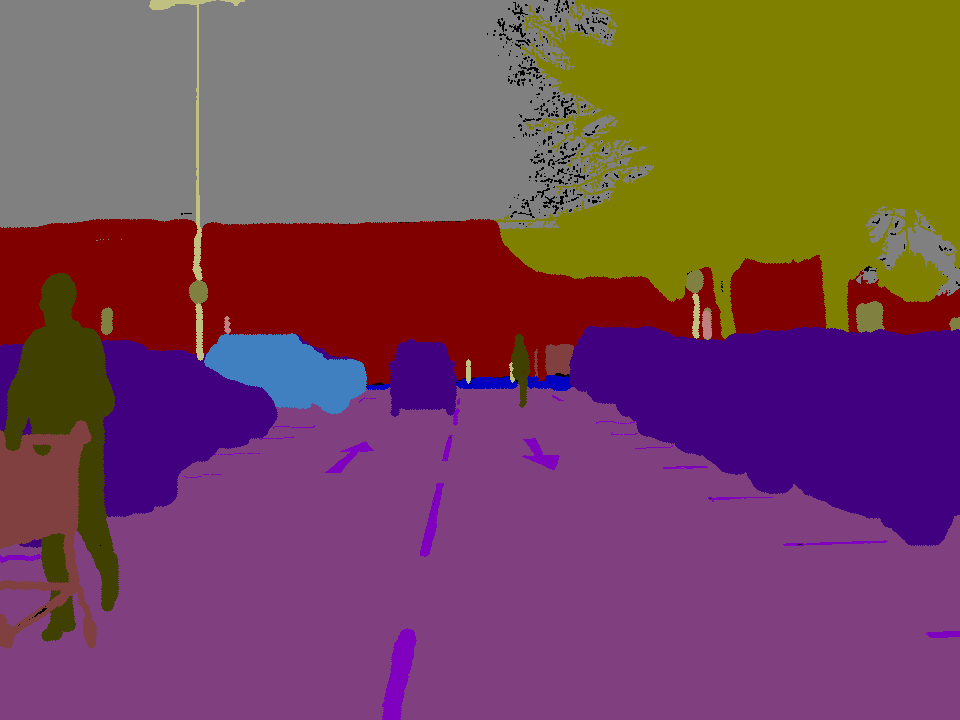} &\hspace{-0.47cm}
\includegraphics[width=0.187\linewidth, height=0.12\linewidth]{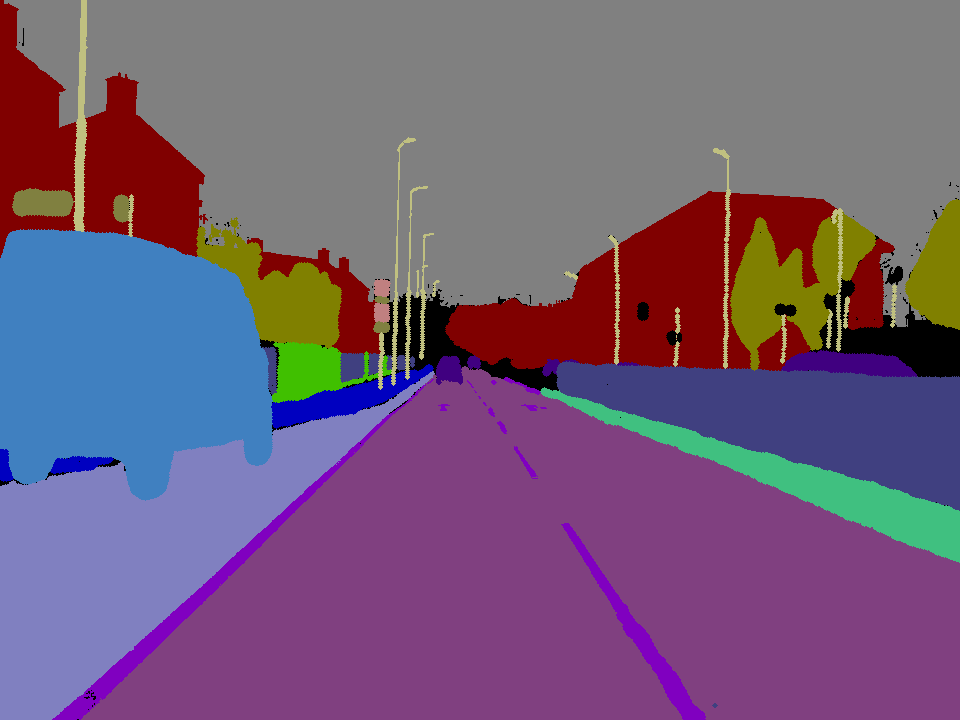} &\hspace{-0.47cm}
\includegraphics[width=0.187\linewidth, height=0.12\linewidth]{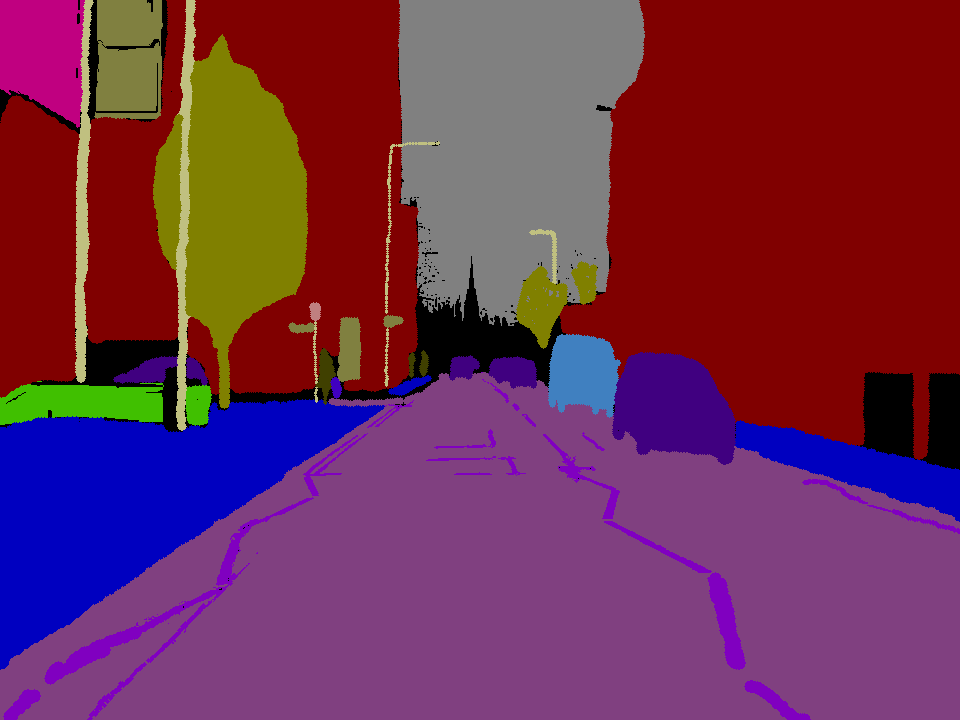} &\hspace{-0.47cm}
\includegraphics[width=0.187\linewidth, height=0.12\linewidth]{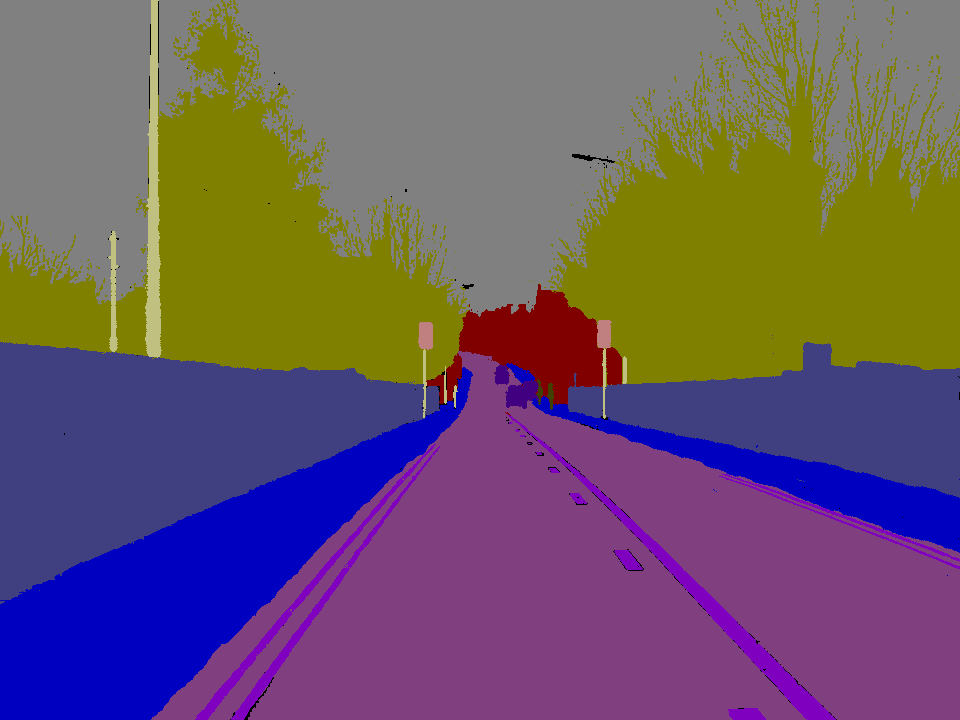}\\
\hline

\verticaltext[26pt]{\tiny}{SCAFFOLD} &\includegraphics[width=0.187\linewidth, height=0.12\linewidth]{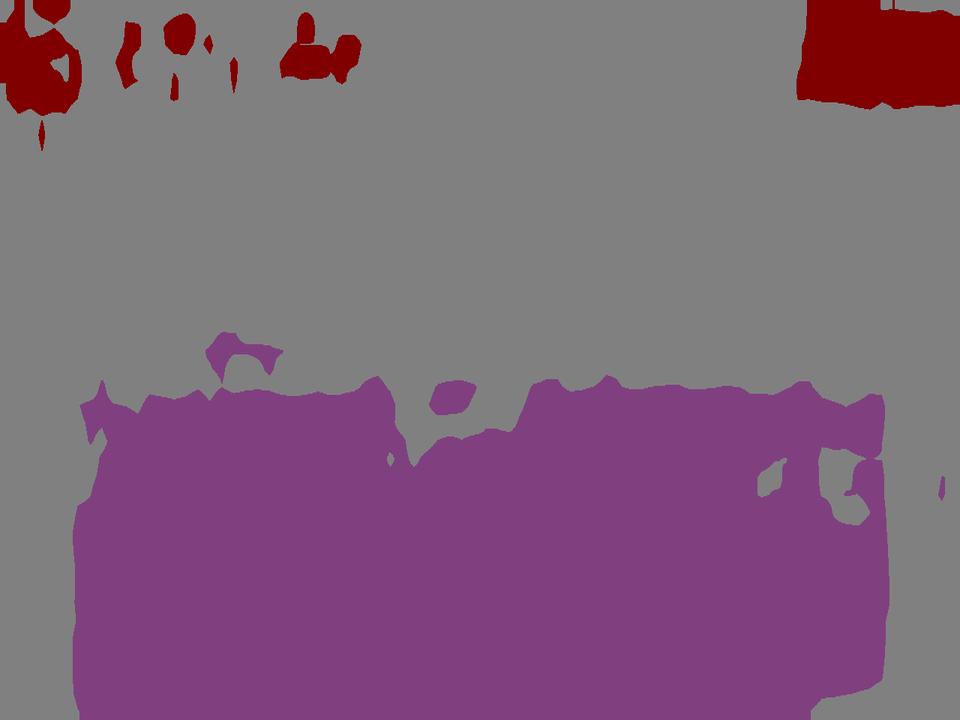} &\hspace{-0.47cm}
\includegraphics[width=0.187\linewidth, height=0.12\linewidth]{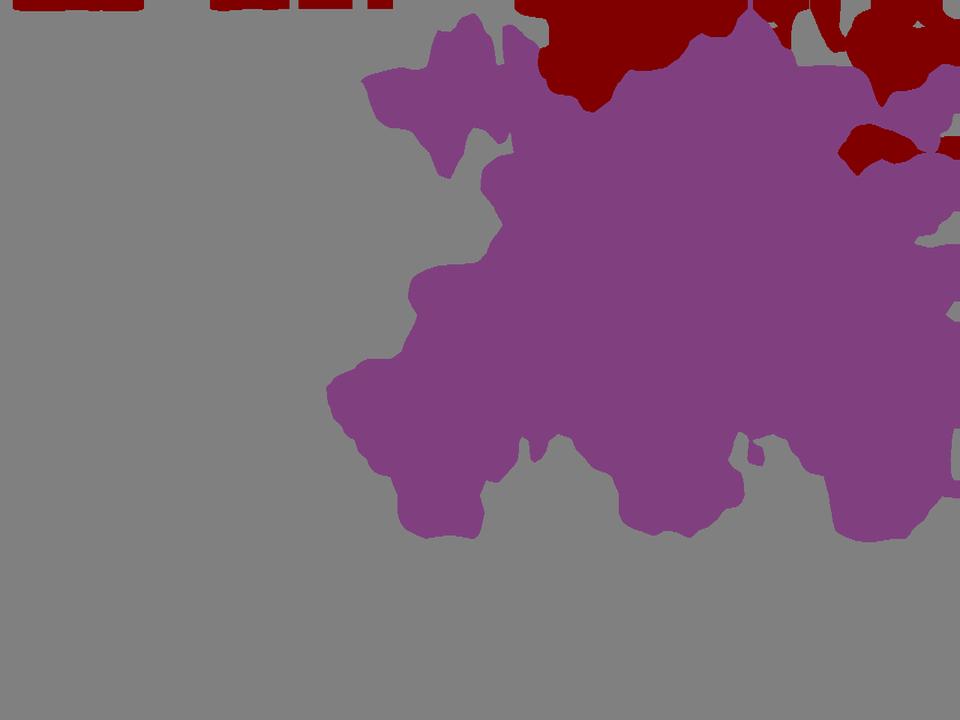} &\hspace{-0.47cm}
\includegraphics[width=0.187\linewidth, height=0.12\linewidth]{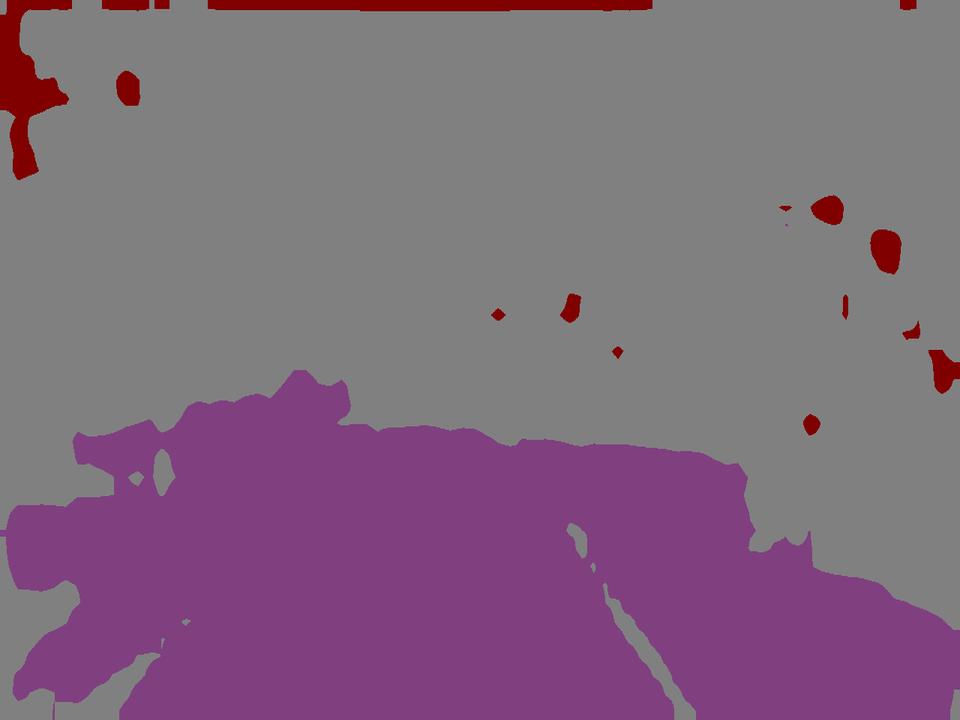} &\hspace{-0.47cm}
\includegraphics[width=0.187\linewidth, height=0.12\linewidth]{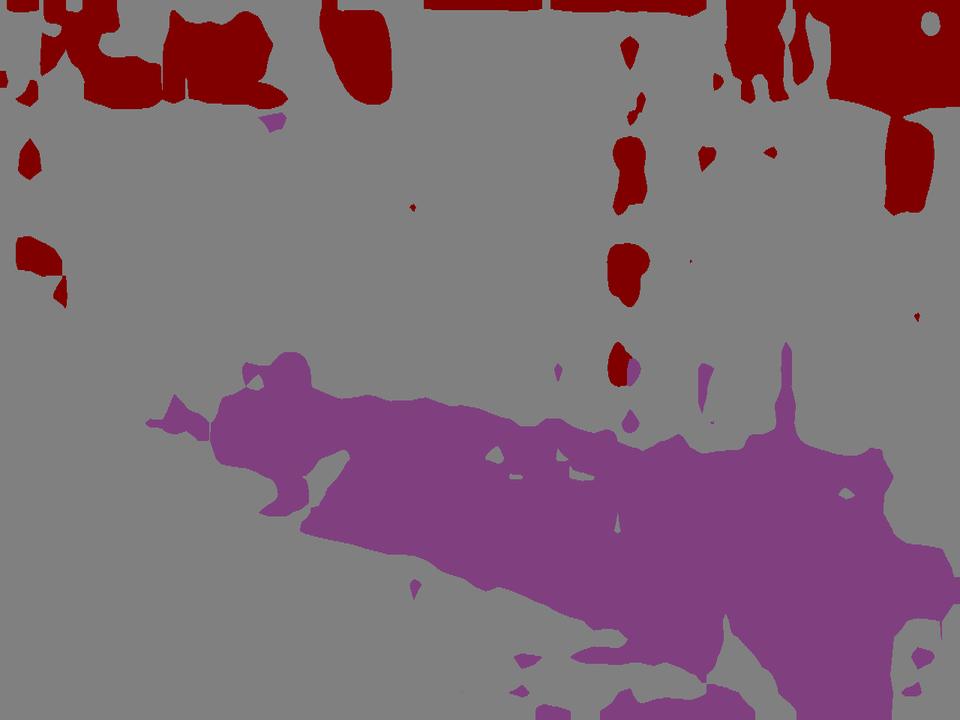} &\hspace{-0.47cm}
\includegraphics[width=0.187\linewidth, height=0.12\linewidth]{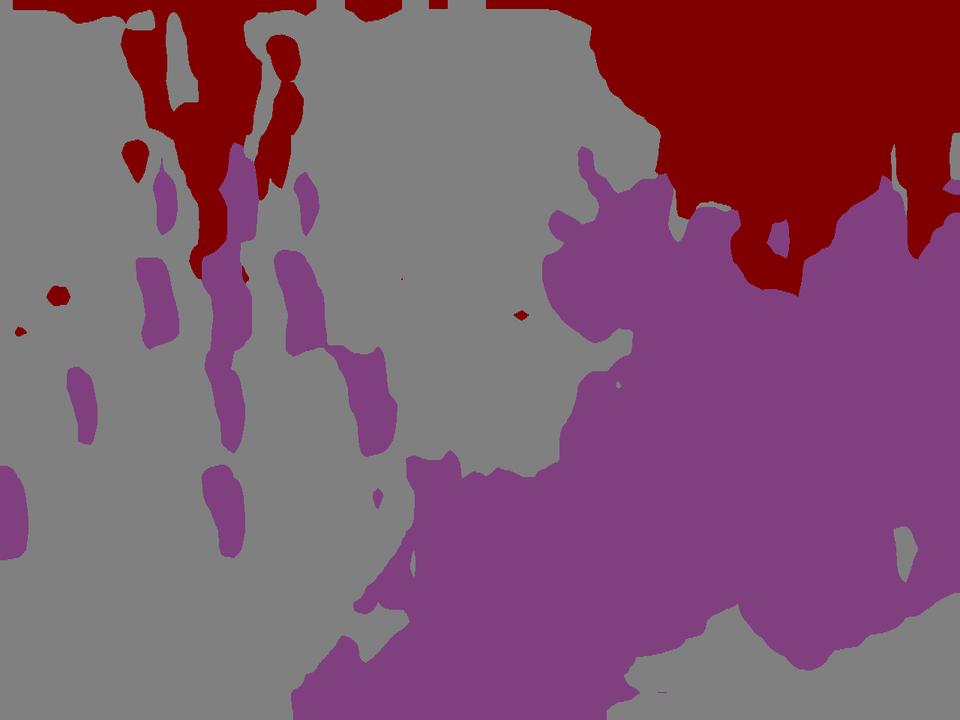}\\
\hline

\verticaltext[26pt]{\tiny}{FedAvg} &\includegraphics[width=0.187\linewidth, height=0.12\linewidth]{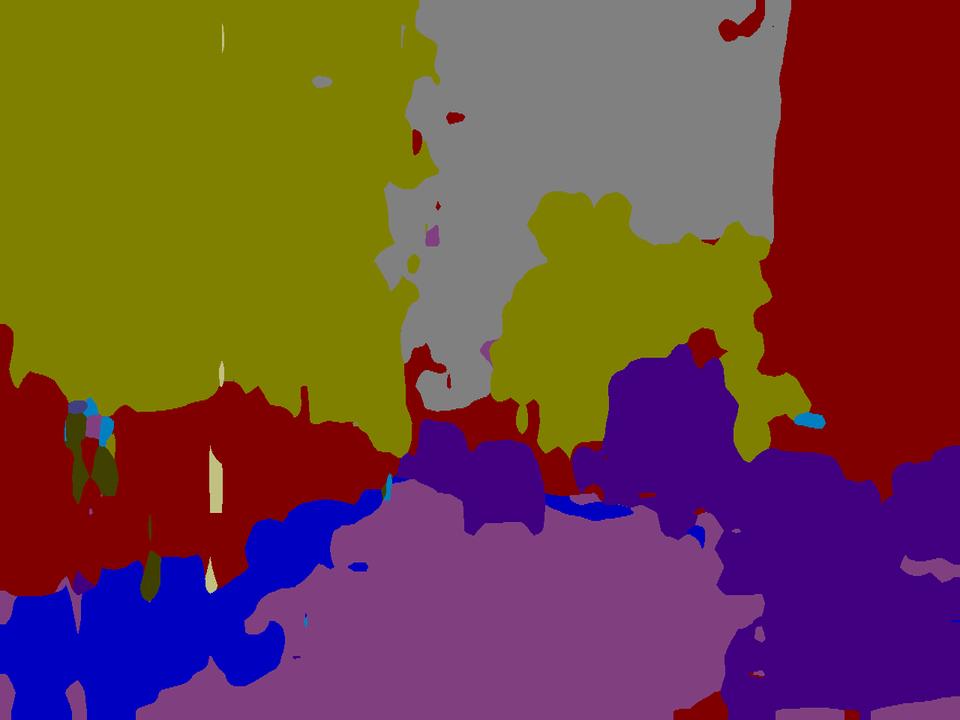} &\hspace{-0.47cm}
\includegraphics[width=0.187\linewidth, height=0.12\linewidth]{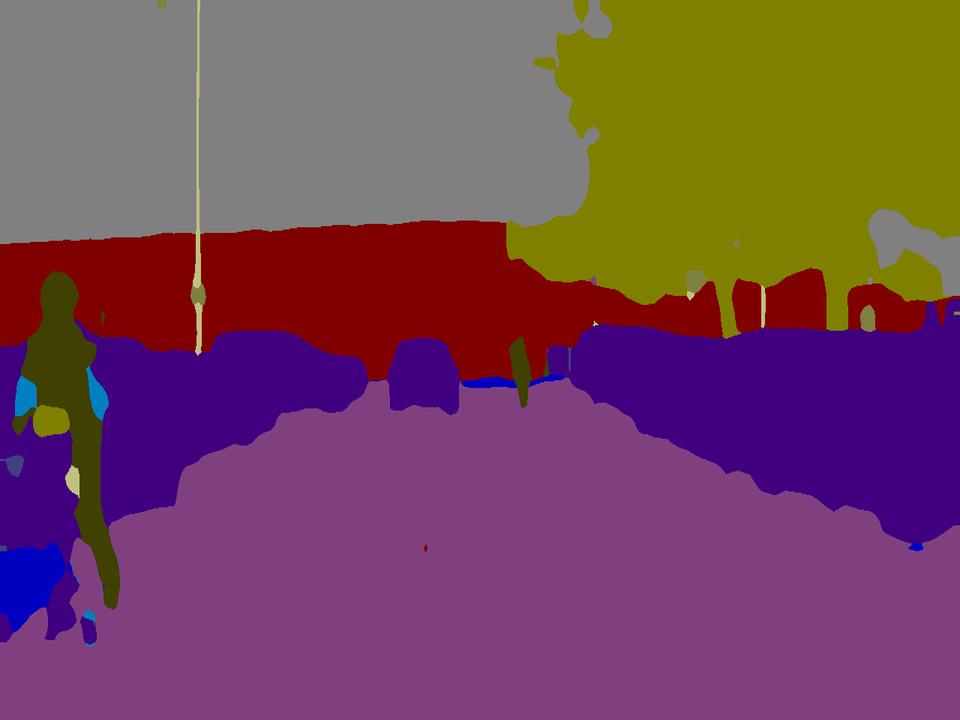} &\hspace{-0.47cm}
\includegraphics[width=0.187\linewidth, height=0.12\linewidth]{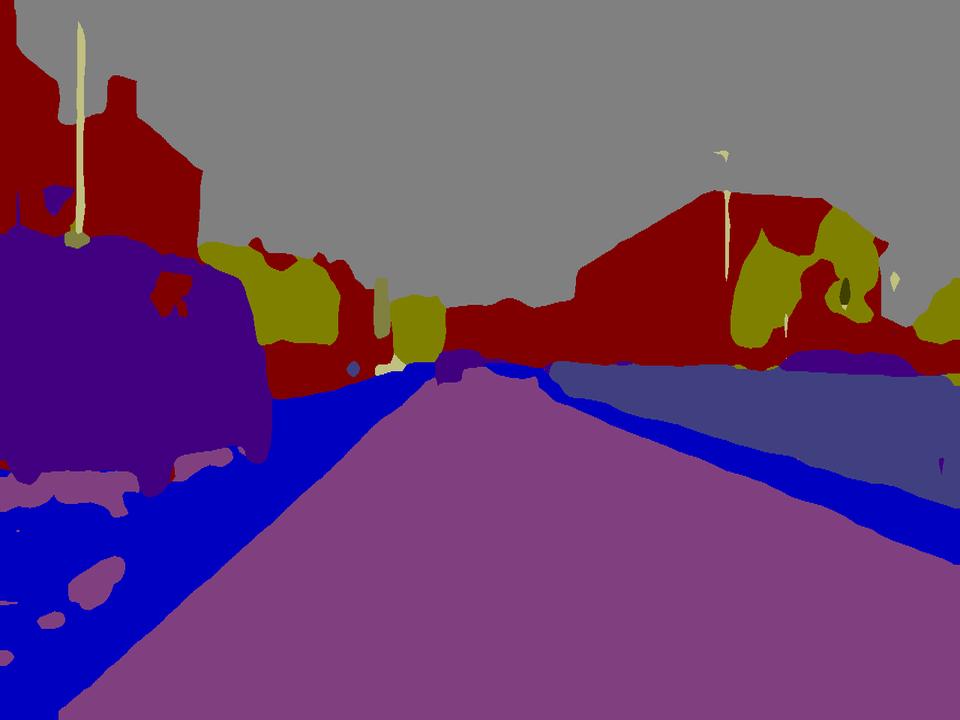} &\hspace{-0.47cm}
\includegraphics[width=0.187\linewidth, height=0.12\linewidth]{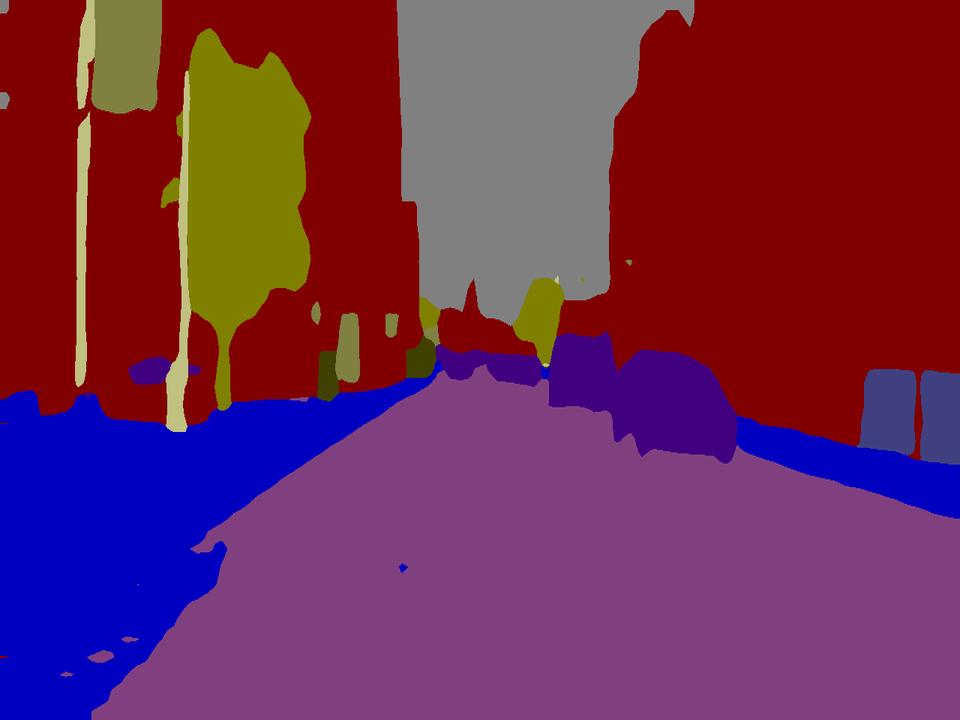} &\hspace{-0.47cm}
\includegraphics[width=0.187\linewidth, height=0.12\linewidth]{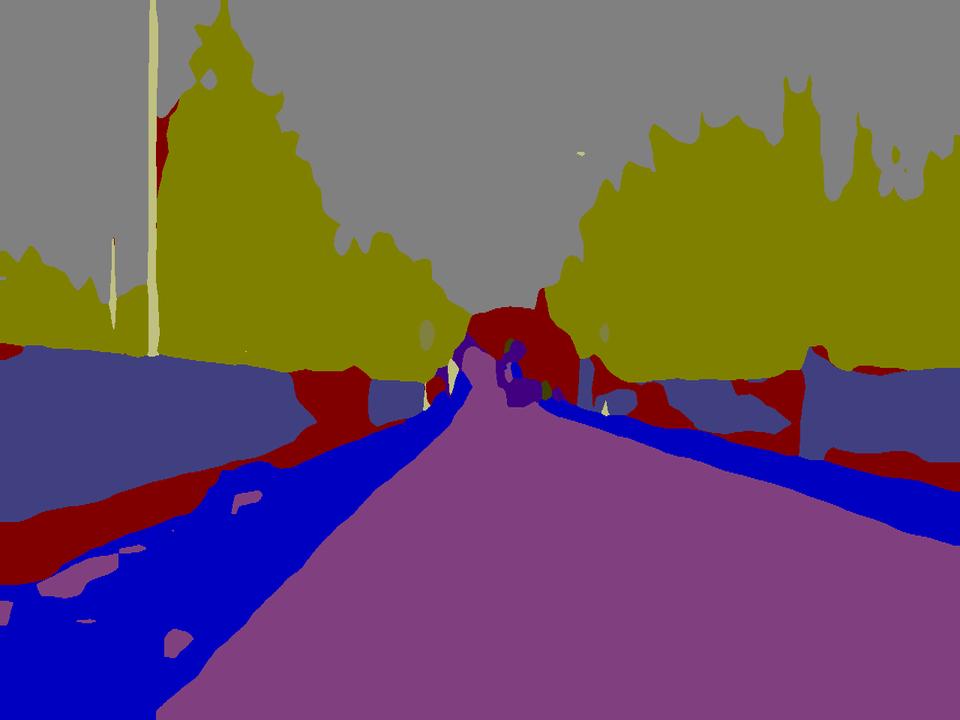}\\
\hline

\verticaltext[26pt]{\tiny}{FedAvgM(0.7)} &\includegraphics[width=0.187\linewidth, height=0.12\linewidth]{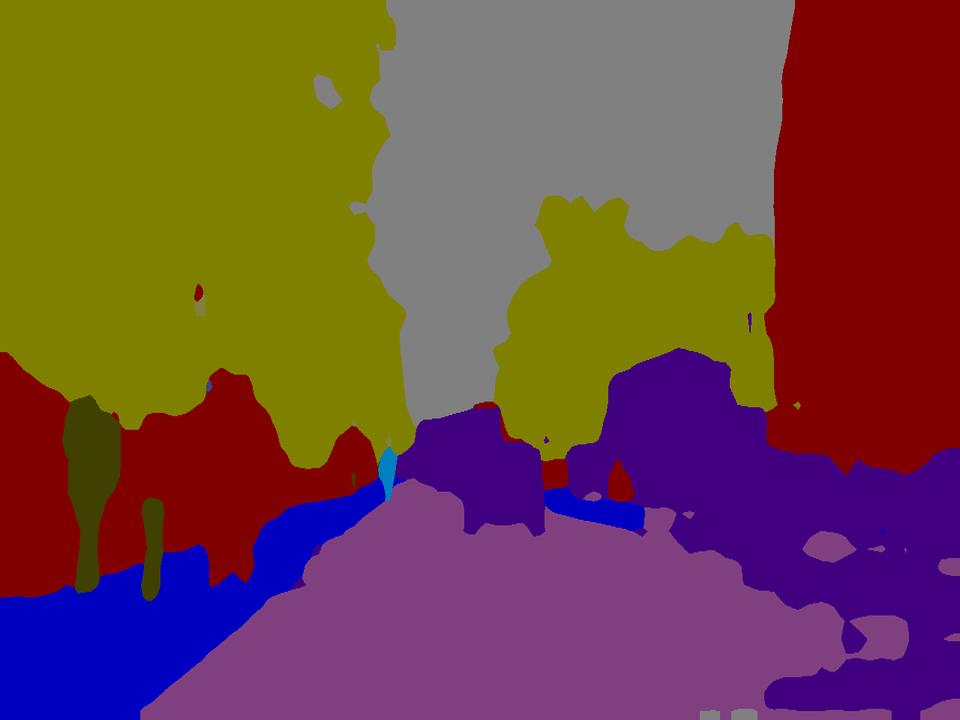} &\hspace{-0.47cm}
\includegraphics[width=0.187\linewidth, height=0.12\linewidth]{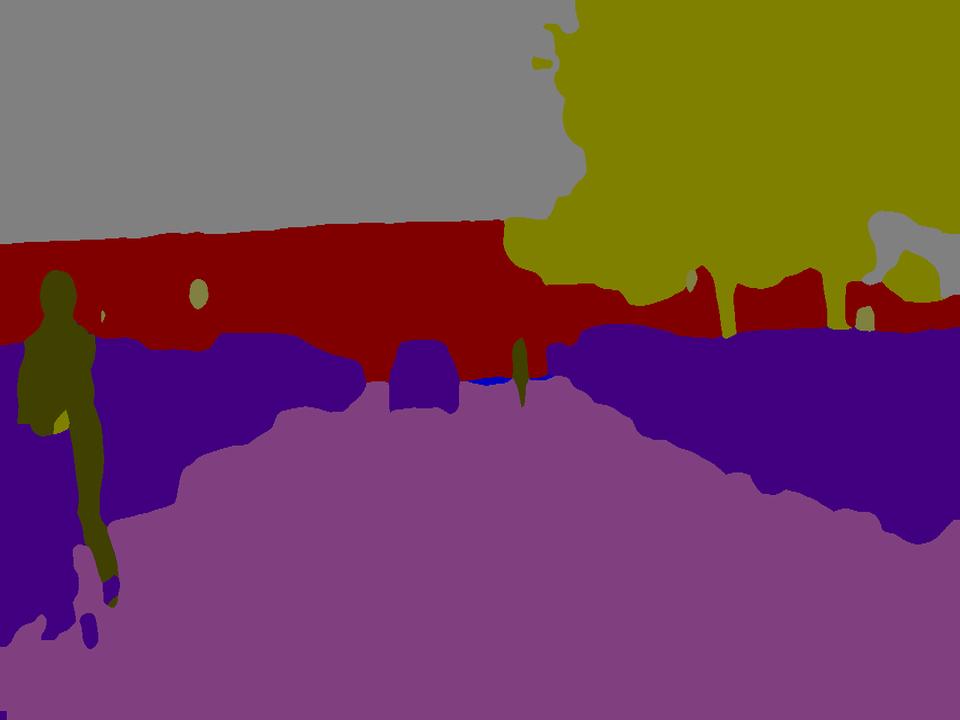} &\hspace{-0.47cm}
\includegraphics[width=0.187\linewidth, height=0.12\linewidth]{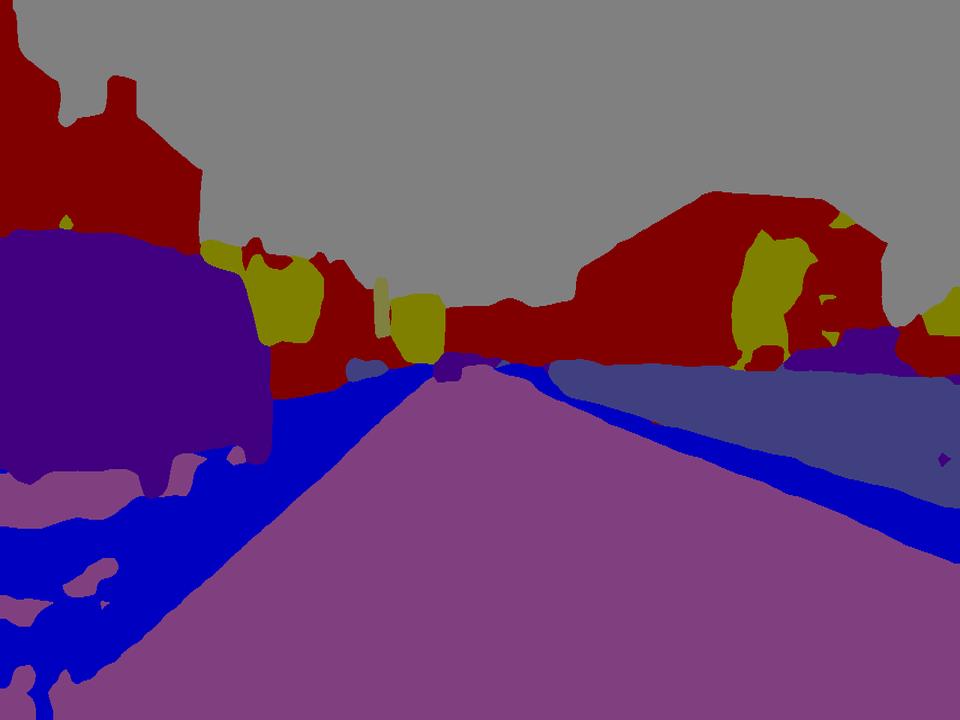} &\hspace{-0.47cm}
\includegraphics[width=0.187\linewidth, height=0.12\linewidth]{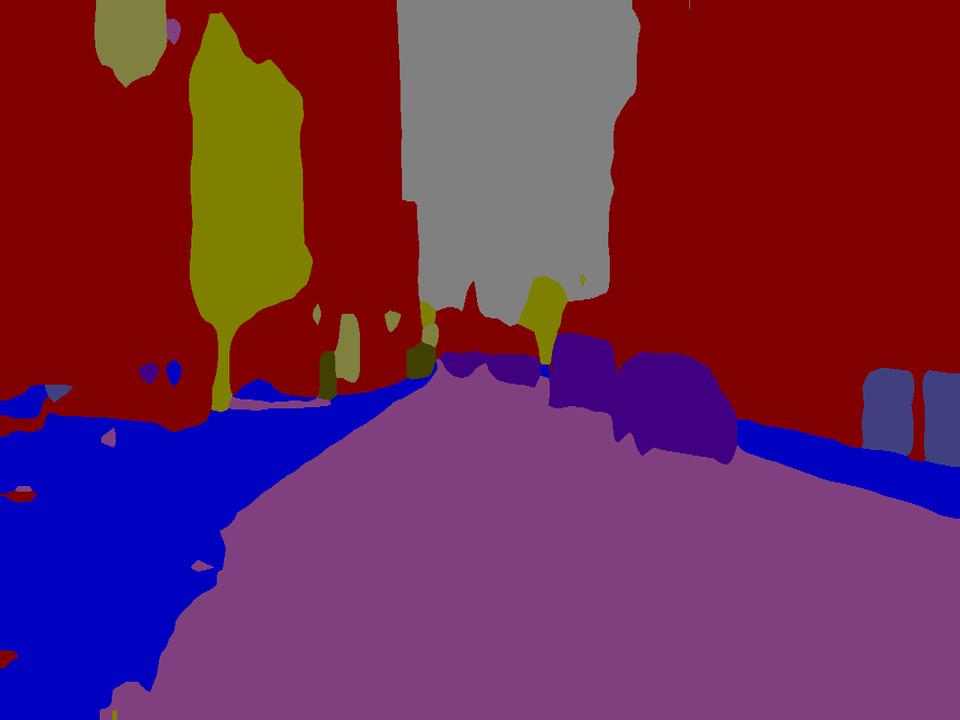} &\hspace{-0.47cm}
\includegraphics[width=0.187\linewidth, height=0.12\linewidth]{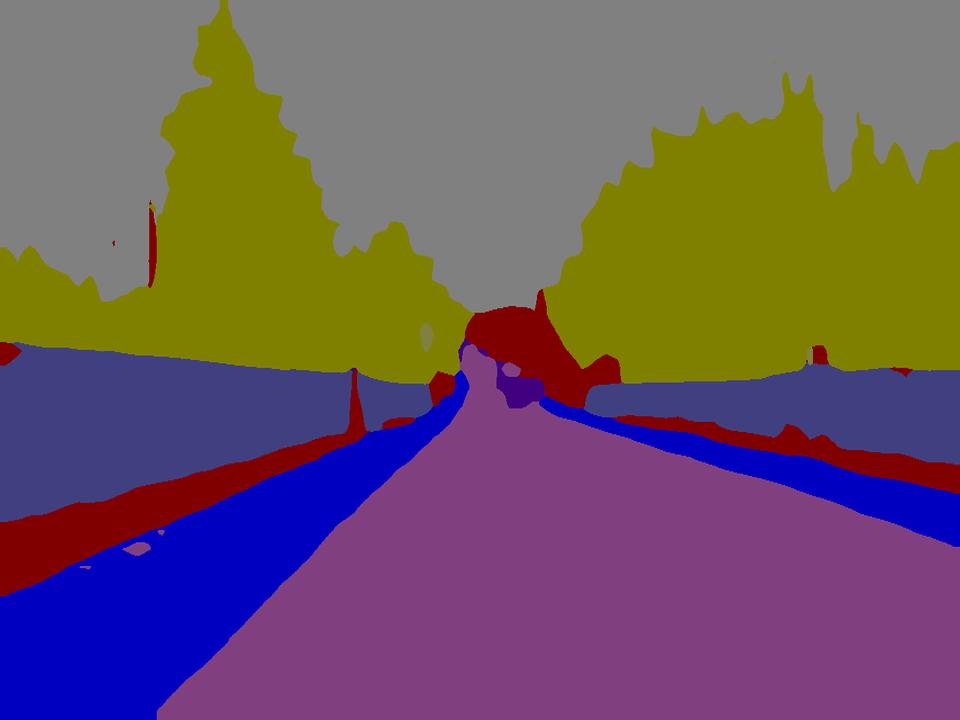}\\
\hline

\verticaltext[26pt]{\tiny}{FedIR} &\includegraphics[width=0.187\linewidth, height=0.12\linewidth]{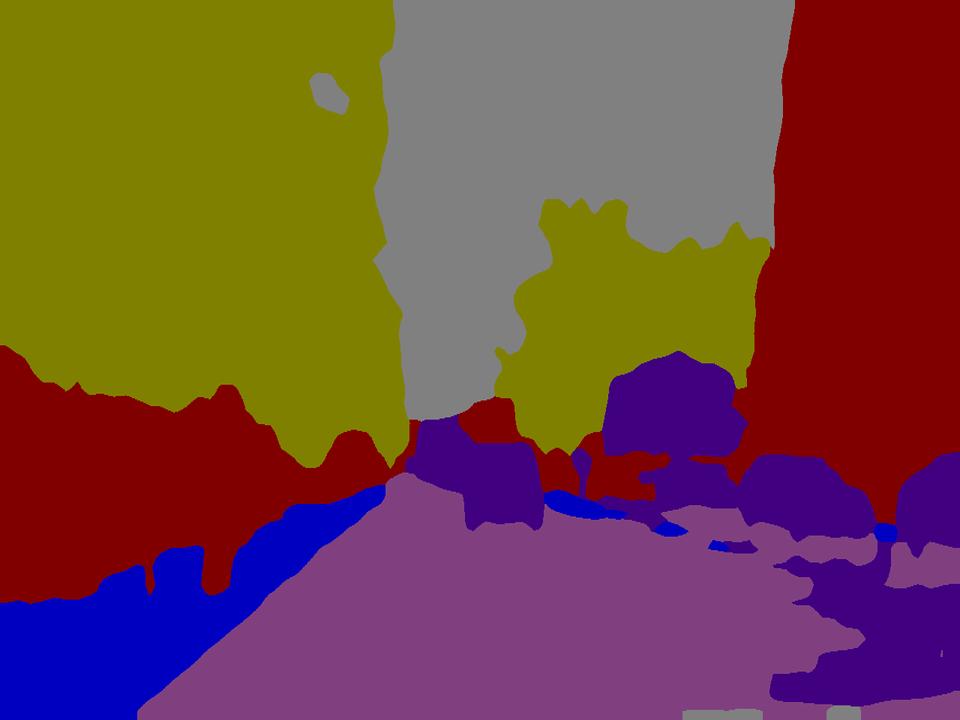} &\hspace{-0.47cm}
\includegraphics[width=0.187\linewidth, height=0.12\linewidth]{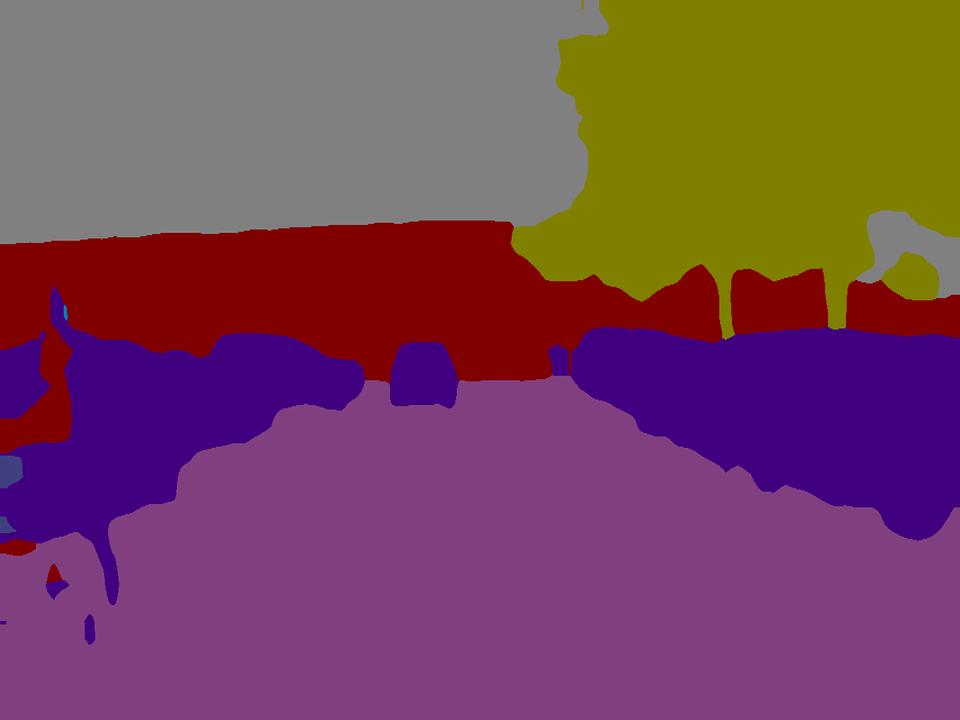} &\hspace{-0.47cm}
\includegraphics[width=0.187\linewidth, height=0.12\linewidth]{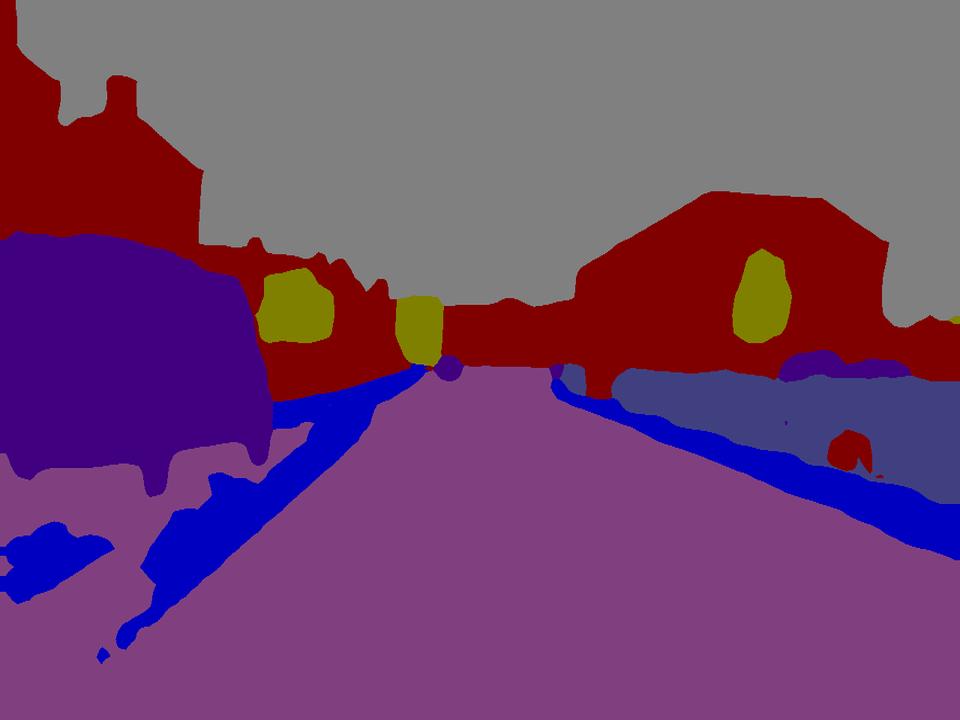} &\hspace{-0.47cm}
\includegraphics[width=0.187\linewidth, height=0.12\linewidth]{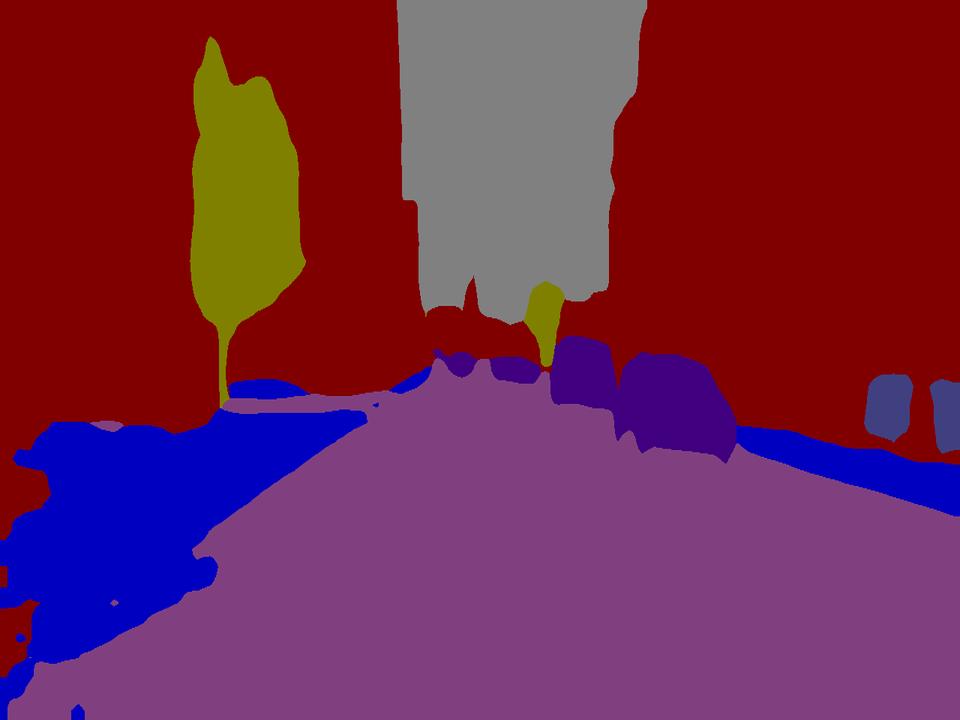} &\hspace{-0.47cm}
\includegraphics[width=0.187\linewidth, height=0.12\linewidth]{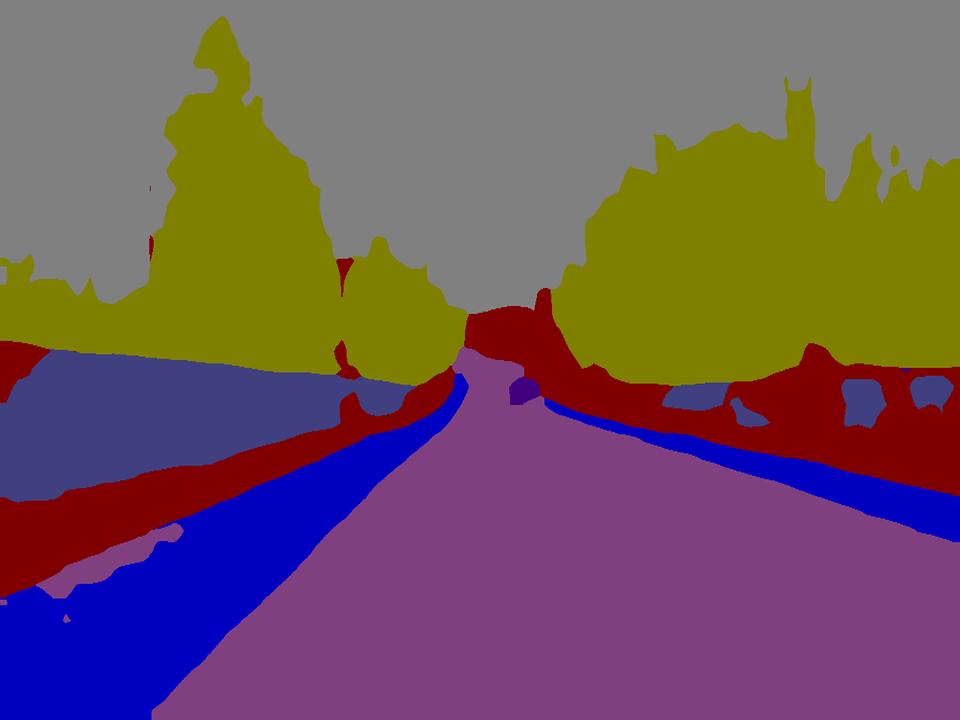}\\
\hline

\verticaltext[28pt]{\tiny}{FedDyn(0.005)} &\includegraphics[width=0.187\linewidth, height=0.13\linewidth]{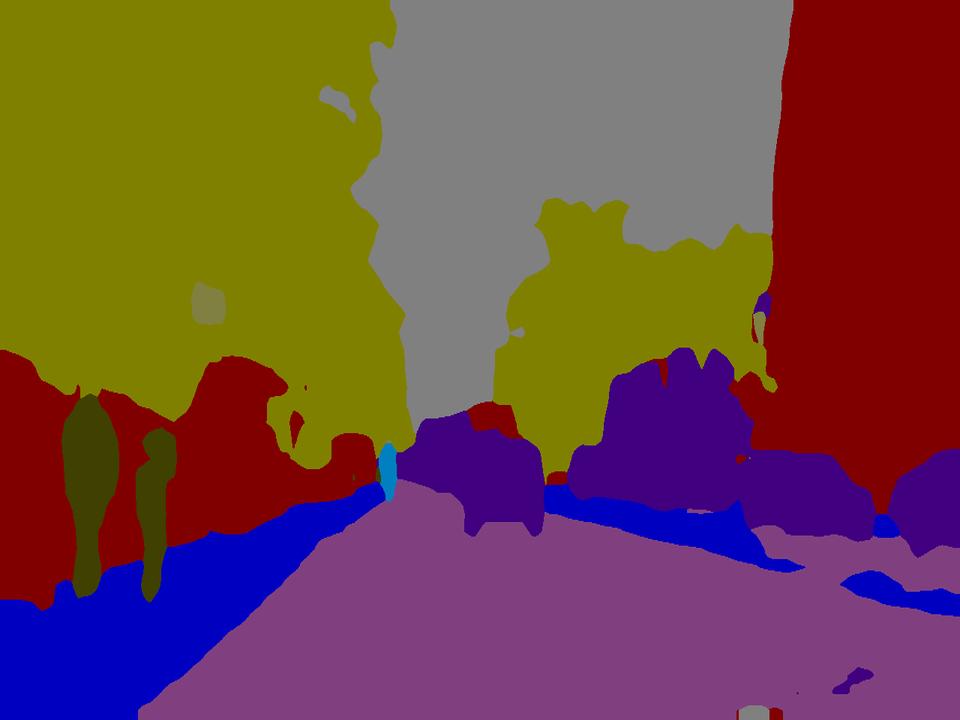} &\hspace{-0.47cm}
\includegraphics[width=0.187\linewidth, height=0.13\linewidth]{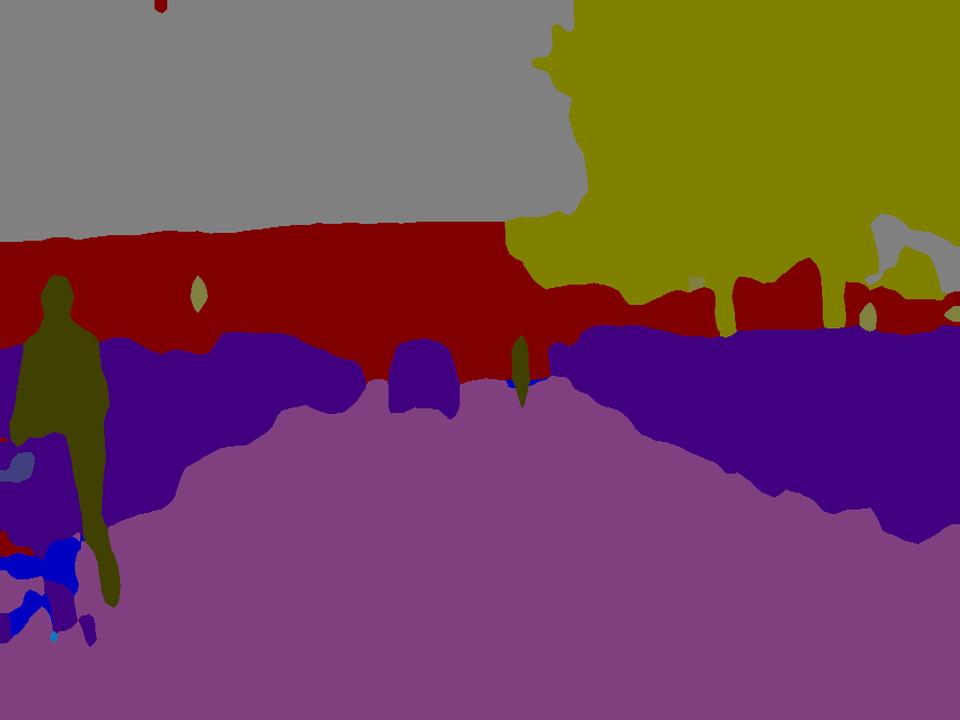} &\hspace{-0.47cm}
\includegraphics[width=0.187\linewidth, height=0.13\linewidth]{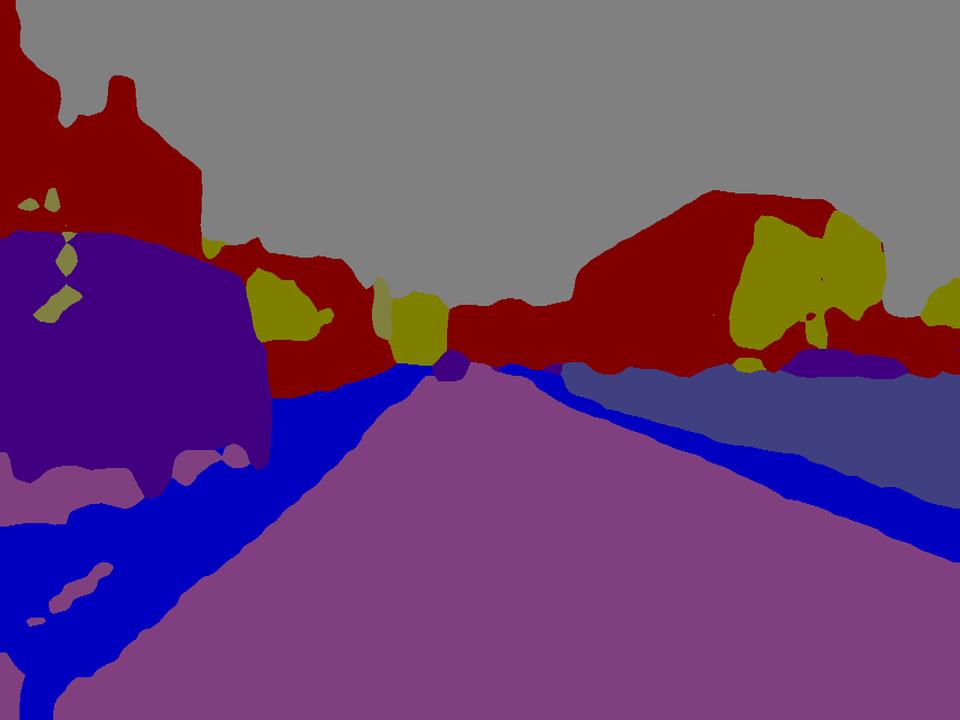} &\hspace{-0.47cm}
\includegraphics[width=0.187\linewidth, height=0.13\linewidth]{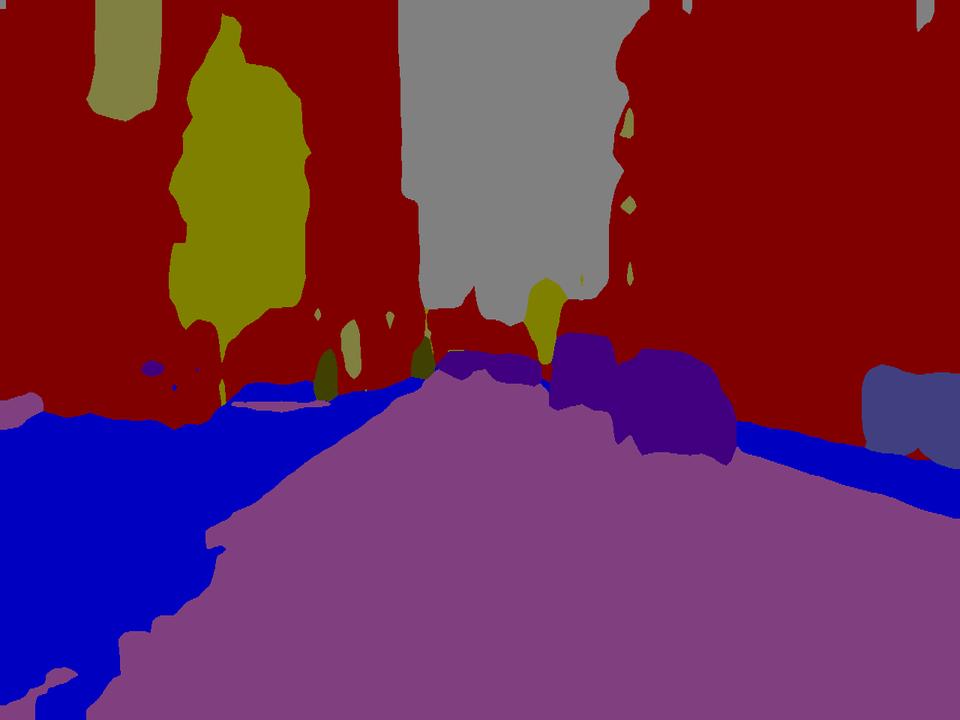} &\hspace{-0.47cm}
\includegraphics[width=0.187\linewidth, height=0.13\linewidth]{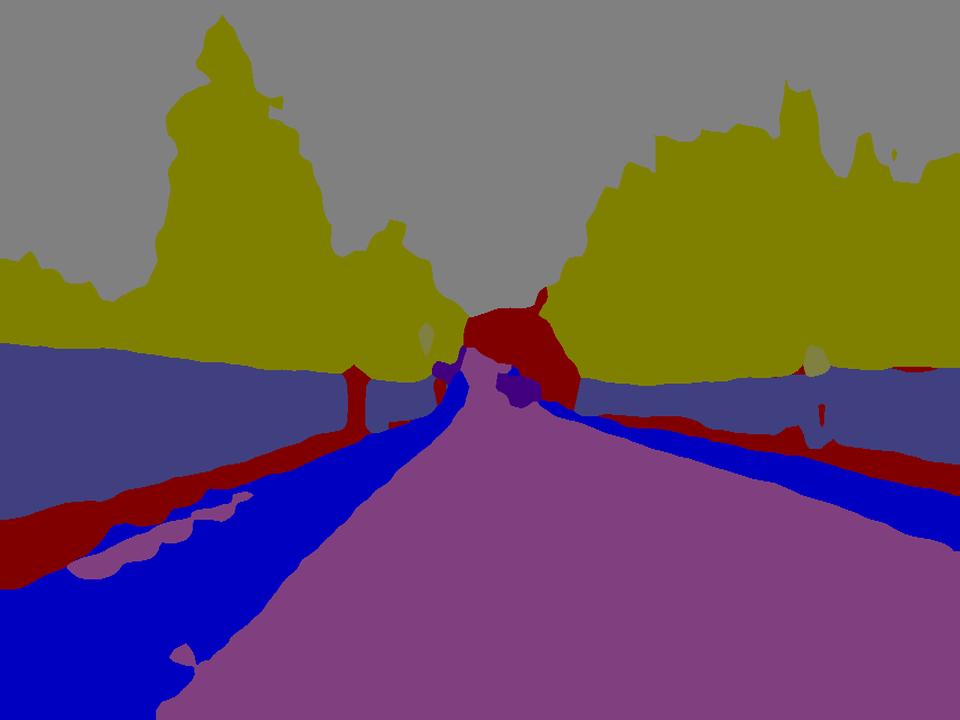}\\
\hline

\verticaltext[28pt]{\tiny}{FedProx(0.005)} &\includegraphics[width=0.187\linewidth, height=0.13\linewidth]{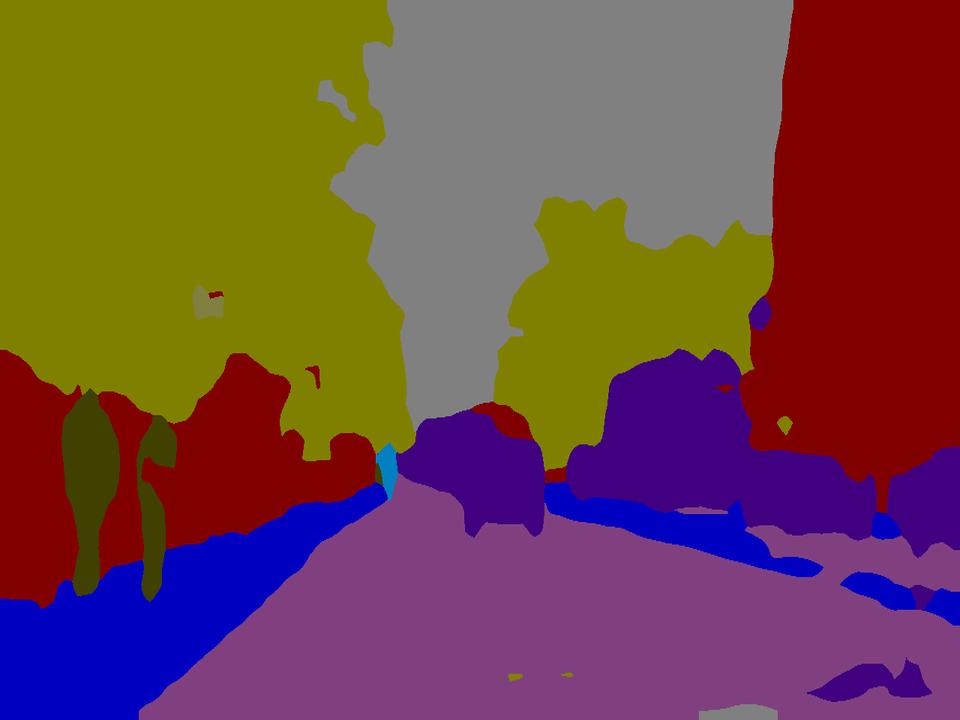} &\hspace{-0.47cm}
\includegraphics[width=0.187\linewidth, height=0.13\linewidth]{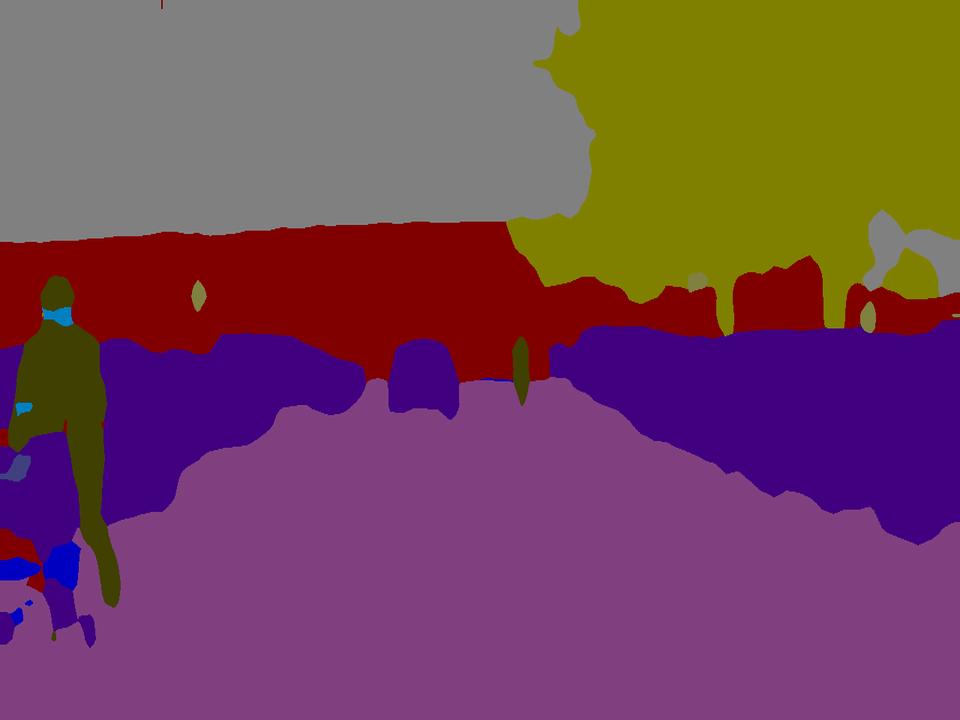} &\hspace{-0.47cm}
\includegraphics[width=0.187\linewidth, height=0.13\linewidth]{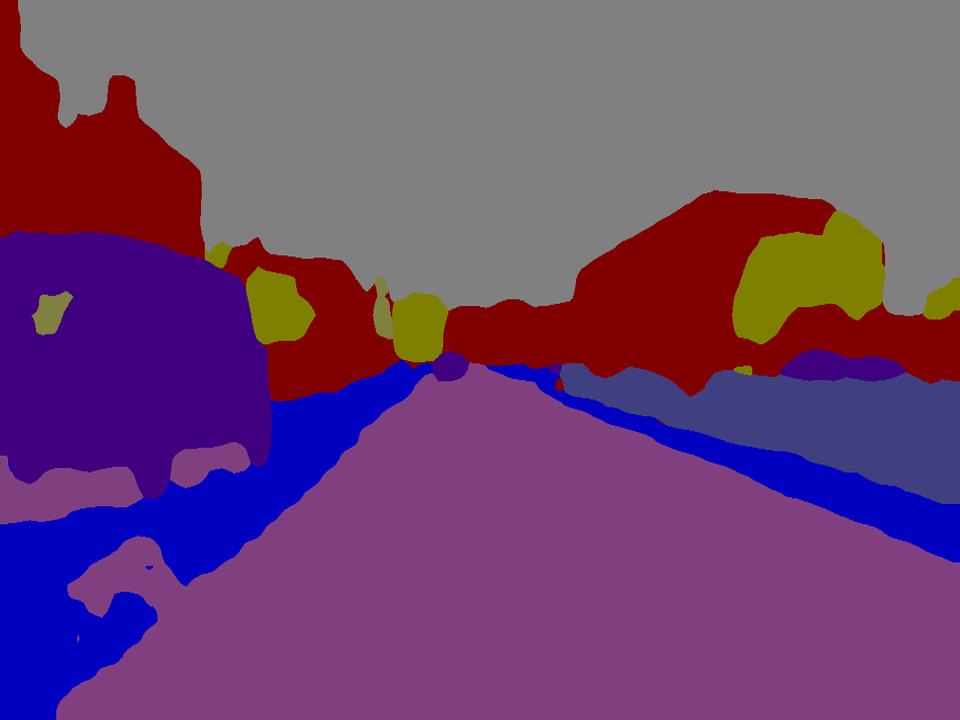} &\hspace{-0.47cm}
\includegraphics[width=0.187\linewidth, height=0.13\linewidth]{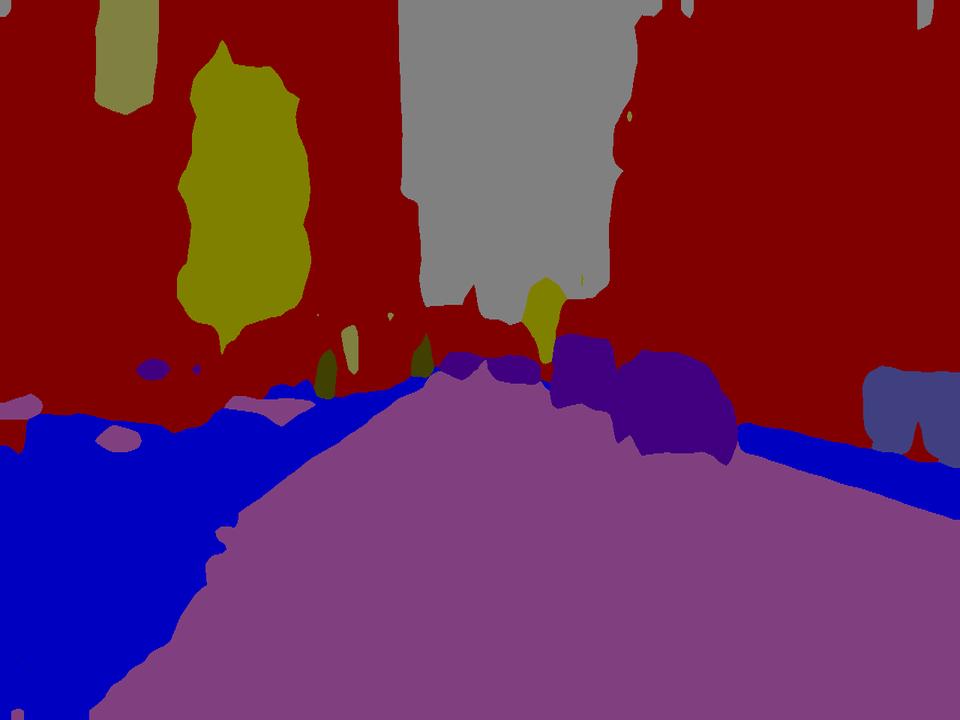} &\hspace{-0.47cm}
\includegraphics[width=0.187\linewidth, height=0.13\linewidth]{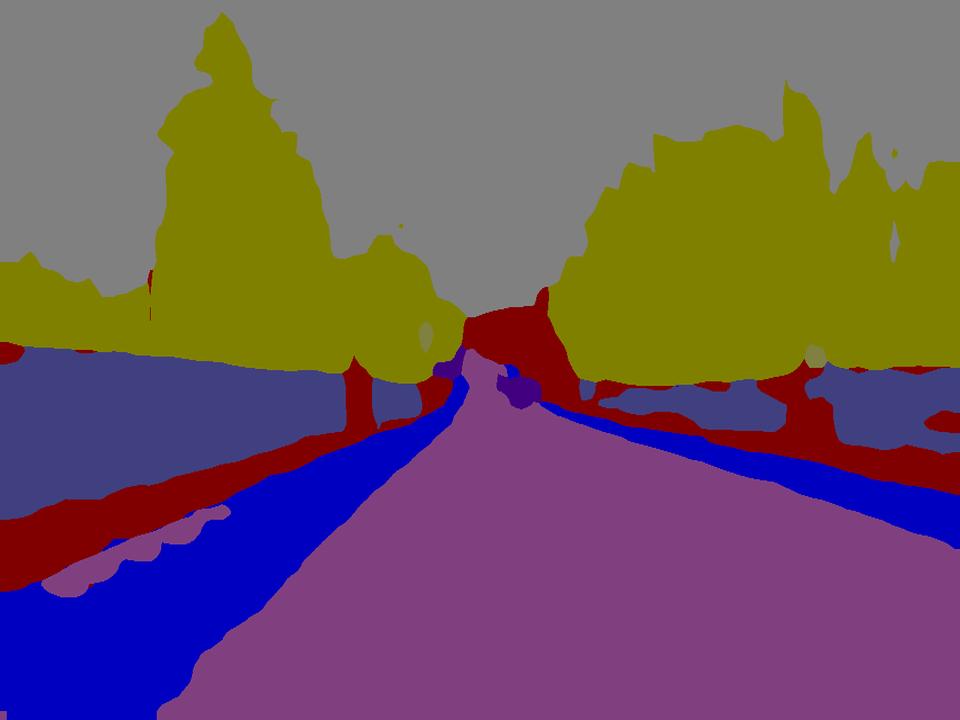}\\
\hline

\verticaltext[26pt]{\tiny}{FedNova} &\includegraphics[width=0.187\linewidth, height=0.12\linewidth]{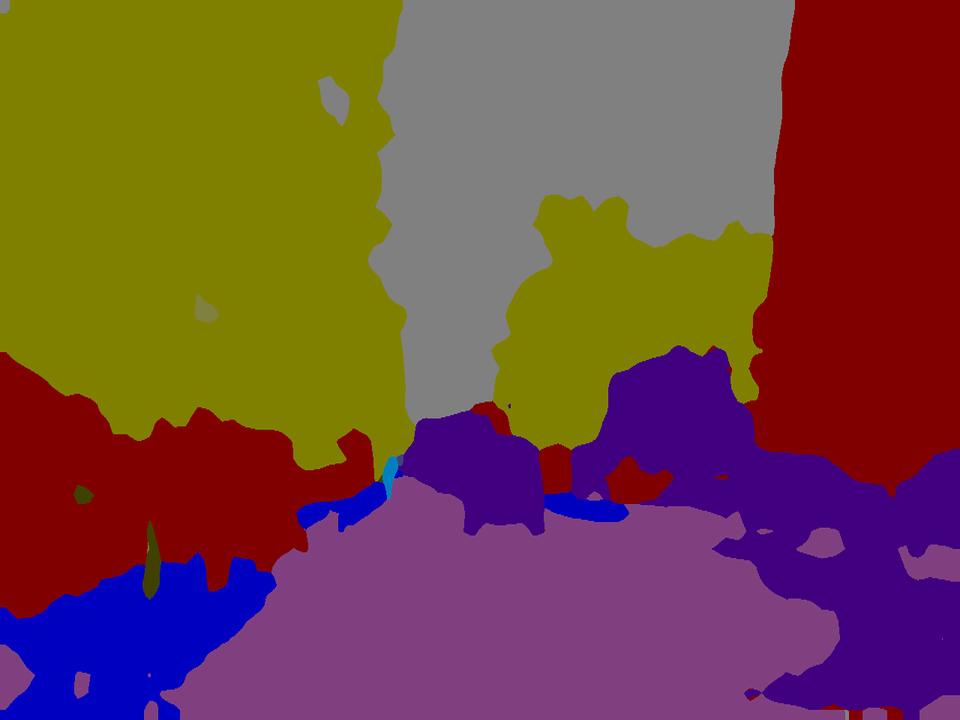} &\hspace{-0.47cm}
\includegraphics[width=0.187\linewidth, height=0.12\linewidth]{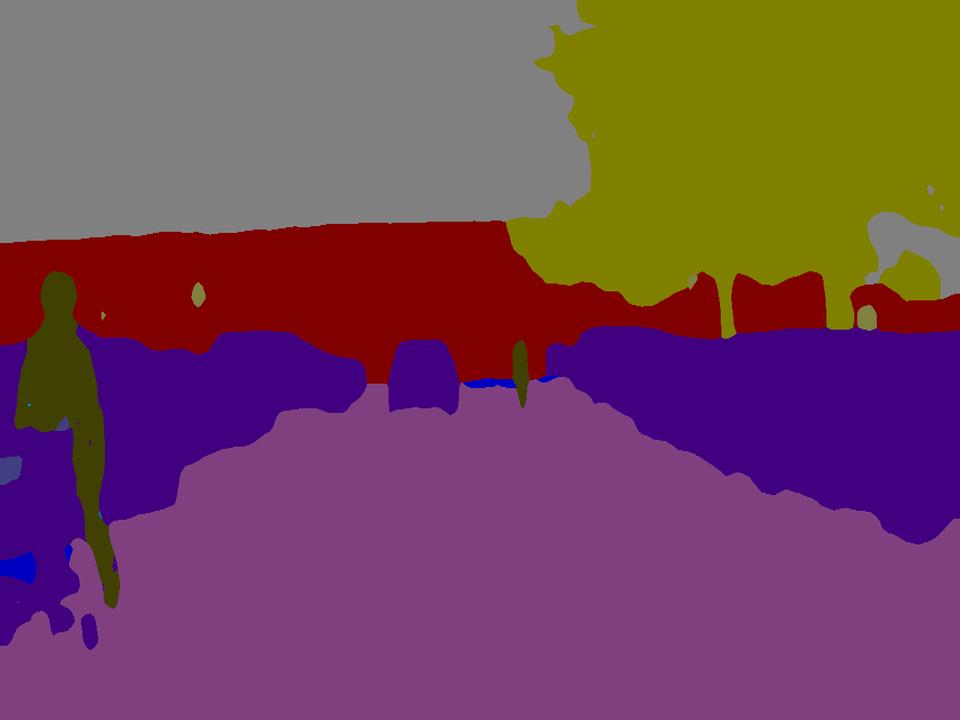} &\hspace{-0.47cm}
\includegraphics[width=0.187\linewidth, height=0.12\linewidth]{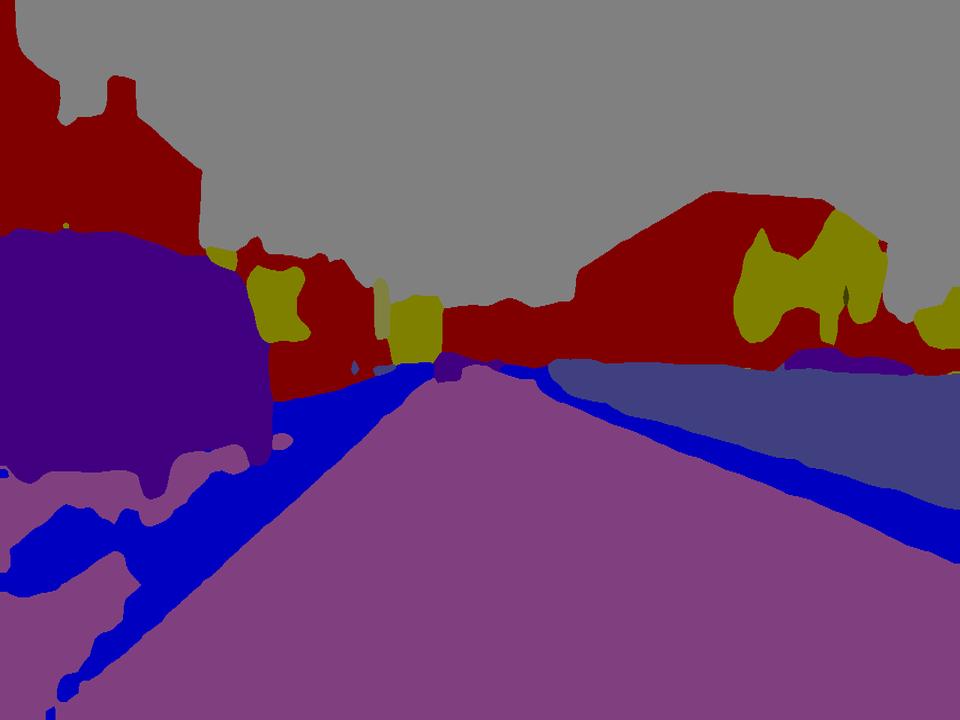} &\hspace{-0.47cm}
\includegraphics[width=0.187\linewidth, height=0.12\linewidth]{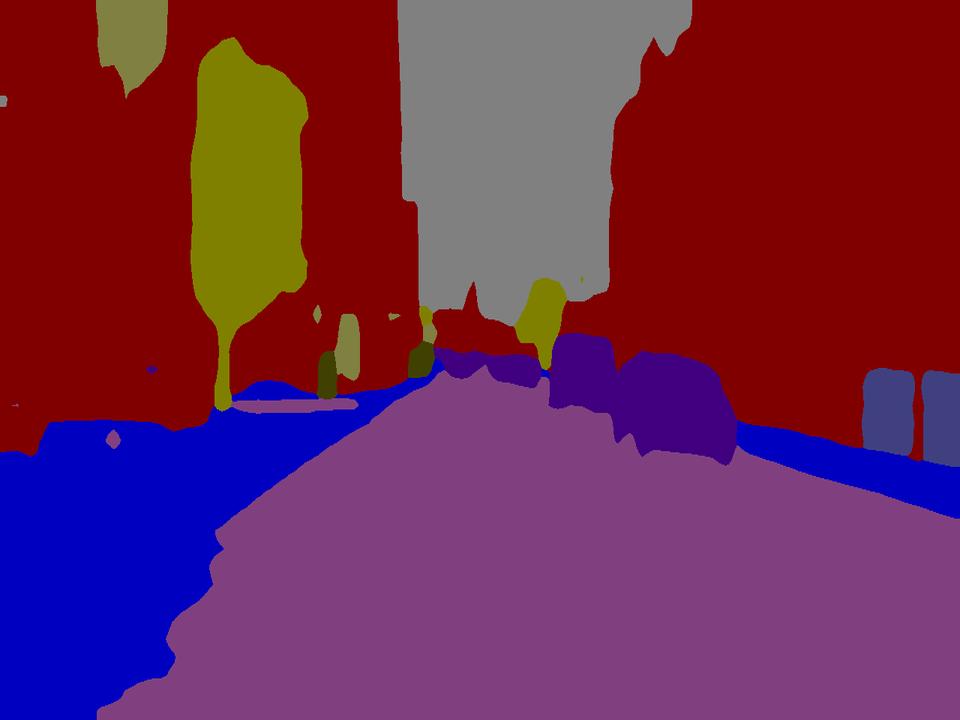} &\hspace{-0.47cm}
\includegraphics[width=0.187\linewidth, height=0.12\linewidth]{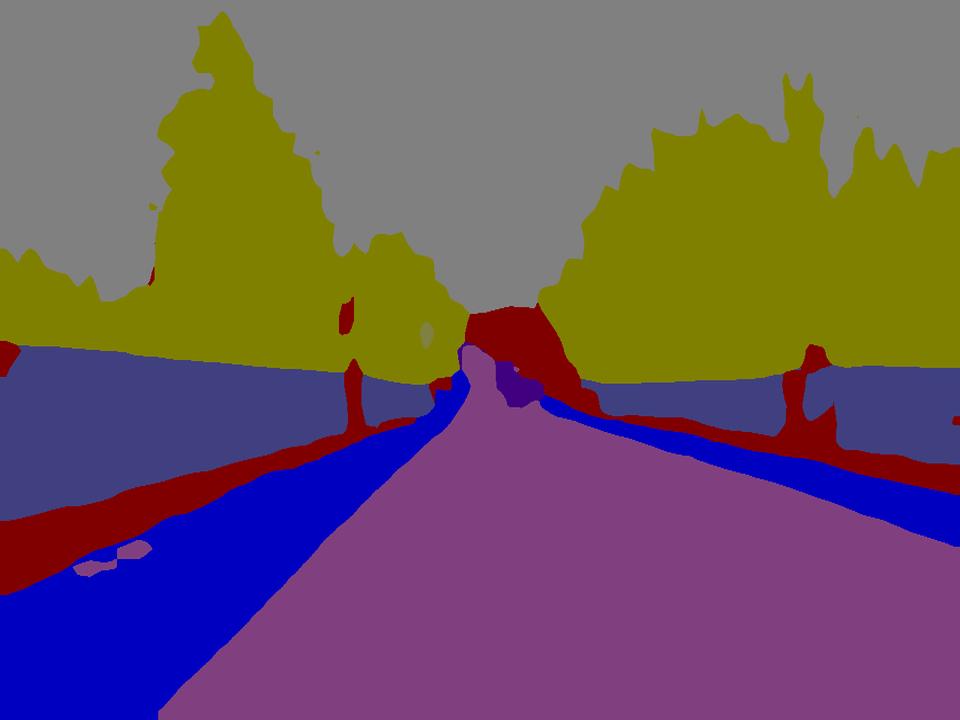}\\
\hline

\verticaltext[33pt]{\tiny}{\textbf{FedGau (Ours)}} &\includegraphics[width=0.187\linewidth, height=0.14\linewidth]{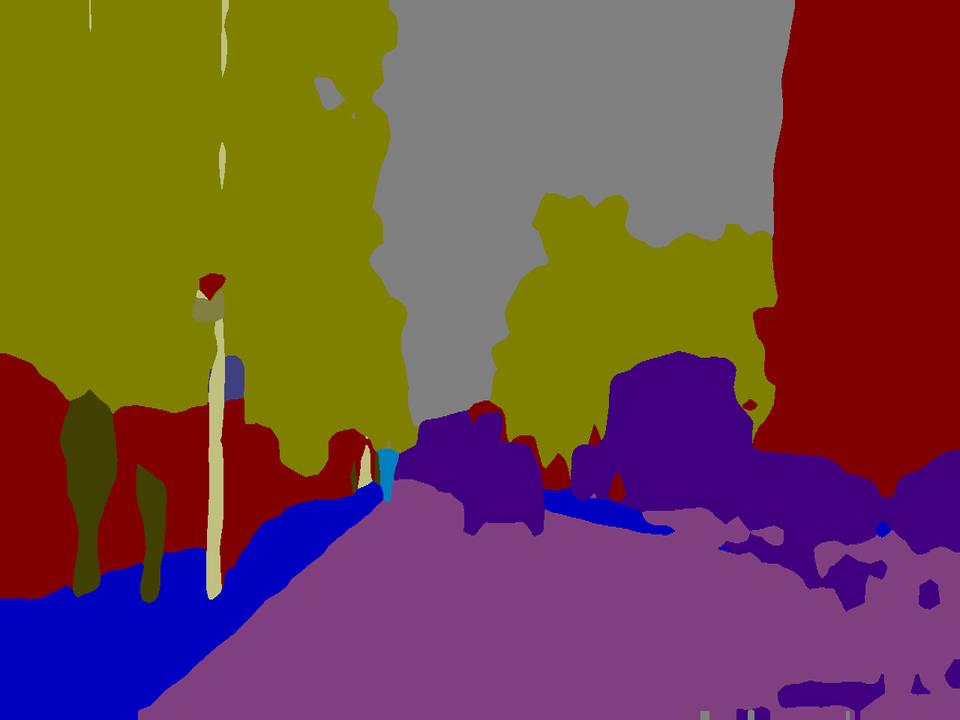} &\hspace{-0.47cm}
\includegraphics[width=0.187\linewidth, height=0.14\linewidth]{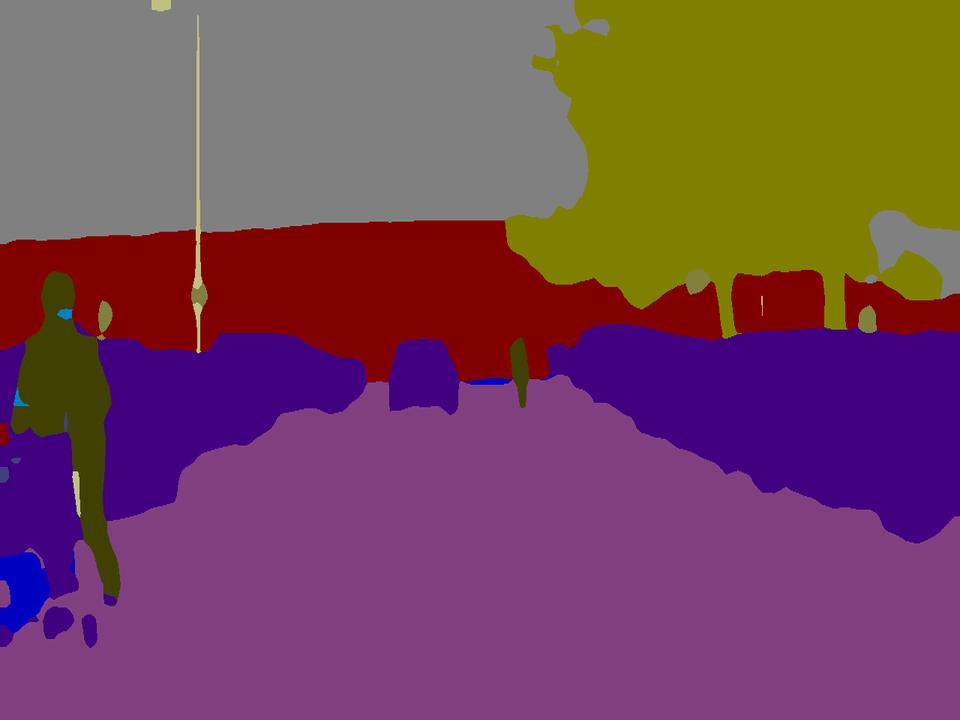} &\hspace{-0.47cm}
\includegraphics[width=0.187\linewidth, height=0.14\linewidth]{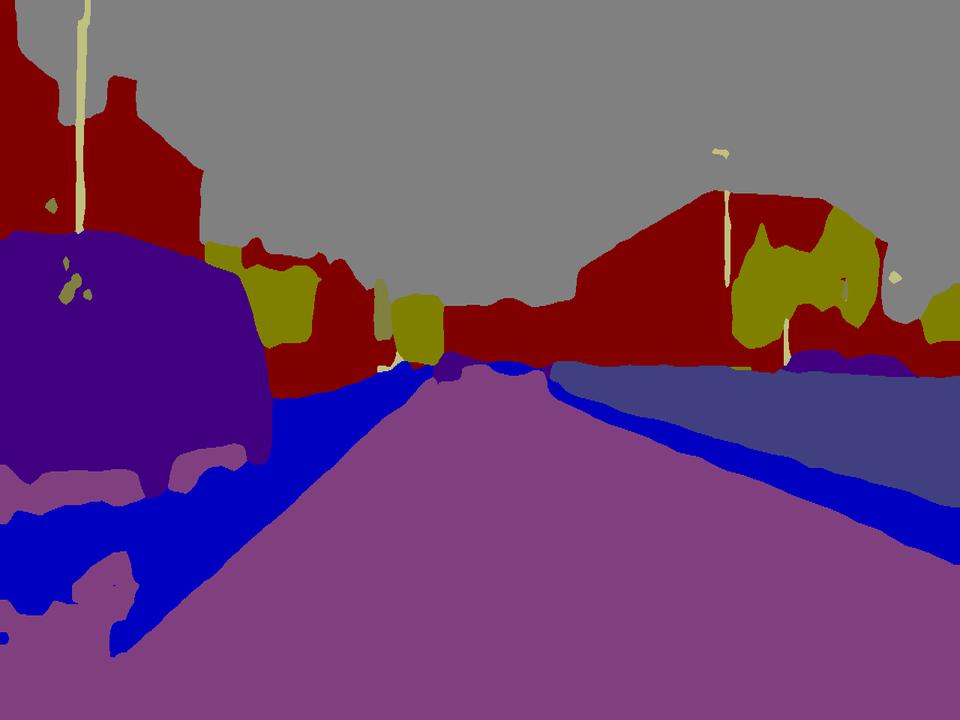} &\hspace{-0.47cm}
\includegraphics[width=0.187\linewidth, height=0.14\linewidth]{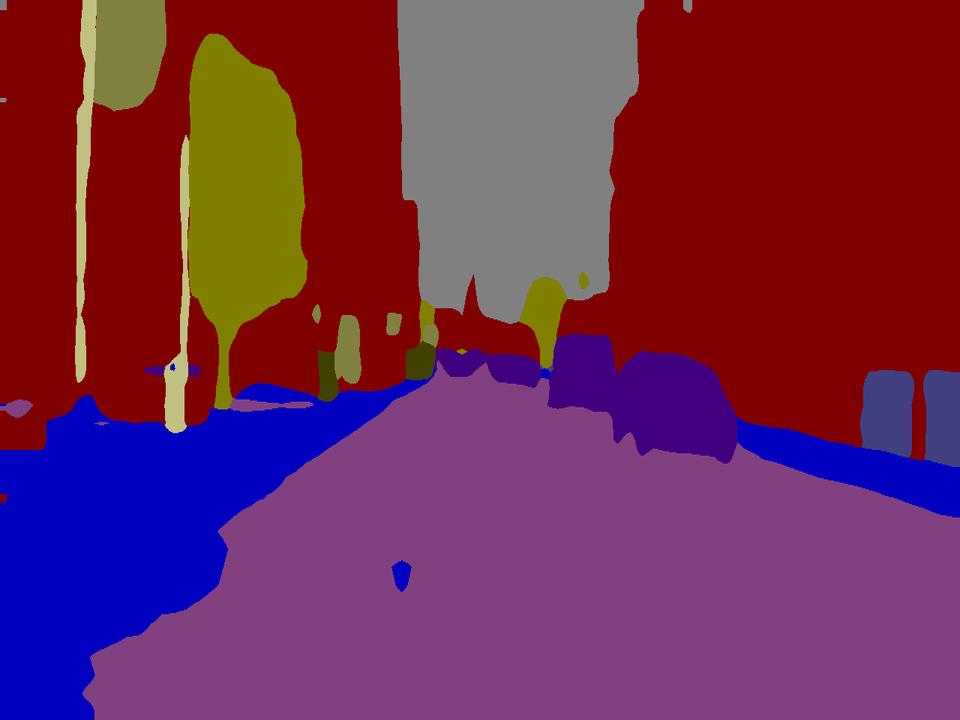} &\hspace{-0.47cm}
\includegraphics[width=0.187\linewidth, height=0.14\linewidth]{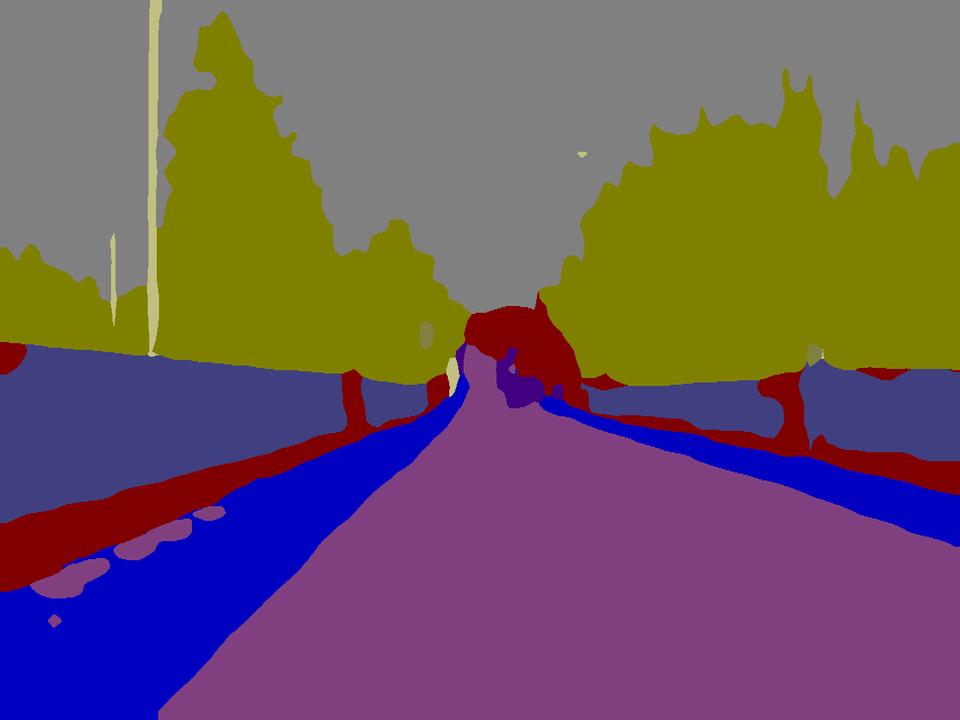}\\ \hline
\end{tabularx}
\label{tab:semantic_pred}
\end{table*}

\subsection{Evaluaton on AdapRS against StatRS}
\label{adaprs_eval}
In this section, we investigate two key aspects of AdapRS: (I) the extent of communication resource that it conserves, and (II) its impact on model performance of FL algorithms, such as FedGau and FedAvg, when compared to StatRS.

\begin{figure*}[tp]
\centering
\subfloat[Communication Resource]{\includegraphics[width=0.33\linewidth]{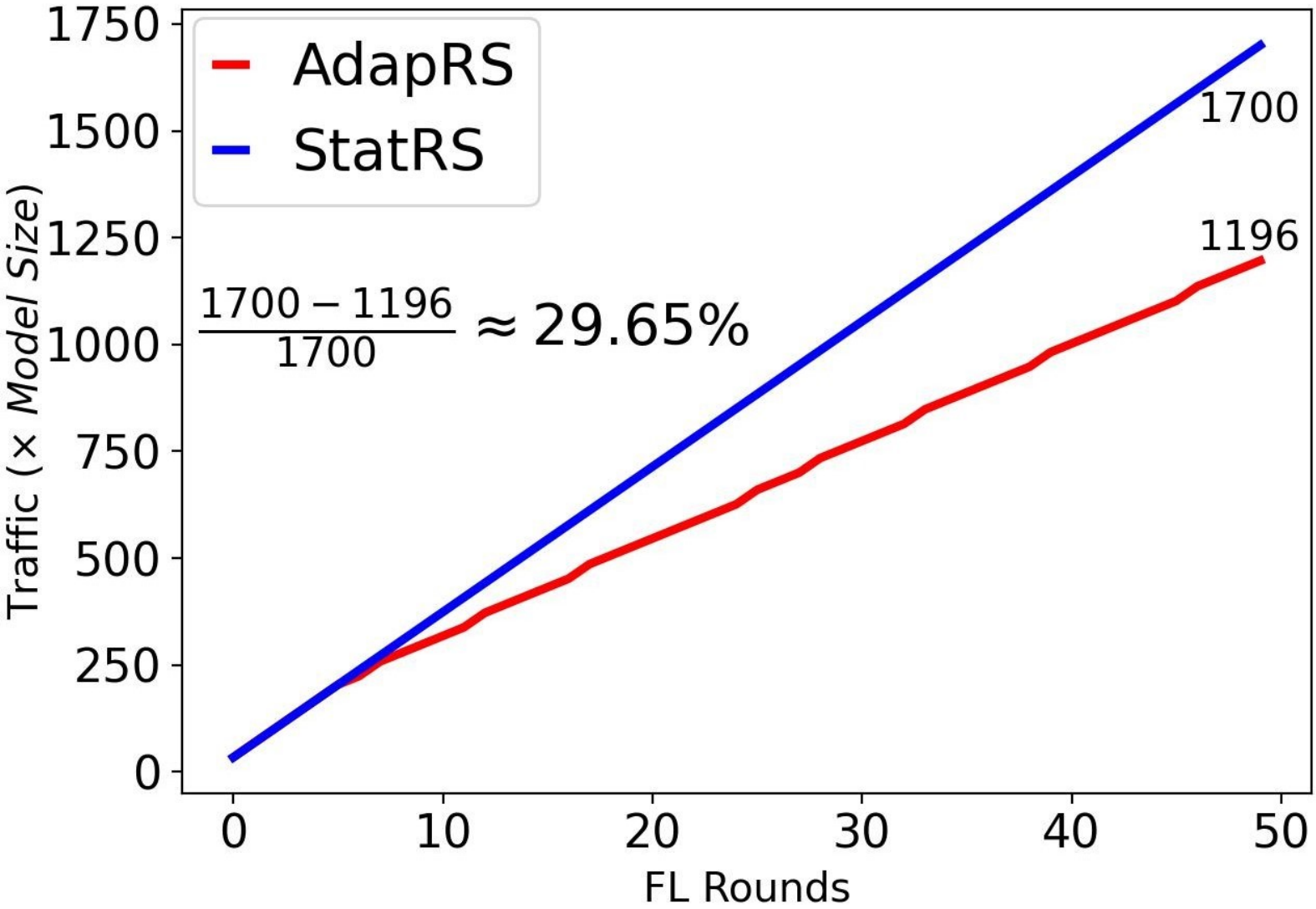}
\label{Fig.AdapRS_resource}
}
\subfloat[Performance (mIoU)]{\includegraphics[width=0.33\linewidth]{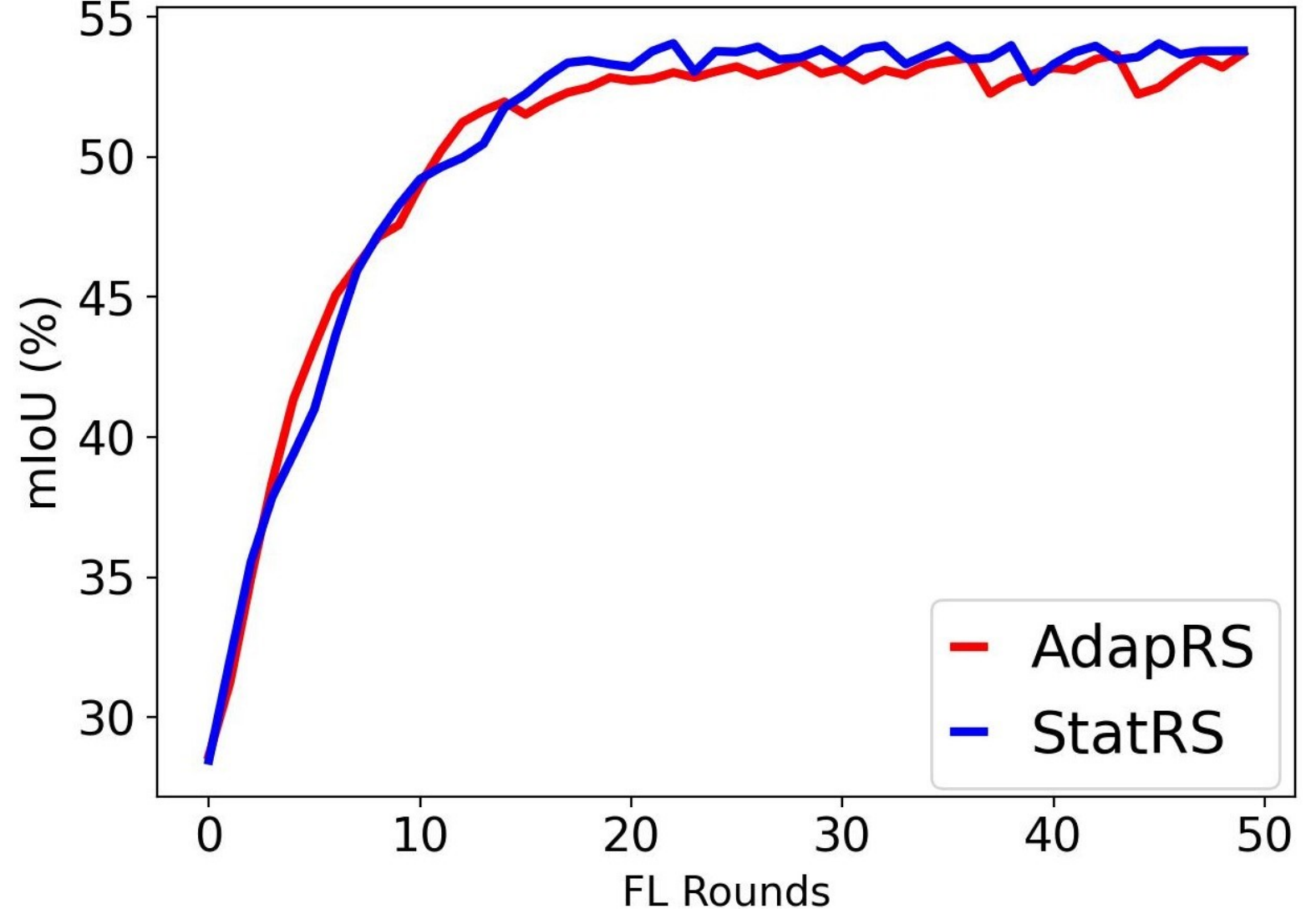}
\label{Fig.AdapRS_mIoU}
}
\subfloat[Evaluation Loss]{\includegraphics[width=0.33\linewidth]{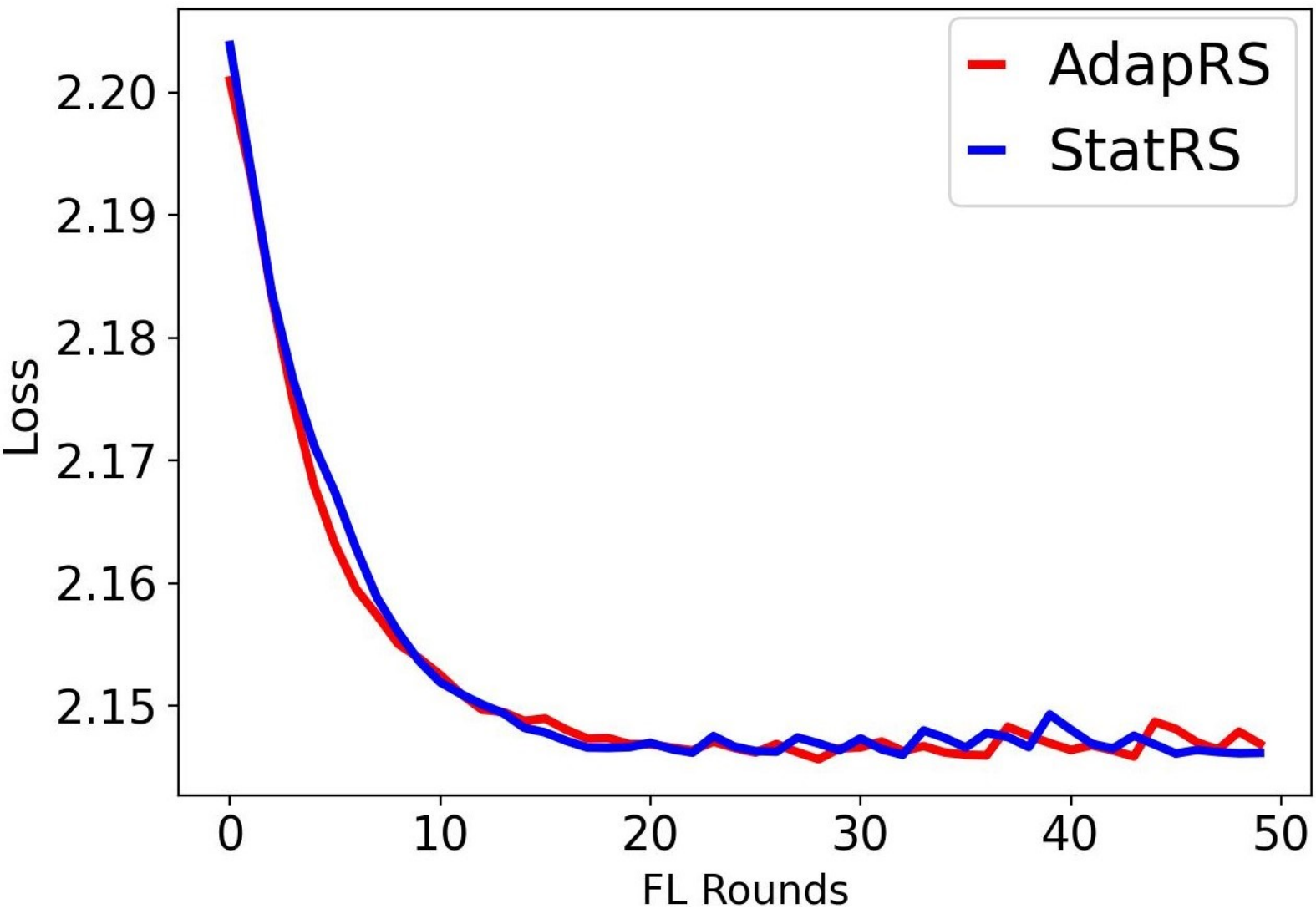}
\label{Fig.AdapRS_eval_loss}
}
\caption{AdapRS Results. AdapRS can save $29.65\%$ communication resource than StatRS while maintaining as good performance as StatRS.}
\label{Fig.AdapRS}
\vspace{-0.5cm}
\end{figure*}

\Cref{Fig.AdapRS} compares AdapRS with StatRS in terms of the communication efficiency and the effect on model performance of FL algorithms. \Cref{Fig.AdapRS_resource} illustrates that, with StatRS, the communication overhead grows linearly in proportion to the number of FL rounds, as the data exchanged in each round remains constant. In contrast, the increase in communication overhead for AdapRS slows down as training progresses, thanks to its ability to dynamically adjust the communication resource allocation. Consequently, \Cref{Fig.AdapRS_resource} demonstrates that AdapRS ultimately achieves a 29.65\% reduction in communication resource consumption when compared to StatRS.

Furthermore, \Cref{Fig.AdapRS_mIoU} and \Cref{Fig.AdapRS_eval_loss} indicate that despite the substantial communication resource savings, AdapRS does not compromise on model performance. Specifically, \Cref{Fig.AdapRS_mIoU} shows that AdapRS maintains a level of mIoU that is comparable to that of StatRS, suggesting that there is no significant drop in predictive performance for the TriSU task. Likewise, \Cref{Fig.AdapRS_eval_loss} compares the evaluation loss of AdapRS against StatRS, confirming that there is no significant increase in loss, further underlining the efficacy of AdapRS.

\begin{figure*}[tp]
\centering
\hspace{-0.03\linewidth}
\subfloat[Performance (mIoU)]{\includegraphics[width=0.49\linewidth, height=0.3\linewidth]{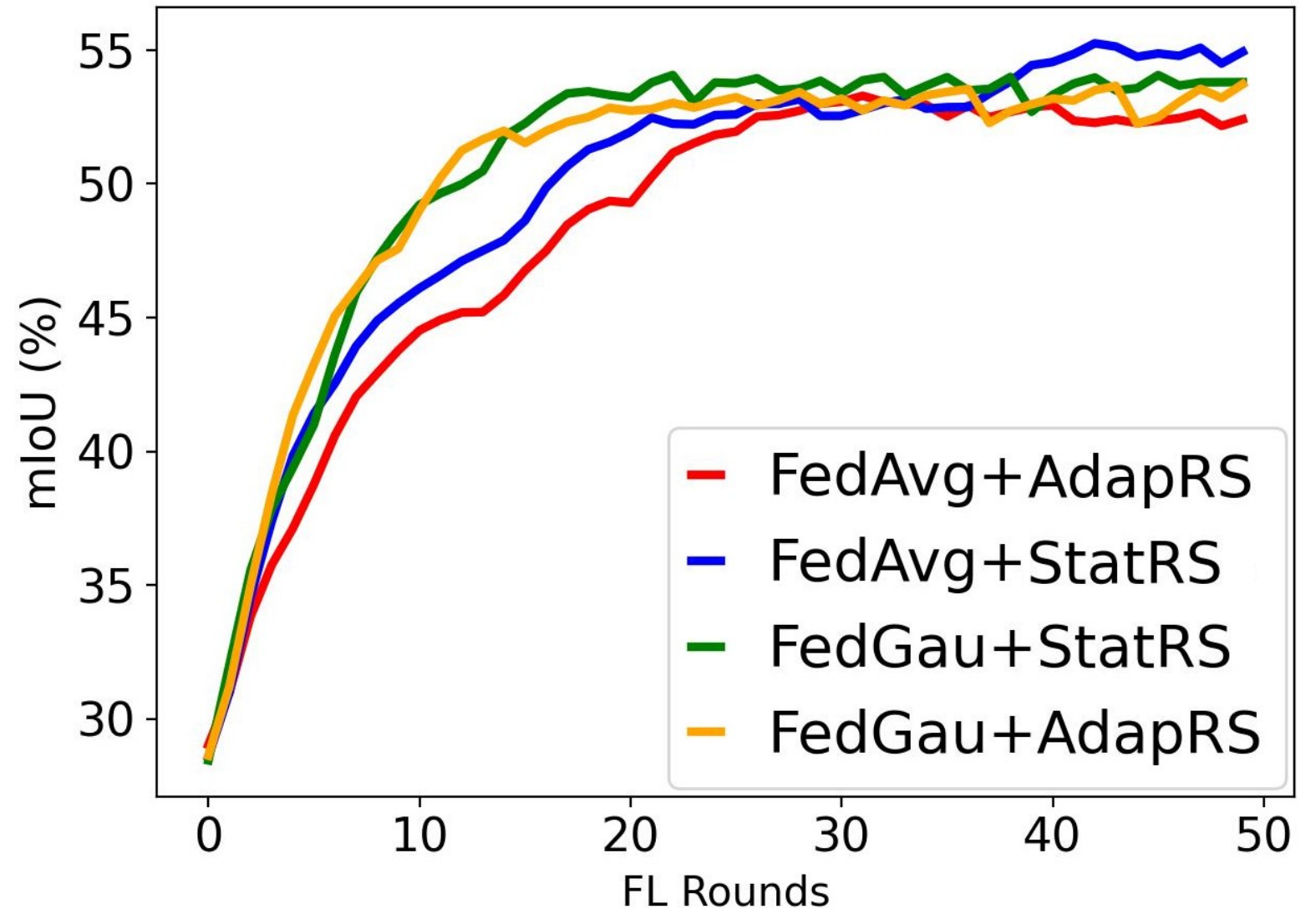}
\label{Fig.Ablation_mIoU}
}
\subfloat[Communication Resource]{\includegraphics[width=0.49\linewidth, height=0.3\linewidth]{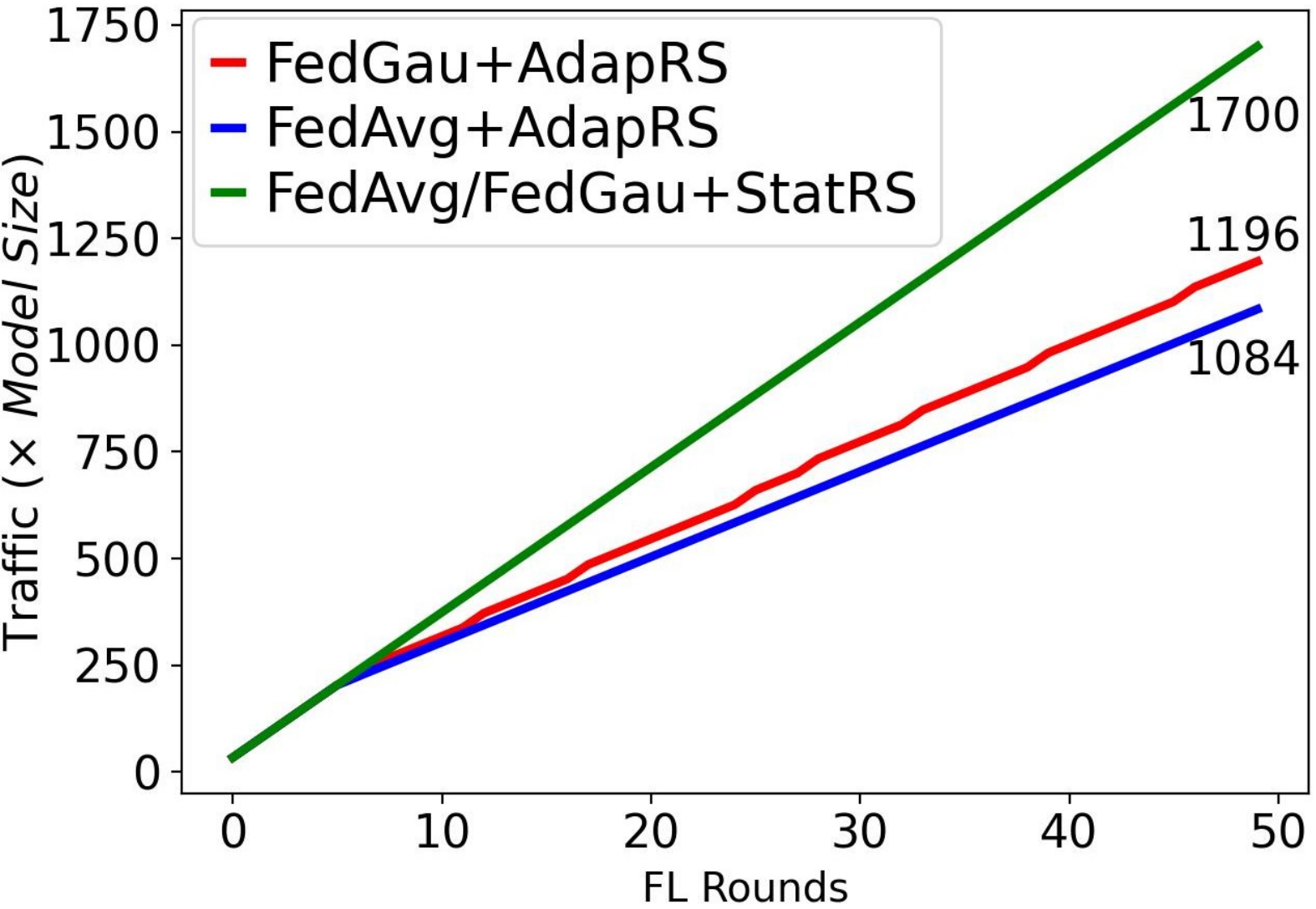}
\label{Fig.Ablation_traffic}
}
\caption{Ablation Study Results.}
\label{Fig.Ablation}
\vspace{-0.5cm}
\end{figure*}

\subsection{Ablation Study}
\label{abl_study}
This paper presents two primary contributions: FedGau and AdapRS. For in-depth exploration of their effectiveness, we conduct ablation studies with following configurations: FedGau+AdapRS, FedGau+StatRS, FedAvg+AdapRS, and FedAvg+StatRS. We compare the performance metric mIoU and communication resource consumption of these four cases. The outcomes of the ablation studies are presented in \Cref{Fig.Ablation}.

Firstly, when comparing the model performance and model convergence of aforementioned four configurations, \Cref{Fig.Ablation_mIoU} indicates that FedGau significantly enhances the convergence of the HFL TriSU model, irrespective of StatRS or AdapRS used. More specifically, configurations employing FedGau (\ie, FedGau+AdapRS and FedGau+StatRS) show faster convergence than those using FedAvg (\ie, FedAvg+AdapRS and FedAvg+StatRS). Additionally, configurations with AdapRS, whether combined with FedAvg or FedGau, underperform slightly those with StatRS, which is the small cost of saving communication resource by AdapRS. 

Secondly, we turn our attention to the communication resource consumption across the four configurations. Since FedAvg+StatRS and FedGau+StatRS share the same StatRS policy and therefore exhibit identical communication resource consumption, we merge these two cases and denote them collectively as FedAvg/FedGau+StatRS. As depicted in \Cref{Fig.Ablation_traffic}, AdapRS significantly mitigates the communication overheads of both FedAvg and FedGau. Concretely, FedAvg/FedGau+StatRS, FedGau+AdapRS, and FedAvg+AdapRS consume communication resource of 1700, 1196, and 1084 times the model size, respectively. Based on this, we can deduce that FedGau+AdapRS achieves a communication resource saving of (1700 - 1196) / 1700 = 29.65\% over FedAvg/FedGau+StatRS. Similarly, FedAvg+AdapRS reduces communication resource consumption by (1700 - 1084) / 1700 = 36.24\% compared to FedAvg/FedGau+StatRS.

\begin{figure*}[tp]
\centering
\hspace{-0.03\linewidth}
\subfloat[Cumulative sum of $QoC$]{\includegraphics[width=0.49\linewidth,height=0.26\linewidth]{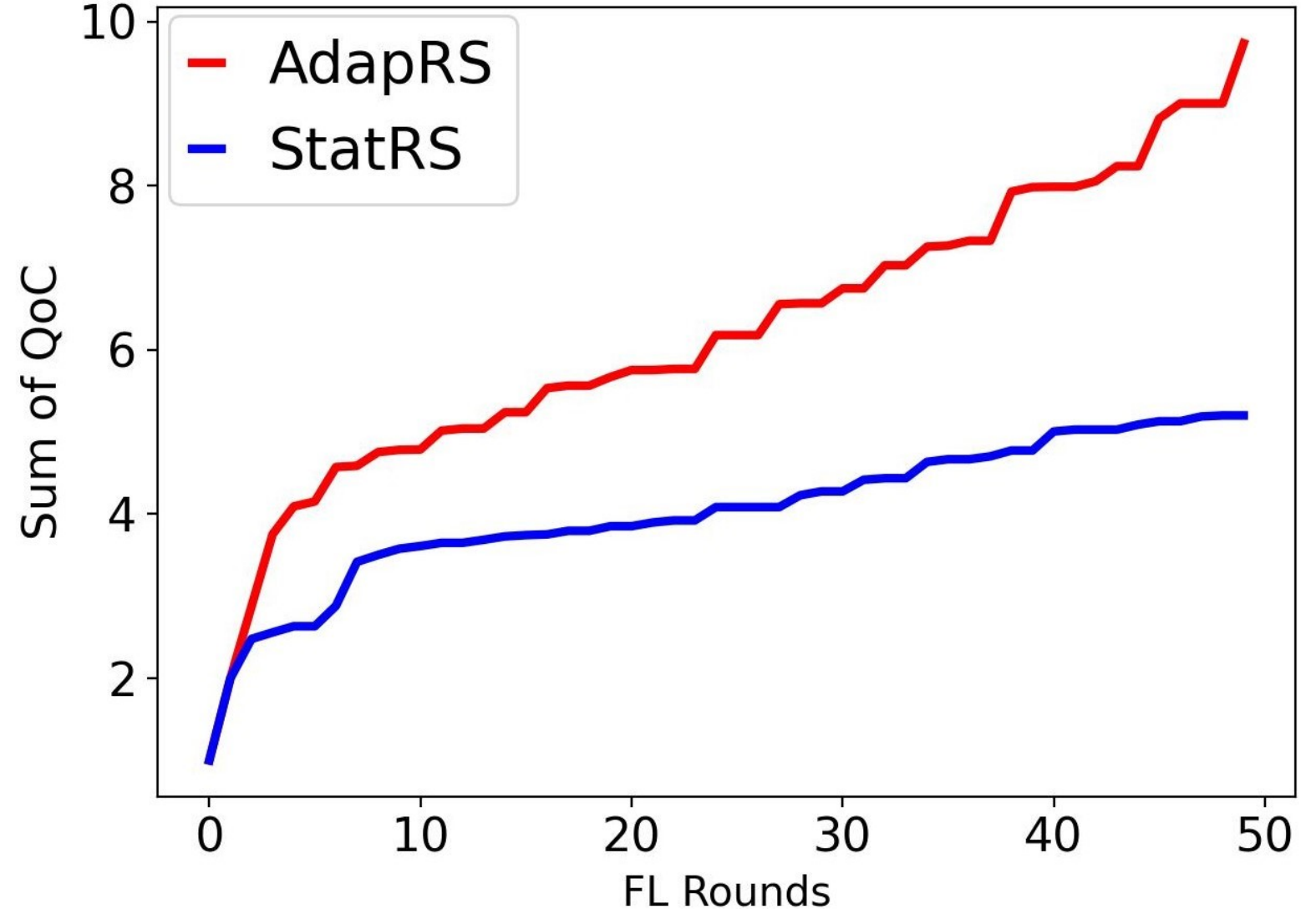}
\label{Fig.cumu_sum_qoc}
}
\subfloat[Global loss $\mathcal{L}(\omega)$ w.r.t EAI ($\tau_1$) and CAI ($\tau_2$)]{\includegraphics[width=0.49\linewidth,height=0.26\linewidth]{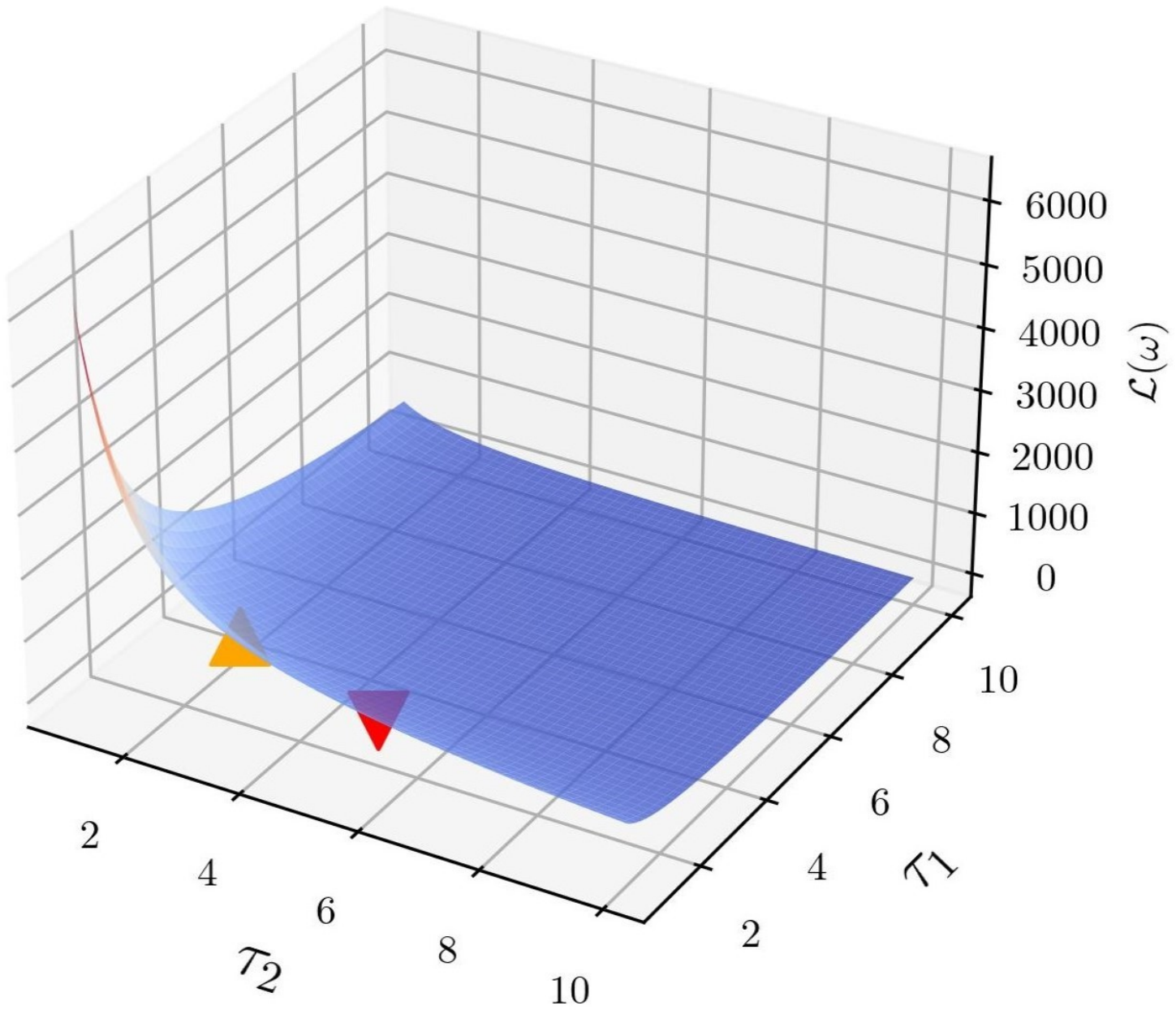}
\label{Fig.loss_opt}
}
\caption{In-depth exploration of AdapRS.}
\label{Fig.adaprs_mechanism}
\label{-0.5cm}
\end{figure*}

To gain a deeper understanding of the proposed AdapRS, \Cref{Fig.cumu_sum_qoc} compares the cumulative sum of Quality of Communication ($QoC$) between AdapRS and StatRS, where $QoC$ is defined as the ratio of the model performance increment (e.g., $\Delta$mIoU) relative to the number of model exchanges (including vehicle-edge exchange and edge-cloud exchange) in each $round$. It reveals that AdapRS presents a significantly higher cumulative sum of $QoC$ over StatRS. Specifically, we can deduce the following points: (I) At the initial phase, the cumulative sum of $QoC$ of both AdapRS and StatRS rises more rapidly than that of later stages, indicating that the initial multiple $rounds$ are crucial for substantial model performance improvements. (II) Throughout the entire training process, AdapRS consistently maintains a higher cumulative sum of $QoC$ than that of StatRS, signifying AdapRS's superior communication quality. (III) The disparity in cumulative sum of $QoC$ between AdapRS and StatRS widens as training moves forward, suggesting that AdapRS is increasingly effective in optimizing communication resource.

Moreover, \Cref{Fig.loss_opt} compares an example of AdapRS's optimized outcome of EAI (Edge Aggregation Interval) and CAI (Cloud Aggregation Interval) with StatRS's fixed EAI and CAI, at the end of the 25-th training $round$, where EAI and CAI are the major factors in HFL to schedule the communication resource. In this instance, we set StatRS's fixed EAI as 3 and CAI as 2, represented by the yellow triangle. The AdapRS's optimized EAI is 5 and CAI is 1, marked by the red triangle. Based on this, the following observations can be made: (I) AdapRS's optimized EAI and CAI results in a lower global loss compared to StatRS's fixed EAI and CAI, confirming AdapRS's superior performance over StatRS. (II) As the training progresses, AdapRS successfully reduces the frequency of model exchanges (\ie, lowering CAI), thereby conserving communication resource, which stands as the primary objective behind AdapRS.

\section{Conclusion}
\label{conclusion}
In this study, we improved TriSU model generalization in inter-city setting using HFL. We presented two major contributions: FedGau and AdapRS. FedGau is designed to overcome inter-city data heterogeneity and accelerate HFL TriSU model convergence, and AdapRS aims to save communication resource by adjusting communication resource allocation dynamically across $rounds$. We conducted comprehensive experiments and compared the results with current state-of-the-art approaches. The findings reveal that FedGau can accelerate HFL TriSU model convergence and AdapRS can reduce communication resource. Future work includes, but is not limited to, the following aspects: 1. Processing the RGB channels separately in FedGau; 2. Applying FedGau to a wider range of AD tasks. 3. Integrating multi-modal data into FedGau framework.

\end{document}